\documentclass[5p]{elsarticle}

\usepackage{lineno}%,hyperref}
\modulolinenumbers[5]

\journal{Computer Science Review}

\usepackage{graphicx}
\graphicspath{{Figures/}}
\usepackage{amsmath}
\usepackage{amssymb}
\usepackage[colorlinks = true,
            linkcolor = blue,
            urlcolor  = blue,
            citecolor = blue,
            anchorcolor = blue]{hyperref}
\usepackage{tabularx}
\usepackage{nccmath}
\usepackage{fancyref}
\usepackage{soul}
\usepackage{setspace}
\usepackage{pifont}% http://ctan.org/pkg/pifont
\usepackage{caption}
\usepackage{subcaption}
\usepackage{rotating}
\usepackage{multirow}
\usepackage{stfloats}
\usepackage{afterpage,refcount}
\usepackage{threeparttable}
\usepackage{tikz}
\usepackage{xparse}
\usepackage{breakcites}
\usepackage{xcolor,colortbl}

\definecolor{darkred}{rgb}{0.55,0,0}
\definecolor{darkbrown}{rgb}{0.4, 0.26, 0.13}
\definecolor{darkblue}{rgb}{0.0, 0.0, 0.55}
\definecolor{darkgreen}{rgb}{0.0, 0.35, 0}
\definecolor{LightBlue}{rgb}{0.7,0.7,1}

\DeclareMathOperator*{\argmin}{arg\,min}
\newcommand\norm[1]{\left\lVert#1\right\rVert}
\newcommand\abs[1]{\left\lvert#1\right\rvert}
\newcolumntype{Y}{>{\centering\arraybackslash}X}

\newcommand{\Lmk}{\vec{\ell}^{\text{ 3D}}}
\newcommand{\LmkProj}{\vec{\hat{\ell}}^{\text{ 2D}}}
\newcommand{\lmk}{\vec{\ell}^{\text{ 2D}}}
\newcommand{\lmkProj}{\vec{\hat{\ell}}^{\text{ 3D}}}
\newcommand{\Iin}{\vec{I}_{\text{in}}}
\newcommand{\Imod}{\vec{I}_{\text{mod}}}
\newcommand{\errorI}{\mathbb{E}_{\mathbf{I}}}
\newcommand{\errorReg}{\mathbb{E}_{\text{reg}}}
\newcommand{\errorlmk}{\mathbb{E}_{\text{lmk2D}}}
\newcommand{\errorLmk}{\mathbb{E}_{\text{lmk3D}}}

\renewcommand{\vec}[1]{\mathbf{#1}}

\renewcommand\tabularxcolumn[1]{m{#1}}
\NewDocumentCommand{\xrightarrows}{ O{}O{} }{%
    \mathrel{%
        \vcenter{\hbox{%
        \begin{tikzpicture}
          \node[minimum width=1cm,minimum height=1ex,anchor=south,align=center] (a){\vphantom{hg}#1\\[0.5ex] \vphantom{hg}#2};
          \draw[->] ([yshift=0.35ex]a.west) -- ([yshift=0.35ex]a.east);
          \draw[<-] ([yshift=-0.35ex]a.west) -- ([yshift=-0.35ex]a.east);
        \end{tikzpicture}
        }}%
    }%
}

\begin{document}

\begin{frontmatter}

% \title{Elsevier \LaTeX\ template\tnoteref{mytitlenote}}
% \tnotetext[mytitlenote]{Fully documented templates are available in the elsarticle package on \href{http://www.ctan.org/tex-archive/macros/latex/contrib/elsarticle}{CTAN}.}
\title{Survey on 3D face reconstruction from uncalibrated images}

\author[1]{Araceli Morales}
\ead{mariadearaceli.morales@upf.com}
\author[1]{Gemma Piella}
\ead{gemma.piella@upf.com}
\author[1]{Federico M. Sukno}
\ead{federico.sukno@upf.com}

\address[1]{Pompeu Fabra University, Barcelona, Spain}

% %% Group authors per affiliation:
% \author{Elsevier\fnref{myfootnote}}
% \address{Radarweg 29, Amsterdam}
% \fntext[myfootnote]{Since 1880.}

% %% or include affiliations in footnotes:
% \author[mymainaddress,mysecondaryaddress]{Elsevier Inc}
% \ead[url]{www.elsevier.com}

% \author[mysecondaryaddress]{Global Customer Service\corref{mycorrespondingauthor}}
% \cortext[mycorrespondingauthor]{Corresponding author}
% \ead{support@elsevier.com}

% \address[mymainaddress]{1600 John F Kennedy Boulevard, Philadelphia}
% \address[mysecondaryaddress]{360 Park Avenue South, New York}

\begin{abstract}
Recently, a lot of attention has been focused on the incorporation of 3D data into face analysis and its applications. Despite providing a more accurate representation of the face, 3D facial images are more complex to acquire than 2D pictures. As a consequence, great effort has been invested in developing systems that reconstruct 3D faces from an uncalibrated 2D image. However, the 3D-from-2D face reconstruction problem is ill-posed, thus prior knowledge is needed to restrict the solutions space. In this work, we review 3D face reconstruction methods proposed in the last decade, focusing on those that only use 2D pictures captured under uncontrolled conditions. We present a classification of the proposed methods based on the technique used to add prior knowledge, considering three main strategies, namely, statistical model fitting, photometry, and deep learning, and reviewing each of them separately. In addition, given the relevance of statistical 3D facial models as prior knowledge, we explain the construction procedure and provide a list of the most popular publicly available 3D facial models. After the exhaustive study of 3D-from-2D face reconstruction approaches, we observe that the deep learning strategy is rapidly growing since the last few years, becoming the standard choice in replacement of the widespread statistical model fitting. Unlike the other two strategies, photometry-based methods have decreased in number due to the need for strong underlying assumptions that limit the quality of their reconstructions compared to statistical model fitting and deep learning methods. The review also identifies current challenges and suggests avenues for future research.
\end{abstract}


\end{frontmatter}

\section{Introduction} \label{intro}

Facial analysis has been widely exploited in many different applications, including human-computer interaction \citep{ZhangESA2013,HansenPAMI2010}, security \citep{KaplanTITS2015,BurtonPS1999}, animation \citep{WeiseToG2011,WeiseESCA2009}, and even health \citep{RaiICCSPA2015,SuttiePediatrics2013,CerrolazaISBI2016,HennessyBP2007}. A recent trend in this field is to incorporate 3D data to overcome some of the intrinsic problems of the ubiquitous 2D facial analysis. Due to the 3D nature of the face, a 2D image is insufficient to accurately capture its geometry, as it collapses one dimension. Furthermore, 3D imaging provides a representation of the facial geometry that is invariant to pose and illumination, which are two of the major inconveniences of 2D imaging.

The advantages brought by 3D facial analysis systems come at the price of a more complex imaging process, which can often limit their scope. 3D facial information is usually captured using stereo-vision systems \citep{AlexanderCGA2010,BeelerToG2010,BeelerToG2011}, 3D laser scanners \citep{LeeACCGIT1995} (e.g. NextEngine and Cyberware), or RGB-D cameras (such as Kinect). The first two capture high quality facial scans but require controlled environments and expensive machinery. In contrast, RGB-D cameras are cheaper and easier to use, but the resulting scans are of limited quality \citep{KhoshelhamSensors2012,LYangSensors2015}.

%Indeed, many databases of 3D facial scans were gathered using these techniques (BU-3DFE database \cite{YinFG2006_BU3DFE}, USF Human ID 3D database, UOY 3D Face Database, Bosphorus \cite{SavranEWBIM2008_Bosphorus}, the Stirling / ESRC 3D Face Database, etc).

An appealing alternative to capturing a 3D scan of the face is to estimate its geometry from an uncalibrated 2D picture \citep{BoothPAMI2018,YGuoPAMI2018,TranCVPR2018}. This 3D-from-2D reconstruction alternative aims to combine the simplicity of capturing 2D images with the benefit of a 3D representation of the facial geometry.

Even though this approach is attractive, it is an inherently ill-posed: the individual facial geometry, the pose of the head and its texture (including illumination and colour) have to be recovered from a single picture, which leads to an underdetermined problem. As a consequence, there are ambiguities in the solution of the 3D-from-2D face reconstruction since a single 2D picture can be generated from different 3D faces, and it is hard to determine which one corresponds to the true geometry.

% Although using several 2D images helps constraining the reconstruction process, only with a sufficiently large amount of images the resulting 3D faces are satisfactory.

Recent methodological progress has helped to achieve remarkably convincing reconstructions, making it possible to use 3D-from-2D face reconstruction in a wide variety of fields \citep{AbatePRL2007,CaoToG2013,CorneanuPAMI2016,LTuMICCAI2018,LTuMICCAI2019}. Some methods are even able to recover local details, such as wrinkles, or to reconstruct the 3D face from images viewed under extreme conditions, such as occlusions or large head poses \citep{BoothPAMI2018,YGuoPAMI2018}.

A key to the success of 3D-from-2D reconstruction methods is the addition of prior knowledge to resolve ambiguities in the solutions. In the last decade, we can distinguish three strategies for adding this prior information, namely, statistical model fitting, photometric stereo, and deep learning. In the first one, prior knowledge is encoded in a 3D facial model, built from a set of 3D facial scans, which is fitted to the input images. In the second one, a 3D template face or a 3D facial model is combined with photometric stereo methods to estimate the facial surface normals. Approaches under this strategy generally use information from multiple images, which further constrains the problem. In the third one, the 2D-3D mapping is implemented by means of deep neural networks that, given the appropriate training data, can learn the priors necessary to relate the geometry and appearance of faces.

In this work, we review the recent research on 3D face reconstruction from one or more uncalibrated 2D images. For each of the three main strategies described above (i.e., statistical model fitting, photometric stereo, and deep learning), we summarise, compare and discuss the most relevant approaches proposed in the last decade. We also introduce a common mathematical framework to all the proposed methods whose notation is summarised in \ref{sec:Notation}.

Although there are other reviews of the field \citep{LevinePRL2009,StylianouIJIG2009,SuenSMC2007,WidanagamaachchiDICTA2008,ZollhoferEUROGRAPHICS2018,EggerToG2020}, none of them provides a in-depth and up-to-date study of the state-of-the-art research on 3D-from-2D face reconstruction. \citet{StylianouIJIG2009} presented a survey on 3D face reconstruction from 2D images, but they only covered works up to 2009. The rapid expansion of the field over the last decade and the emergence of the deep learning techniques make this work obsolete. Also, \citet{LevinePRL2009} presented a review of this topic, but narrowed it to reconstruction from single images, focusing on model fitting approaches for face recognition. More recently, \citet{ZollhoferEUROGRAPHICS2018} presented another review on 3D face reconstruction from single images, but they focused only on optimisation-based approaches, also missing methods based on deep learning and photometric stereo. This last work was updated in 2020 by \citet{EggerToG2020}, who presented a very extensive survey especially focused on statistical facial models, reviewing 3D data acquisition, 3D facial model construction, and 2D image generation. Even though they included 3D face reconstruction, it is only reviewed as an application of the 3D facial models and they focused on single-image reconstruction (both RGB and RGB-D) and model fitting, discussing briefly methods based on deep learning. Other surveys of 3D face reconstruction, such as \citet{SuenSMC2007} and \citet{WidanagamaachchiDICTA2008}, study the general strategies, discussing their strengths and drawbacks, but without providing an in-depth review of the most relevant works within each of them. Thus, our review complements the existing ones by providing an updated, comprehensive and complete review of 3D-from-2D reconstruction methods.

The remainder of this survey is organised as follows: in Section \ref{sec:StatisticalModels}, we first introduce the most popular way of constructing a statistical 3D facial model and list the publicly available models that have been most used for 3D-from-2D face reconstruction. In Section \ref{sec:ModelFittingMethods}, we review methods based on statistical model fitting. In Sections \ref{sec:Photometry} and \ref{sec:DL}, we review the photometry-based and deep learning approaches, respectively. Section \ref{sec:OtherML} groups methods that use other machine learning approaches, such as regression. Finally, in Section \ref{sec:Applications}, we review the main applications of 3D face reconstruction from uncalibrated images, and in Section \ref{sec:Conclusions} conclusions are provided.

\section{Background: Statistical 3D Facial Models} \label{sec:StatisticalModels}

As stated in the introduction, 3D-from-2D face reconstruction is an ill-posed problem, thus it requires some kind of prior knowledge to resolve the otherwise underdetermined solution.

Statistical 3D facial models are the most popular way of adding this prior information since they encode the geometric variations of the face, possibly in conjunction with the appearance. These models consist of a mean face along with the modes of variation of its geometry and appearance. Fitting a 3D facial model to a photograph is done by estimating, apart from the model parameters, the 3D pose and illumination such that the projection into the image plane of the resulting 3D face produces an image as similar as possible to the given picture.

In this section, we explain how the 3D facial models are built and provide a list of the most popular ones that are publicly available. We refer to \citep{EggerToG2020} for a detailed review on statistical facial models.

% \subsection{3D facial models} \label{subsec:3Dmodel}

\subsection{Construction of a 3D facial model} \label{subsec:Build3Dmodel}
The most widespread statistical models of 3D faces are the 3D Morphable Models (3DMM), which were introduced to the community by \citet{BlanzVetterSIGGRAPH1999}. A 3DMM consists of a shape (i.e., geometry) model and, optionally, an albedo (a.k.a texture or colour) model, separately constructed using principal component analysis (PCA). In this work, we use texture, albedo or colour indistinctly, and we explicitly indicate when the lighting is modelled separately from raw colour.

Let $M$ be the number of 3D faces in the training set and $N$ the number of vertices in each mesh. Let $\vec{x} = (x_1,y_1,z_1, \cdots, x_N, y_N, z_N) \in \mathbb{R}^{3N}$ be the shape vector of a mesh, and $\vec{c} = (R_1,G_1,B_1, \cdots, R_N, G_N, B_N) \in [0,1]^{3N}$ the albedo vector that contains the $R$ (red), $G$ (green), and $B$ (blue) values of the RGB colour model for each of the $N$ vertices. The idea behind the 3DMM is that, if the set of 3D faces is sufficiently large, one can express any new textured shape as a linear combination of the shapes and textures of the training 3D faces:
\begin{ceqn}
    \begin{equation*} \label{eq:LI_shapes}
    \vec{x}_{\text{new}} = \sum_{m=1}^M a_m \vec{x}_m, \quad \vec{c}_{\text{new}} = \sum_{m=1}^M b_m \vec{c}_m 
    \end{equation*}
\end{ceqn}
with $a_m, b_m \in \mathbb{R} \ \forall m = 1,\cdots,M$.

Thus, we can parametrise any new face by its shape $\vec{x}_{\text{new}} = (a_1,\cdots, a_M)^\text{T}$ and albedo $\vec{c}_{\text{new}} = (b_1,\cdots, b_M)^\text{T}$. However, this parametrisation gets more complicated when the number of shapes in the training set $M$ is large. PCA helps compressing the data, performing a basis transformation to an orthogonal coordinate system defined by the eigenvectors $\vec{\phi}_i$ and $\vec{\psi}_i$ of the covariance matrices computed over the shapes and albedos in the training set. In the orthogonal basis given by PCA,
\begin{ceqn}
\begin{align}
    &\vec{x}_{\text{new}} = \overline{\vec{x}} +\sum_{i=1}^{M-1} \alpha_i \vec{\phi}_i = \overline{\vec{x}} + \vec{\Phi}\vec{\alpha}, \label{eq:shapeModel} \\ 
    &\vec{c}_{\text{new}} = \overline{\vec{c}} + \sum_{i=1}^{M-1} \beta_i \vec{\psi}_i = \overline{\vec{c}} + \vec{\Psi}\vec{\beta}, \label{eq:textureModel}
\end{align}
\end{ceqn}
with $\overline{\vec{x}} = \frac{1}{M}\sum_{m=1}^M \vec{x}_m$ the mean shape, $\vec{\alpha} = (\alpha_1,$ $\dots,\alpha_{M-1})^\text{T} \in \mathbb{R}^{M-1}$ the shape parameters of the model, and $\vec{\Phi} = (\vec{\phi}_1, \dots,$ $\vec{\phi}_{M-1}) \in \mathbb{R}^{3N\times(M-1)}$ the shape basis matrix of the model; $\overline{\vec{c}}$, $\vec{\beta}$ and $\vec{\Psi}$ are analogously defined for the texture. The probability of the shape parameters $p(\vec{\alpha})$ is given by
\begin{ceqn}
\begin{equation}\label{eq:prior_prob}
    p(\vec{\alpha}) \propto \exp{\left[-\frac{1}{2}\sum_{i=1}^{M-1} \left(\frac{\alpha_i}{\sigma_{\alpha_i}}\right)^2\right]}
\end{equation}
\end{ceqn}
where $\sigma_{\alpha_i}^2$ are the eigenvalues of the corresponding eigenvectors $\vec{\phi}_i$. The probability of the albedo parameters $p(\vec{\beta})$ is defined analogously.

Finally, the shape model of the 3DMM is defined by the mean shape, $\overline{\vec{x}}$, the eigenvectors of the shape covariance matrix, $\vec{\Phi}$, and the corresponding eigenvalues, $\{\sigma_{\alpha_i}^2\}_{i = 1}^{M-1}$. Similarly, the albedo model is given by $\overline{\vec{c}}$, $\vec{\Psi}$, and $\{\sigma_{\beta_i}^2\}_{i = 1}^{M-1}$.

% Notice that, when computing PCA, the dimensionality of the parametrisation is reduced from $M$ to $M-1$, which is not a great improvement when $M$ is large. However, as we have mentioned before, the eigenvalues of the model are the variances of the training dataset along the dimensions represented by the corresponding eigenvectors. Thus, to reduce the dimensionality considerably, we can keep the dimensions that represent most of the variance of the training set. 

However, some of the variation modes (eigenvectors $\vec{\phi}_i$, $\vec{\psi}_i$) may have very small variance (eigenvalues $\sigma_{\alpha_i}^2$, $\sigma_{\beta_i}^2$), thus they are dispensable. Keeping only the directions that represent most of the variance of the training set allows us to reduce the dimension of the data, which is very useful when $M$ is large. Assuming the eigenvalues $\sigma_{\xi_i}^2$ (denoting either $\sigma_{\alpha_i}^2$ or $\sigma_{\beta_i}^2$) are ordered in descending order, the $\widetilde{M}$ first eigenvectors with higher eigenvalues
\begin{ceqn}
\[
\widetilde{M} = \argmin_{1 \leq m \leq M-1} \left\{ \sum_{i=1}^m \sigma_{\xi_i}^2 \geq \eta V_\text{total} \right\}.
\]
\end{ceqn}
keep the $100 \eta \%$ of the total variance $V_\text{total}$.

Although most of the existing 3D statistical facial models are based on the procedure explained above, this method has two limitations that have been noted by several researchers. Firstly, PCA estimates basis vectors that globally model the input data, so subtle information, such as wrinkles, is not captured, and thus reconstructing facial details by fitting a 3DMM becomes a hard task. Some works  \citep{NeumannToG2013,BruntonECCV2014,HJinCAGD2017,FerrariToM2017,LuthiPAMI2018} highlighted the importance of modelling facial deformations locally and proposed different approaches to do so. \citet{NeumannToG2013} and \citet{FerrariToM2017} proposed to decompose the matrix of the training shapes by imposing sparse components. \citet{BruntonECCV2014} applied a wavelet transform to every training shape, obtaining a multi-scale decomposition of the surface, and computed localised multilinear models on the estimated wavelet coefficients. \citet{HJinCAGD2017} applied non-negative matrix factorisation (NMF) since it decomposes a shape into localised features. And, finally, \citet{LuthiPAMI2018} modelled shape variations using Gaussian processes, which provide a way of adding local models to global models, thus combining the information at multiple scales.

The second drawback of 3DMMs was noted by \citep{RanjanECCV2018,BouritsasICCV2019,ZJiangCVPR2019}, who argued that facial shape variations are not perfectly linear and thus cannot be modelled accurately using linear models. Their approach consists in learning a latent space of facial deformations using a mesh-to-mesh autoencoder. \citet{RanjanECCV2018} and \citet{BouritsasICCV2019} modelled all the shape variations in a single latent space, as opposed to \citet{ZJiangCVPR2019}, who estimated two separated latent spaces, one corresponding to identity-related deformations and the other one corresponding to expression-related deformations. Whereas \citet{RanjanECCV2018} and \citet{ZJiangCVPR2019} used spectral convolutional operators, \citet{BouritsasICCV2019} proposed a spiral convolution that uses anisotropic filters, which allow
a one-to-one mapping between the neighbours of a vertex and the parameters of the local filter.

% Although most of the existing 3D statistical facial models are based on the procedure explained above, \citet{HJinCAGD2017} claimed that a drawback of PCA is that it learns the data globally, so subtle information, such as wrinkles, is not captured. Therefore, it is very hard to reconstruct facial details by fitting a 3DMM. They proposed to apply non-negative matrix factorisation (NMF) instead of PCA since it decomposes a shape into localised features, which improves the reconstruction of the details.

% Other techniques to build a 3D statistical facial mo\-del have been proposed, such as \citet{BruntonECCV2014}, \citet{LuthiPAMI2018} and \citet{LTranCVPR2018}. However, we only explain the construction of the 3DMM, since they are the most-widespread 3D facial models.

%%% Quitado el 22 feb
% Nevertheless, a common requirement to all the methods is the need of full (i.e. dense) point-to-point correspondence across all of the faces in the training set. This is not an easy task since any small inaccuracy in the correspondence may lead to a model of poor quality. There are two main ways of establishing this dense correspondence: directly in 3D, usually deforming a 3D template non-rigidly, or in a common 2D space where all the 3D faces are projected, usually a UV-space.

\subsection{Available 3D facial models} \label{subsec:3Dmodels}

In the last decades, several 3DMM have been built and made public. \citet{BlanzVetterSIGGRAPH1999} constructed a morphable model with 200 laser scans of heads of young adults (100 males and 100 females). They put the 3D faces of the training set in point-to-point correspondence using an optical flow algorithm based on the flattening of the 3D faces to a UV-space in 2D. Each 3D face is associated to a 2D cylindrical parametrisation by a bijective mapping. Notice that establishing a dense correspondence between two UV images implicitly establishes a 3D-to-3D dense correspondence (due to the bijection). \citet{PaysanAVSBS2009_BFM} constructed the well-known Basel Face Model (BFM) by applying the nonrigid iterative closest point (NICP) algorithm \citep{AmbergCVPR2007_NICP} to compute these dense correspondences directly between 3D faces. The BFM was built also with 100 female and 100 male subjects between 8 and 62 years old, with an average of 24.97 years old. One technical improvement of the BFM with respect to the 3DMM of Blanz and Vetter is the scanner, which is able to capture the facial geometry with higher resolution and precision in shorter time. The BFM was extended by \citet{GerigFG2018_BFM} who, instead of using NICP, established dense correspondence with a Gaussian process deformation model taking as mean deformation the zero function and as deformation basis the multi-scale B-spline kernel, introduced in \citet{OpferACM2006}.

% Gerig et al. [2018] establish dense correspondence with a Gaussian process deformation model with the spatially varying kernel.

\begin{figure}[ht]
    % \centering
    % \begin{subfigure}[b]{0.3\textwidth}
         \centering
         \includegraphics[scale = 0.6]{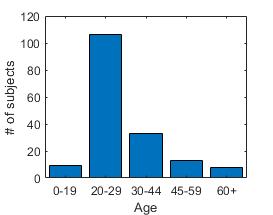}
         \caption{SFM dataset age distribution}\label{fig:SFM_age}
    %  \end{subfigure}
    % %  \hfill
    % \hspace{10em}
    %  \begin{subfigure}[b]{0.3\textwidth}
    %      \centering
    %      \includegraphics[scale = 0.65]{Figures/SFM_ethnicity.png}
    %      \caption{}\label{fig:SFM_ethnicity}
    %  \end{subfigure}
    % \caption{SFM dataset (a) age distribution, and (b) ethnicity distribution.}
    % \label{fig:SFM}
\end{figure}

\citet{HuberVISIGRAPH2016_SFM} presented the Surrey Face Model (SFM). It was built from 169 subjects, very diverse both in age (see Figure \ref{fig:SFM_age}) and in ethnicity (60\% Caucasian, 20\% Eastern Asian, 6\% Black African and 14\% of other ethnicities comprising South Asian, Arabic, and Latin). The scans were put in dense correspondence using the iterative multi-resolution dense 3D registration method \citep{TenaAVSS2006_IMDR}, which registers two 3D faces in three stages (global mapping, local matching, and energy minimisation) in a coarse-to-fine manner. To build the texture model, the 2D images captured by the camera system that generated the 3D scans were mapped to the registered 3D faces, obtaining in this way textured 3D faces. These textures were mapped back to 2D through a mapping that preserves the geodesic distances. The texture model was built from these resulting 2D representations.

\begin{table*}[ht]
\begin{threeparttable}[b]
\caption{Characteristics of the most popular available 3DMMs.}
\renewcommand{\arraystretch}{1.5}
\renewcommand\tabularxcolumn[1]{m{#1}}
\begin{tabularx}{\textwidth}{m{10em}YYcYYY}
\hline 
\textbf{3DMM} & \textbf{\# Cites}\tnote{1} & \textbf{\# Subjects} & \textbf{\% Males} & \begin{tabular}[c]{@{}c@{}}\textbf{Age}\\\textbf{(years old)}\end{tabular} & \textbf{Ethnicity} & \textbf{Expression} \\ \hline

BlanzVetter 1999 \citep{BlanzVetterSIGGRAPH1999} & 4824 & 200 & 50 & ``young adults'' & $-$ & $\times$ \\ \hline

BFM 2009 \citep{PaysanAVSBS2009_BFM} & 840 & 200 & 50 & 8-62 & ``most Europeans'' & $\times$\\ \hline

FaceWarehouse \citep{CaoVCG2014_FaceWarehouseFM} & 612 & 150 & $-$ & 7-80 & ``various'' & $\checkmark$\\ \hline

SFM \citep{HuberVISIGRAPH2016_SFM} & 165 & 169 & $-$ & Fig. \ref{fig:SFM_age}& 60\% Caucasian & $\times$  \\ \hline

CoMA \citep{RanjanECCV2018} & 159 & 20466 meshes from 12 subjects & $-$ & $-$ & $-$ & $\checkmark$\\ \hline

FLAME \citep{TLiToG2017_FLAME} & 156 & 3800 & 48 & ``wide range'' & ``wide range'' & $\checkmark$\\ \hline

LSFM \citep{BoothIJCV2017_LSFM} & 126 & 9663 & 48 & Fig. \ref{fig:LSFM_age} & 82\% White & $\times$ \\ \hline

BFM 2017 \citep{GerigFG2018_BFM} & 85 & 200 & 50 & $-$ & $-$ & $\checkmark$\\ \hline

LYHM \citep{DaiICCV2017_LYHM,DaiIJCV2020_LYHM} & 67 & 1212 & 50 & Fig. \ref{fig:LYHM_age} & $-$ & $\times$ \\ \hline

Multilinear Wavelet model \citep{BruntonECCV2014} & 59 & 99 & $-$ & $-$ & $-$ & $\checkmark$\\ \hline
\end{tabularx}

\begin{tablenotes}
\vspace{0.5em}
\item [1] \footnotesize{Number of cites extracted from Google Scholar on 22th February 2021.}
\end{tablenotes}
\end{threeparttable}
\end{table*}

% number of cites (according to google scholar)
% BlanzVetter 1999 \citep{BlanzVetterSIGGRAPH1999}      4824
% BFM 2009\citep{PaysanAVSBS2009_BFM}                   842
% BFM 2017\citep{GerigFG2018_BFM}                       85
% SFM \citep{HuberVISIGRAPH2016_SFM}                    165
% LSFM \citep{BoothIJCV2017_LSFM}                       126
% LYHM \citep{DaiICCV2017_LYHM,DaiIJCV2020_LYHM}        67
% FaceWarehouse \citep{CaoVCG2014_FaceWarehouseFM}      612
% Multilinear Wavelet model \citep{BruntonECCV2014}     59
% FLAME \citep{TLiToG2017_FLAME}                        156
% CoMA \citep{RanjanECCV2018}                           159

\begin{figure}[ht]
    % \centering
    % \begin{subfigure}[b]{0.3\textwidth}
         \centering
         \includegraphics[width = 0.95\columnwidth]{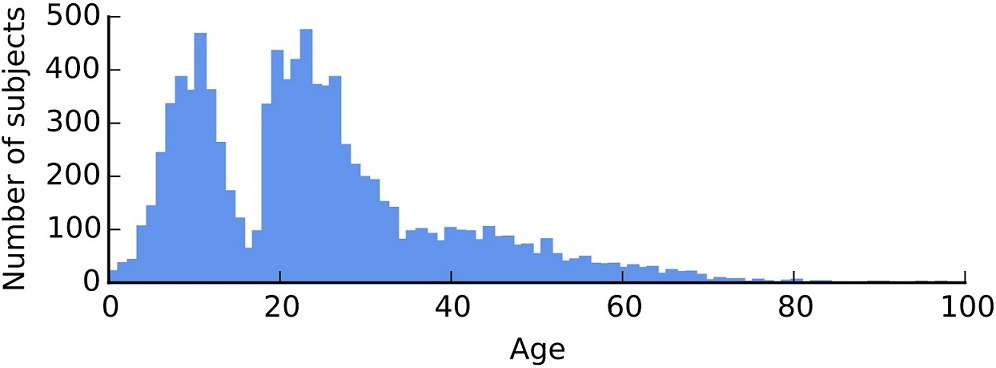}
         \caption{LSFM dataset age distribution (extracted from \citep{BoothIJCV2017_LSFM}).}\label{fig:LSFM_age}
    %  \end{subfigure}
    %  \hfill
    % \hspace{10em}
    %  \begin{subfigure}[b]{0.3\textwidth}
    %      \centering
    %      \includegraphics[scale = 0.65]{Figures/LSFM_ethnicity.png}
    %      \caption{}\label{fig:LSFM_ethnicity}
    %  \end{subfigure}
    % \caption{LSFM dataset age distribution (extracted from \citet{BoothIJCV2017_LSFM}) (left) and ethnicity distribution (right).}
    % \label{fig:LSFM}
\end{figure}

\citet{BoothIJCV2017_LSFM} and \citet{DaiICCV2017_LYHM,DaiIJCV2020_LYHM} proposed fully automated pipelines to construct a 3DMM, both consisting mainly of an automatic detection of landmarks, the computation of a shared triangulation across all the meshes and the model construction. To locate a set of 3D landmarks in the facial meshes, \citet{BoothIJCV2017_LSFM} detected 2D landmarks on multiple 2D renderings of each mesh using a state-of-the-art landmark detector. These 2D landmarks were mapped to the 3D face by inverting the rendering. Then, dense correspondences were established by deforming a predefined template mesh to fit the facial shapes using the NICP algorithm. From an initial PCA model of all fittings, erroneous correspondences can be identified by the corresponding shape vectors that behave as outliers. The final model was then obtained by applying PCA on the training set after excluding the outliers. With this pipeline, they presented the largest 3DMM until now, the Large Scale Facial Model (LSFM), built from 9,663 individuals covering a wide variety of ages (Figure \ref{fig:LSFM_age}), gender (48\% male), and ethnicity (82\% White, 9\% Asian, 5\% mixed heritage, 3\% Black and 1\% other ethnicities). The size of this dataset allowed the construction of smaller models from shapes of a specific age range and ethnicity.

\begin{figure}[h]
    \centering
    \includegraphics[width = 0.8\columnwidth]{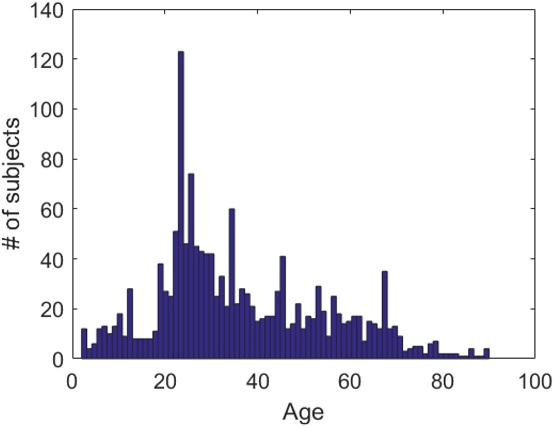}
    \caption{LYHM dataset age distribution (extracted from \citet{DaiICCV2017_LYHM}).}
    \label{fig:LYHM_age}
\end{figure}

The pipeline presented by \citet{DaiICCV2017_LYHM,DaiIJCV2020_LYHM} is similar to the one in \citep{BoothIJCV2017_LSFM}; however, rather than rendering images from the 3D scans to detect landmarks in 2D, \citet{DaiICCV2017_LYHM,DaiIJCV2020_LYHM} used the 2D image captured by the image system, like \citet{HuberVISIGRAPH2016_SFM}. The dense correspondences were also established by deforming a 3D facial template to each facial shape, but instead of using NICP, they used an approach based on the coherent point drift algorithm \citep{MyronenkoPAMI2010CoherentPointDrift} and refined the correspondences using optical flow for the texture channel. In their subsequent work \citep{DaiIJCV2020_LYHM}, before the dense morphing of the template, they first personalised the input template to better align with the scan, allowing them to obtain better correspondences. Finally, the facial model was built from the meshes in dense correspondence using PCA. They built a 3D model of the whole head, the Liverpool-York Head Model (LYHM), with 1,212 3D scans of subjects from a wide range of ages (Figure \ref{fig:LYHM_age}) and balanced in gender.

%% Build a 3D-tensor - do SVD - build bilinear model

In contrast to all the previous models, which capture geometric variations due to shape only, other works \citep{BruntonECCV2014,CaoVCG2014_FaceWarehouseFM,TLiToG2017_FLAME,RanjanECCV2018} modelled the facial deformations related to expression variations separately from identity. \citet{CaoVCG2014_FaceWarehouseFM} constructed the FaceWarehouse model from depth maps of 20 expressions of 150 individuals aged between 7 and 80 years old. In order to obtain meshes in dense correspondence, the Blanz and Vetter face model \citep{BlanzVetterSIGGRAPH1999} was fitted to the depth maps. Then, a bilinear face model was built by applying higher-order singular value decomposition (HOSVD) to the 3-rank data tensor (vertices $\times$ identities $\times$ expressions) constructed from the vectorised meshes. A very similar approach was proposed by \citet{BruntonECCV2014}, who also constructed a bilinear facial model by applying HOSVD to separate identity from expression facial variations, and using a training set of facial scans from 99 subjects with 25 expressions each. However, \citet{BruntonECCV2014} applied HOSVD to the wavelet coefficients extracted from the training shapes, instead of directly to the training shapes. The wavelet transform decomposes the facial surfaces in a multi-scale manner, allowing them to model coarse-scale shape variations separately from localised fine details. Dense correspondences were established in the training set by registering a 3D facial template using \cite{SalazarMVA2014}.

In contrast, \citet{TLiToG2017_FLAME} proposed a pipeline that jointly builds the facial model and refines the registration of the template by fitting it to the training scans. Then, they applied PCA to the meshes in dense correspondence with neutral expression to build the shape model. The expression model was constructed by applying PCA to the expression deformation fields obtained by removing the neutral face mesh from the expressive faces.

Differently from all the other works, \citet{RanjanECCV2018} used deep learning to build a non-linear facial model. They trained an autoencoder architecture to learn a latent space from the 3D facial scans by forcing the network to reconstruct the original mesh from the estimated latent representation. They captured 3D facial scans from 12 individuals for a range of 12 facial expressions, which in total makes a training set of $20,446$ meshes.

% In contrast to all the previous models, which capture geometric variations due to shape only, \citet{CaoVCG2014_FaceWarehouseFM} constructed the FaceWarehouse model, a bilinear face model that includes also expression variations. They used a Kinect system to capture 20 expressions (one of them being the neutral) of 150 individuals aged between 7 and 80 years old. For each of the expressions, 74 landmarks were located on the 2D picture using active shape model \citep{CootesCVIU1995_ASM}. These landmarks were then used to guide the fitting process of the Blanz and Vetter face model \citep{BlanzVetterSIGGRAPH1999} to the depth map, in order to obtain meshes in correspondence. Individual-specific expression blendshapes were then generated from the meshes obtaining 47 expressions for each subject, which were used as the face database. To build the bilinear face model, they assembled all the vectorised meshes into a 3-rank data tensor (vertices $\times$ identities $\times$ expressions) and applied N-mode singular value decomposition to decompose it.

\section{Statistical Model Fitting Methods} \label{sec:ModelFittingMethods}
% ////////////////////////////////////////////////////////////////////////////////////////
%                                      MINI CLUSTERS
% ////////////////////////////////////////////////////////////////////////////////////////
% \begin{itemize}
%     \item \textbf{NMF (instead of PCA)}: \cite{HJinCAGD2017}
%     \item \textbf{Cascaded regression with local features}: \cite{HuberICIP2015}
%     \item \textbf{Probabilistic approach}: \cite{SchonbornGCPR2013}
%     \item \st{\textbf{Linear approach}:}
%     \begin{itemize}
%         \item \st{Linear approach on camera, shape and texture:} \cite{AldrianSmithBMVC2010,AldrianSmithBMVC2011,AldrianSmithICCV2011,AldrianSmithPAMI2013,DingTheVisualComputer2014}
%         \item \st{Linear on camera, shape + non-linear refinement: }\cite{BasACCV2016}
%     \end{itemize}
%     \item \st{\textbf{Non-linear cost function minimisation}:}
%     \begin{itemize}
%         \item \st{By stages: \textit{Global+Detail}: }\cite{GarridoTOG2016,HJinCAGD2017}\st{ // \textit{Geometry} \textit{+Photometry}: }\cite{GHuPR2017}.
%         \item \st{Directly: \textit{Local approach}: }\cite{PiotraschkeCVPR2016}\st{ // \textit{Feature-based texture model}:} \cite{BoothCVPR2017,BoothPAMI2018}
%     \end{itemize}
%     \item \textbf{Local approach}: \cite{PiotraschkeCVPR2016,DingTheVisualComputer2014}
%     \item \textbf{Multiple images/video}: \cite{BoothPAMI2018,PiotraschkeCVPR2016,GarridoTOG2016,HJinCAGD2017}
% \end{itemize}

Using a statistical model to encode the prior knowledge of the 3D facial structure allows for a reconstruction of a new 3D face from one or more photographs by finding the linear combination of the model bases that best fits to the given 2D image(s). Essentially, fitting a 3D facial model to 2D images implies optimising a non-linear cost function, although other approaches have also been explored, such as the linearisation of the cost function or a probabilistic formulation.

Reconstructing the 3D facial geometry and appearance of a person from a single in-the-wild picture is much more challenging than using multiple images. This is why most works have mainly focused on 3D face reconstruction from a single image. Even so, some researchers have considered multiple images to improve the reconstruction accuracy by observing the subject's face under different poses and illumination conditions. In the same line, others have proposed to use 2D video sequences as a simple way of obtaining a set of pictures from the same person. However, when using a video, the temporal relation between frames has to be taken into account.

In this section, we summarise and compare the most relevant proposed approaches to fit a 3D facial model to one or more 2D images presented in the last decade. We review the three different approaches that we have identified among the 3DMM fitting works: non-linear optimisation of a cost function (Section \ref{subsec:MF_NonLinearCost}), linear approaches (Section \ref{subsec:LinearApproach}), and probabilistic approaches (Section \ref{subsec:ProbApproach}). Finally, in Section \ref{subsec:localApproach}, we explain how some researchers have proposed to fit a 3DMM in a local manner by dividing the face into subregions. The reviewed works are compiled in Tables \ref{tab:ModelFitting1} and \ref{tab:ModelFitting2}, and main conclusions are summarised in Section \ref{subsec:ModelFitting_conclusions}.

% First, we overview the different types of input data that have been used, analysing their strengths and weaknesses (Section \ref{subsec:MF_input}). Then,

% ////////////////////////////////////////////////////////////////////////////////////////
\begin{table*}[htp]
\caption{3D-from-2D face reconstruction approaches based on 3DMM fitting - part I.} \label{tab:ModelFitting1}
\begin{tabularx}{\textwidth}{X|m{1.3cm}|m{2.6cm}|m{1.8cm}m{1.8cm}m{1.8cm}m{1.8cm}}
\hline
\multirow{2}{*}{Reference} & \multirow{2}{*}{Images} & 
\multirow{2}{*}{\begin{tabular}[m{2.2cm}]{@{}l@{}}Features\\ used to fit\end{tabular}} & \multicolumn{4}{l}{Fitting}\\ \cline{4-7} 
& & & Camera parameters  & Shape & Texture & Illumination model \\
\hline
\citet{AldrianSmithBMVC2010} & Single & Landmarks & Linear & Maximise posterior & Colour channels ratios & Lambertian \\ \hline
\citet{AldrianSmithBMVC2011} & Single & Landmarks & Linear  & Maximise posterior & Minimise error & Lambertian + specular \\ \hline
\citet{AldrianSmithICCVW2011} & Single & Landmarks & Linear  & Maximise posterior & Minimise error & Lambertian + specular \\ \hline
\citet{AldrianSmithPAMI2013} & Single & Landmarks & Linear & Maximise posterior & Minimise error & Lambertian + specular \\ \hline
\citet{SchonbornGCPR2013} & Single & Render image, landmarks & Minimise error & Maximise posterior & Maximise posterior & Spherical harmonics \\ \hline
\citet{FShiTOG2014} & Video & Landmarks & Minimise error & Minimise error & $-$ &$-$ \\ \hline
\citet{DingTheVisualComputer2014} & Single & Landmarks & Linear & Maximise posterior & $-$ & $-$\\\hline
\citet{QuAVSS2014} & Multiple & Landmarks & Minimise error & Minimise error & $-$ & $-$\\ \hline
\citet{QuBMVC2015} & Single & Landmarks & Minimise error & Minimise error & $-$ & $-$ \\ \hline
\citet{HuberICIP2015} & Single & Features around landmarks & Cascaded regression & Cascaded regression & $-$ & $-$\\ \hline
\citet{XZhuCVPR2015} & Single & Landmarks & Minimise error & Minimise error & $-$ & $-$ \\ \hline
\citet{XZhuFG2015} & Single & Features around landmarks & Cascaded regression & Cascaded regression & $-$ & $-$ \\ \hline
\citet{BasACCV2016} & Single & \begin{tabular}[l]{@{}l@{}}Landmarks,\\texture edges\end{tabular} & Minimise error & Minimise error & $-$ & $-$ \\ \hline
\citet{PiotraschkeCVPR2016} & Multiple & Render image, landmarks & Minimise error & Minimise error & Minimise error & Phong\\\hline
\citet{GarridoTOG2016} & Video & Render image, landmarks & Minimise error & Minimise error & Minimise error & Lambertian \\ \hline
\citet{ThiesCVPR2016} & Single, video & Render image, landmarks & Minimise error & Minimise error & Minimise error & Lambertian \\ \hline
\citet{GHuPR2017} & Single & Render image, landmarks & Minimise error & Minimise error & Minimise error & Phong \\ \hline
\citet{HernandezComGraph2017} & Video & Image features around projected 3D vertices & Minimise error & Minimise error & $-$ & $-$ \\ \hline
\citet{BoothCVPR2017} & Single & Render image, landmarks & Minimise error & Minimise error & Minimise error & \begin{tabular}[l]{@{}l@{}}Image\\features\end{tabular} \\\hline
\citet{HJinCAGD2017} & Two & Render image, landmarks & Regression model & Minimise error & Minimise error & Phong \\ \hline
\citet{BoothPAMI2018} & Single, video & Render image, landmarks & Minimise error & Minimise error & Minimise error & \begin{tabular}[l]{@{}l@{}}Image\\features\end{tabular} \\ \hline
\end{tabularx}
\end{table*}

\begin{table*}[htp]
\caption{3D-from-2D face reconstruction approaches based on 3DMM fitting - part II.} \label{tab:ModelFitting2}
\begin{tabularx}{\textwidth}{X|m{1.3cm}|m{2.6cm}|m{1.8cm}m{1.8cm}m{1.8cm}m{1.8cm}}
\hline
\multirow{2}{*}{Reference} & \multirow{2}{*}{Images} & 
\multirow{2}{*}{\begin{tabular}[m{2.2cm}]{@{}l@{}}Features\\ used to fit\end{tabular}} & \multicolumn{4}{l}{Fitting}\\ \cline{4-7} 
& & & Camera parameters  & Shape & Texture & Illumination model \\
\hline
\citet{LJiangTIP2018} & Single & Render image, landmarks & Minimise error & Minimise error & Minimise error & Lambertian \\ \hline
\citet{GecerCVPR2019} & Single & Render image, landmarks \& image features & Minimise error & Minimise error & Minimise error & Phong \\ \hline
\citet{PLiuMIPR2019} & Single & Landmarks & Minimise error & Minimise error & $-$ & $-$ \\ \hline
\citet{SariyanidiECCV2020} & Mutliple & Render image, landmarks & Minimise error & Minimise error & Minimise error & Phong \\ \hline
\citet{KoujanCVMP2018} & Mutliple & Subset of 3D vertices & Minimise error & Minimise error & $-$ & $-$ \\ \hline
\citet{FerrariToM2017} & \color{darkblue}Single & Landmarks & Linear & Minimise error & $-$ & $-$ \\ \hline
\end{tabularx}
\end{table*}

\subsection{Optimisation of a non-linear cost function} \label{subsec:MF_NonLinearCost}

\citet{BlanzVetterSIGGRAPH1999} not only introduced the 3DMMs to the community but also proposed a method to fit them to a single facial image, indicating a way of extending it to many images, that inspired many others. Their fitting procedure was based on a simple idea: if the input image and the image rendered from the fitted 3D face are similar, the 3D reconstruction is faithful to the input image. Specifically, they optimised the shape $\vec{\alpha}$ and the albedo $\vec{\beta}$ parameters alongside with a set of rendering parameters $\vec{\rho}$ (including intrinsic camera parameters, projection parameters as well as illumination parameters) such that they produced an image $\Imod$ as close as possible to the input $\Iin$; i.e., they minimised the sum of the Euclidean distance between each pixel $(x,y)$ of the input and the reconstructed images:

\begin{equation} \label{eq:errorI_BV1999}
    \errorI(\vec{\alpha}, \vec{\beta}, \vec{\rho}) = \sum_{x,y} \norm{ \Iin(x,y) - \Imod(x,y)}_2^2.
\end{equation}
$\Imod$ was rendered using the Phong reflectance model \citep{Phong1975} and a perspective projection defined by $\mathcal{P}_{\text{PP}}: \mathbb{R}^3 \to \mathbb{R}^2$
\begin{equation}\label{eq:perspectiveProj}
\begin{array}{l}
    \mathcal{P}_{\text{PP}}(\vec{v},\vec{\zeta}) = \frac{f}{v'_z} \begin{pmatrix}v'_x\\v'_y\end{pmatrix} + \begin{pmatrix}o_x\\o_y\end{pmatrix}\\
    \text{where}     \begin{pmatrix}v'_x\\v'_y\\v'_z\end{pmatrix} = \vec{R}\vec{v} + \vec{\tau}
\end{array}
\end{equation}
for a point $\vec{v} \in \mathbb{R}^3$. The projection parameters $\vec{\zeta} = \{ \vec{R}, \vec{\tau}, f, \vec{o}\}$ include the rotation matrix $\vec{R}$ and the translation $\vec{\tau}$ of the face, the focal length $f$ and the coordinates of the optical axis in the image plane $\vec{o} = (o_x,o_y)^{\text{T}}$.

However, only by minimising the distance between the input and the rendered images, a face-like surface is not guaranteed since the minimisation is ill-posed. To solve this issue, Blanz and Vetter added to the cost function a regularisation term for the parameters using the prior probabilities of the model coefficients (eq. \eqref{eq:prior_prob}) and an ad-hoc prior for the rendering coefficients:

\begin{equation*}\label{eq:errorReg_BV1999}
    \errorReg(\vec{\alpha}, \vec{\beta}, \vec{\rho}) = \sum_{i = 1}^{\widetilde{M}_\vec{x}} \frac{\alpha_i^2}{\sigma_{\alpha_i}^2} + \sum_{i = 1}^{\widetilde{M}_\vec{c}} \frac{\beta_i^2}{\sigma_{\beta_i}^2} + \sum_{j} \frac{(\rho_j - \bar{\rho}_j)^2}{\sigma_{\rho_j}^2}.
\end{equation*}

Later, in \citep{BlanzVetterPAMI2003}, they extended their work by adding another term that enforces a fixed set of facial feature points (landmarks) in the reconstructed facial shape, $\{\Lmk_i\}_{i=1}^L$, to be such that their projections, $\LmkProj_i$, lie over the corresponding set of manually annotated landmarks in the input image, $\lmk_i$,

\begin{equation}\label{eq:errorlmk_BV2003}
    \errorlmk(\vec{\zeta}) = \sum_{i = 1}^{L}  \norm{\lmk_i - \LmkProj_i}_2^2.
\end{equation}

This 3DMM fitting approach based on the minimisation of the difference between the input and the rendered image has been followed by many others \citep{FShiTOG2014,QuAVSS2014,QuBMVC2015,XZhuCVPR2015,PiotraschkeCVPR2016,HJinCAGD2017,ThiesCVPR2016,BasACCV2016,GarridoTOG2016,HernandezComGraph2017,BoothCVPR2017,BoothPAMI2018,LJiangTIP2018,PLiuMIPR2019,SariyanidiECCV2020}. In particular, \citet{PiotraschkeCVPR2016} used the method proposed by \citep{BlanzVetterPAMI2003} to obtain single-image reconstructions from several pictures of the same person. Then, they combined them to obtain a single 3D face. \citet{ThiesCVPR2016} also used \citep{BlanzVetterPAMI2003} but aiming at face reenactment. They fitted a 3DMM to the source and the target images and then transferred the estimated expression from the source image to the target image.

In contrast, other works \citep{HernandezComGraph2017,QuBMVC2015,PLiuMIPR2019,XZhuCVPR2015} used only one of the error terms proposed by \citep{BlanzVetterSIGGRAPH1999}, either the image error (eq. \eqref{eq:errorI_BV1999}) or the landmarks error (eq. \eqref{eq:errorlmk_BV2003}). On the one hand, \citet{HernandezComGraph2017} proposed a multi-view 3D face reconstruction method focusing on ensuring photometric consistency between the input and the rendered images, but also taking advantage of the input video sequence by imposing photometric consistency between consecutive frames. Specifically, they minimised a local representation $\mathbf{F}$ around the projection over consecutive frames, $t$ and $t+1$, of the 3D vertices, $i = 1,\cdots,N$ of the estimated 3D faces for each frame, $\vec{x}^t = \vec{x}(\vec{\alpha}^t)$ and $\vec{x}^{t+1} = \vec{x}(\vec{\alpha}^{t+1})$:
\begin{equation*}
    \mathbb{E} = \sum_t \sum_{i=1}^N \norm{\mathbf{F}(\mathcal{P}(\vec{x}_i^t,\vec{\zeta}^{t})) - \mathbf{F}(\mathcal{P}(\vec{x}_i^{t+1},\vec{\zeta}^{t+1}))}_2^2,
\end{equation*}
where $\vec{x}_i^t$ represents the $i$th vertex of $\vec{x}^t$, $\mathcal{P}(\vec{x}^t_i, \vec{\zeta}^t)$ is the projection of the vertex $\vec{x}_i^t$ according to the projection parameters of frame $t$, $\vec{\zeta}^t$, and $\mathbf{F}$ is the local feature extraction function (image colour, image intensity, complex image features, etc.). On the other hand, \citep{QuBMVC2015,PLiuMIPR2019,XZhuCVPR2015} used only the landmark term (eq. \eqref{eq:errorlmk_BV2003}), including the contour landmarks, so as to further constrain the 3D face reconstruction problem. Specifically, \citet{QuBMVC2015} separated the in-face and the contour landmarks in two terms. Since the contour landmarks are not well defined in non-frontal poses, treating contour landmarks separately allowed them to define the 2D contour landmark in a softer manner: a 2D contour landmark corresponding to a 3D contour landmark is the nearest landmark in the image to the projection of the 3D landmark. Similarly, \citet{PLiuMIPR2019} iteratively updated the 2D contour landmarks while estimating the shape and projection parameters. This update consisted in taking as new landmarks the nearest points on the 2D contour line to the projected 3D contour landmarks. In this way, the contour landmarks' correspondence is improved, helping to better estimate the 3D face. Contrary to \citep{QuBMVC2015,PLiuMIPR2019}, \citet{XZhuCVPR2015} jointly estimated the 3DMM parameters, the projection parameters, and the position of 3D contour landmarks on the 3DMM given the 2D landmarks on the image. They observed that the pose parameters and the contour landmarks depended on each other; thus, they proposed to iteratively estimate first the shape and pose parameters and then update the position of the contour landmarks.

\citet{KoujanCVMP2018} also used only a set of corresponding 3D-2D points to fit the 3DMM to images but, contrary to \citep{QuBMVC2015,PLiuMIPR2019,XZhuCVPR2015}, they established dense correspondence between the input images and the 3D face. They used a video sequence to estimate the 3D facial geometry of a person by first computing optical flow from a reference frame to all the rest, and then finding correspondences between the set of moving 2D points identified by the optical flow algorithm and the 3D vertices of the facial shape, which was previously initialised using only landmarks. Also, instead of incorporating the 3DMM as a hard constraint by estimating its parameters, they penalised solutions that deviate much from the 3DMM space. This soft constraint allowed them to obtain reconstructions that capture details that cannot be represented by the face model.

\subsubsection{Redefining the image error}
Differently from the works above, which measured the Euclidean distance between the input and the rendered images as in eq. \eqref{eq:errorI_BV1999}, \citep{SariyanidiECCV2020,BoothCVPR2017,BoothPAMI2018,GecerCVPR2019} completely redefined the image error: whereas the former \citep{SariyanidiECCV2020} used the gradient of the images to compute the image error, \citep{BoothCVPR2017,BoothPAMI2018,GecerCVPR2019} measured the difference between the images in a feature space. In particular, \citet{SariyanidiECCV2020} computed the gradient correlation \cite{TzimiropoulosICCV2011_GradCorr} between the input and the rendered images, $\errorI = {\nabla \Iin}^\text{T} \ \nabla \Imod$, which they argued is more robust against illumination variations and occlusions. However, their main contribution is a regularisation of the model parameters based on adding inequality constraints to the minimisation problem, whose upper and lower bounds are extracted from the 3DMM. The authors highlighted that adding a regularisation term to the cost function (eq. \eqref{eq:errorReg_BV1999}) may lead to unsatisfying results, either because of an oversmoothed 3D face or the opposite, since the weight of the regularisation on the minimisation problem is controlled by a parameter that is chosen ad-hoc.
%  The gradients $\nabla \mathbf{I}_{*}$ are computed as the concatenation of the mag\-ni\-tu\-de-normalised differences along the vertical and horizontal axes for each of the $K$ pixels.

On the other hand, \citet{BoothCVPR2017,BoothPAMI2018} constructed a texture model using a dense feature-based representation, instead of using the per-vertex RGB values of the 3D face. The training samples used to build the texture model (eq. \eqref{eq:textureModel}) were image feature vectors, $\vec{f}_i$, that were obtained by applying a dense feature extraction function to each of the images in the training set. Then, assuming known shape parameters $\vec{\alpha}$ (thus, known 3D facial shape, $\vec{x}(\vec{\alpha})$) and projection parameters $\vec{\zeta}$ for each training image, they computed the projection of the 3D faces $\vec{x}(\vec{\alpha})$ according to $\vec{\zeta}$ into the image plane $\mathcal{P}\left(\vec{x}(\vec{\alpha})),\vec{\zeta}\right)$, and sampled the feature vectors on the location of the projection of the 3D faces. Therefore, for each training image, they obtained a vector $\vec{f}_i$ composed of the feature vectors on the location of the corresponding projected 3D face, and the image error $\errorI$ was redefined as

\begin{equation}\label{eq:errorI_BoothCVPR2017}
    \errorI(\vec{\alpha},\vec{\beta},\vec{\zeta}) = \norm{\mathbf{F}(\mathcal{P}(\vec{x}(\vec{\alpha}),\vec{\zeta})) - \vec{f}(\vec{\beta})}_2^2,
\end{equation}
where $\mathbf{F}$ is the function that extracts the features from the pixels in the input image corresponding to the location of the projected reconstructed 3D face $\mathcal{P}(\vec{x}(\vec{\alpha}),\vec{\zeta})$, and $\vec{f}(\vec{\beta})$ is the instance of the feature-based texture model corresponding to the parameters $\vec{\beta}$.

With a similar idea, \citet{GecerCVPR2019} extracted image features from $\Iin$ and $\Imod$ using the ArcFace network \citep{DengCVPR2019_arcface}, which is a network trained for face recognition. Therefore, by minimising the cosine distance between the feature vectors and the Euclidean distance between features from intermediate layers, the authors forced the estimated 3D face to have the same identity as the input image, thus obtaining more faithful reconstructions. In addition, and differently from the other works, \citet{GecerCVPR2019} trained a generative adversarial network (Section \ref{subsubsec:DL_GANs}) to estimate a refined texture UV-map from the texture parameters estimated by minimising the cost function.

\subsubsection{Multi-stage optimisation}
The works studied above \citep{BoothCVPR2017,BoothPAMI2018,HernandezComGraph2017,PiotraschkeCVPR2016,QuBMVC2015,ThiesCVPR2016,GecerCVPR2019} jointly optimised the shape, texture, and projection parameters by minimising a single cost function. However, other approaches \citep{GarridoTOG2016,HJinCAGD2017,QuAVSS2014,FShiTOG2014,BasACCV2016,XZhuFG2015,HuberICIP2015,LJiangTIP2018} proposed to reconstruct the 3D facial shape by minimising a cost function in different stages. Some researchers \citep{GarridoTOG2016,HJinCAGD2017,QuAVSS2014,FShiTOG2014,BasACCV2016} proposed to reconstruct the 3D face in stages where each progressively added details to refine the previous one, others \citep{HuberICIP2015,XZhuFG2015} trained a cascaded regressor, and \citep{LJiangTIP2018} split the 3D facial reconstruction process into the estimation of the facial geometry of the subject and the estimation of the appearance, including the illumination.

In the coarse-to-fine category, \citet{QuAVSS2014} refined the single-image reconstruction obtained by minimising the landmarks error (eq. \eqref{eq:errorlmk_BV2003}) using multiple images. In contrast, \citep{HJinCAGD2017,FShiTOG2014,GarridoTOG2016} divided the fitting process into a global and a fine-scale reconstruction stages. \citet{HJinCAGD2017} and \citet{FShiTOG2014} obtained the coarse 3D face by minimising only the landmarks error (eq. \eqref{eq:errorlmk_BV2003}). Then, they departed from the 3DMM and iteratively recovered facial details by refining the surface normals using the image error (eq. \eqref{eq:errorI_BV1999}) and adding a term that enforced the refined shape to be similar to the coarse one. The refinement of the surface normals allowed them to obtain detailed reconstructions with notable accuracy. Although both works \citep{HJinCAGD2017,FShiTOG2014} followed very similar approaches, the main difference lies on the type of model used: whereas \citet{FShiTOG2014} used a PCA-based model, \citet{HJinCAGD2017} used a facial model based on non-negative matrix factorisation, which, according to the authors, allows for a local decomposition of the shape variations (see Section \ref{subsec:Build3Dmodel}). On the other hand, \citet{GarridoTOG2016} proposed a 3-stage reconstruction method, where the coarse shape was obtained by minimising the image error and the landmark error. Then, medium-scale corrections were estimated as a 3D deformation field modelled by manifold harmonics \citep{ValletCGF2008}, whereas the fine-scale deformation field was estimated by inverse rendering optimisation such that the synthesised shading gradients matched the gradients of the illumination in the corresponding input image as good as possible. With this approach, \citet{GarridoTOG2016} obtained highly detailed 3D faces, improving the results from \citep{FShiTOG2014}.

%Whereas \citet{HJinCAGD2017} assumed a Phong reflectance model and used statistical regularisation of the model parameters, \citet{FShiTOG2014} adopted a Lambertian reflectance model and regularised the parameters by minimising the difference to reference parameters. 

Similarly to the coarse-to-fine scheme, \citet{HuberICIP2015} and \citet{XZhuFG2015} trained a cascaded regressor to jointly optimise the camera $\vec{\zeta}$ and the shape $\vec{\alpha}$ parameters given an initial feature vector. The feature vectors they used were SIFT features extracted from patches of the input image $\Iin$ around the projections of the 3D-landmarks into the image plane $\LmkProj$. Hence, at each step $t$, given the parameters $\vec{\theta}_t = \{\vec{\zeta}_t,\vec{\alpha}_t\}$ estimated by the previous weak regressor $\mathcal{R}_{t-1}$, the 3D-landmarks in $\vec{x}(\vec{\alpha}_t)$, $\{\Lmk_i(\vec{\alpha}_t)\}_{i=1}^L$, were projected into the input image plane $\mathcal{P}(\Lmk_i(\vec{\alpha}_t),\vec{\zeta}_t) = \LmkProj_i(\vec{\alpha}_t)$, and image features were extracted from patches of $\Iin$ centred at $\LmkProj_i(\vec{\alpha}_t)$, $\big\{\vec{f}_t^i = \vec{f}(\vec{\theta}_t, \LmkProj_i(\vec{\alpha}_t))\big\}_{i=1}^L$. Each weak regressor $\mathcal{R}_t$ inputed a vector of features $\vec{f}_t = [\vec{f}_t^1, \dots, \vec{f}_t^L]$ and outputed an optimal parameter update $\vec{\Delta} \vec{\theta}_t$ such that $\vec{\theta}_{t+1} = \vec{\theta}_t + \vec{\Delta} \vec{\theta}_t$.

Contrary to \citep{HJinCAGD2017,FShiTOG2014,GarridoTOG2016}, who split the reconstruction process into the estimation of a global facial shape and the recovery of geometric details, \citet{BasACCV2016} followed the coarse-to-fine strategy by first estimating the shape and projection parameters in a linear manner and then refining it with a non-linear cost function. In the linear stage, the shape $\vec{\alpha}$ and projection $\vec{\zeta} = \{\vec{R},\vec{\tau},s\}$ parameters are estimated by solving the linear system of equations
\begin{equation} \label{eq:projLmks}
\vec{p}_i = \mathcal{P}(\vec{v}_i,\vec{\zeta})    
\end{equation}
for a set of corresponding vertices in the 3D face $\vec{v}_i\in\mathbb{R}^3$ and pixels in the image $\vec{p}_i\in\mathbb{R}^2$, where $ \mathcal{P}(\vec{v},\vec{\zeta})$ is the weak-perspective projection of $\vec{v}\in\mathbb{R}^3$ into the image plane according to $\vec{\zeta}$. To further constrain the reconstruction, they computed correspondences between texture edges and the occluding boundary of the 3D face, defined as the set of vertices that lie on the mesh edge whose adjacent mesh faces have a change of visibility. In this way, the global structure of the face is recovered more accurately due to the dense correspondences obtained in the facial boundary. However, fine details are hardly captured since the landmarks used in the inner face constitute a sparse set, not being sufficient to capture subtle details.

As stated above, in contrast to the coarse-to-fine strategy, \citet{LJiangTIP2018} divided the 3D-from-2D face reconstruction problem into the estimation of the geometry and the estimation of the texture. They estimated the geometry by minimising the landmarks error (eq. \eqref{eq:errorlmk_BV2003}), which was used in the photometric stage to estimate the albedo and illumination parameters by minimising the image error $\errorI$. However, they took advantage of the estimated texture to refine the reconstruction resulting from the geometric stage by reestimating the surface normals. To further improve the reconstruction of fine geometric details, they reestimated the surface normals by minimising the difference in intensity gradients between the input and the rendered images.

\subsection{Linear approaches} \label{subsec:LinearApproach}

All the previously mentioned works exploit the optimisation of a non-linear error function, similarly to the approach presented originally by \citet{BlanzVetterSIGGRAPH1999,BlanzVetterPAMI2003}. However, \citet{AldrianSmithBMVC2010} observed that the optimisation problem proposed by \citet{BlanzVetterPAMI2003} is highly complex, computationally expensive and prone to be stuck on local minima due to its ill-posed nature. To avoid that, they proposed an alternative: to fit the 3DMM in a linear manner, separately for the projection parameters, the facial geometry and the texture. Unlike \citep{BasACCV2016}, who refined in a non-linear manner the reconstruction resulting from the linear stage, \citet{AldrianSmithBMVC2010} proposed a completely linear approach that was later further explored \citep{FerrariToM2017,AldrianSmithICCVW2011,AldrianSmithBMVC2011,AldrianSmithPAMI2013,DingTheVisualComputer2014}.

Specifically, \citet{AldrianSmithBMVC2010} estimated the projection parameters by solving the linear system of equations from eq. \eqref{eq:projLmks} for a set of corresponding 3D-2D landmarks, and assuming an affine camera model, which allowed them to rewrite eq. \eqref{eq:projLmks} into
\begin{equation*}
    \vec{p}_i = \vec{C}\vec{v}_i,
\end{equation*}
where $\vec{C}\in\mathbb{R}^{3\times4} $ is the camera projection matrix. Then, they estimated the shape model parameters by maximising the posterior probability over the shape parameters $\vec{\alpha}$ given the set of 2D-landmarks in the image $\vec{\mathcal{L}}^\text{2D}$, $p(\vec{\alpha}|\vec{\mathcal{L}}^\text{2D})$, which is differentiable and leads to a linear system of equations for $\vec{\alpha}$. Finally, to estimate the texture, they assumed a white illumination and a Lambertian reflectance model (diffuse-only reflectance). These assumptions allowed them to compute the texture by imposing the ratios between pairs of colour channels in the input image to be equal to the ratios of texture in the corresponding vertices of the 3D face. A very similar approach was followed by \citet{FerrariToM2017} but, instead of maximising the posterior probability like \citep{AldrianSmithBMVC2010}, they linearised the landmarks error (eq. \eqref{eq:errorlmk_BV2003}). In this way, they obtained a closed-form solution for the shape parameters, since the projection parameters were estimated in a previous step.

Whereas linearising the fitting process results on an increased computational efficiency, the resulting reconstruction are not very accurate and strong assumptions are required. Consequently, Aldrian and Smith, in their following works \citep{AldrianSmithICCVW2011,AldrianSmithBMVC2011,AldrianSmithPAMI2013}, imposed less restrictive assumptions on the illumination model of the face. Where\-as in \citep{AldrianSmithBMVC2010} they assumed a Lambertian reflectance model, thereafter they adopted a dichromatic reflectance model, which is a more realistic model since it comprises additive diffuse and specular terms. In \citep{AldrianSmithICCVW2011,AldrianSmithBMVC2011} they proposed different approaches to estimate the texture and illumination parameters, which are later summarised in \citep{AldrianSmithPAMI2013}.

In \citep{AldrianSmithICCVW2011}, they proposed to use a specular-invariant representation of the texture so that the diffuse re\-flec\-tan\-ce could be estimated independently from the specular reflectance. This specular-invariant representation is the SUV colour space \citep{ZicklerIJCV2008_SUV}, which is a rotation of the RGB colour space such that one of the axes is aligned with the direction of the colour of the light source. Therefore, the representation of the input image in the SUV colour space depends only on the diffuse term. Even though the image-formation model adopted in \citep{AldrianSmithICCVW2011} was more realistic than their earliest work \citep{AldrianSmithBMVC2010}, to be able to construct the specular invariant space, they assumed that all light sources had the same fixed known colour. This strong assumption was further relaxed in their latter work \citep{AldrianSmithBMVC2011} by considering an unconstrained illumination, and assuming non-specular reflectance to estimate the diffuse part.
%To estimate the specular reflectance, they computed the specular-only image as the difference between the input image and the diffuse-only image, which was estimated in the diffuse fitting stage.

\citet{DingTheVisualComputer2014} argued that the approach proposed by \citet{AldrianSmithBMVC2010,AldrianSmithICCVW2011,AldrianSmithBMVC2011,AldrianSmithPAMI2013} to linearly estimate the geometry has two main drawbacks: first, in the camera projection matrix estimation, they do not take into account the orthonormal nature of the rotation matrix, and second, estimating the camera and the shape parameters separately, in a two-step manner, does not guarantee to reduce the cost function in each step. To solve the first drawback, \citet{DingTheVisualComputer2014} proposed to rescale and orthogonalise the camera projection matrix resulting from Aldrian and Smith's algorithm to enforce orthonormality of the rotation matrix. The second disadvantage was solved by jointly estimating the camera projection matrix and the shape parameters.

Although \citet{DingTheVisualComputer2014} improved the results from \citep{AldrianSmithBMVC2010,AldrianSmithBMVC2011,AldrianSmithICCVW2011,AldrianSmithPAMI2013}, they all assumed an affine camera model, which cannot model perspective effects. This limitation was highlighted by \citet{GHuPR2017}, who adopted a perspective camera model to estimate the projection parameters. Also, differently from \citep{AldrianSmithBMVC2010,AldrianSmithBMVC2011,AldrianSmithICCVW2011,AldrianSmithPAMI2013,DingTheVisualComputer2014}, they split the reconstruction process into geometric and photometric stages, iterating them in turn a few times to refine the solution, and assumed a Phong's reflectance model, which is more realistic that the dichromatic model adopted by \citep{AldrianSmithBMVC2011,AldrianSmithICCVW2011,AldrianSmithPAMI2013}. This allowed them to obtain more accurate results.

%More precisely, the geometric stage consisted of three steps: first, only the projection parameters are estimated, then used to estimate the shape parameters in the second step, and in the last step the contour landmarks are estimated and added to the set of landmarks used in the previous steps to iteratively refine the solution. The photometric stage is also divided in three steps where the light direction, the light intensity and the texture parameters are estimated separately by minimising the image error (eq. \eqref{eq:errorI_BV1999}) while keeping fixed the rest of the variables.

\subsection{Probabilistic approach} \label{subsec:ProbApproach}
A very different approach for fitting a 3DMM was proposed by \citet{SchonbornGCPR2013} where they reformulated the process of fitting a 3DMM as a probabilistic inference problem. Their approach is based on drawing samples from the posterior distribution over the set of all the parameters $\vec{\theta}$ - containing the 3DMM’s parameters (shape $\vec{\alpha}$ and texture $\vec{\beta}$), the illumination and the projection parameters $\vec{\zeta}$ - given the input image $\Iin$, $p(\vec{\theta}| \Iin)$. Although the use of the posterior probability was also exploited by others \citep{AldrianSmithBMVC2010,AldrianSmithICCVW2011,AldrianSmithBMVC2011,AldrianSmithPAMI2013,DingTheVisualComputer2014}, \citet{SchonbornGCPR2013} reinterpreted the model fitting procedure. They used the Metropolis-Hastings algorithm to draw samples of the parameters $\vec{\theta}$  distributed according the posterior distribution $p(\vec{\theta}| \Iin)$ by stochastically accepting or rejecting samples. Such posterior distribution was computed by applying the Bayes' theorem, $p(\vec{\theta}| \Iin) \propto p(\Iin| \vec{\theta})p(\vec{\theta})$. The first term, $p(\Iin| \vec{\theta})$, is a Gaussian distribution where each pixel is treated independently, considering different distributions for foreground and background pixels. The background distribution is trained on all the background pixels of the input image and the foreground distribution is known from the 3DMM. In the second term, $p(\vec{\theta})$, they integrated all sources of information to estimate a distribution of the parameters. Basically, they observed that, given a completely automatic pipeline where a face detector and a landmark detector are needed; both detectors are forced to make an early decision that might be unreliable due to strong pose and illumination variations. Thus, they included the results of the face and landmark detectors as detection maps, assigning each pixel in the image the likelihood of having a face (or a specific landmark) in that position. In this way, for each detected face box$_i$, $p(\vec{\theta})$ is biased with the position and size of the box and with the landmark detection map of the box $\mathcal{L}_i$, $p(\vec{\theta} | \text{box}_i,\mathcal{L}_i)$. Then, all these candidate distributions (one per detected face) are combined constructing a global distribution $p(\vec{\theta}| \text{allboxes},\mathcal{L})$ by computing the mean of the face box-distributions, which includes the knowledge about all the detections.

\subsection{Local approaches} \label{subsec:localApproach}
Most of the proposed techniques to reconstruct the 3D face of a person approach the problem in a global manner, fitting the whole facial model to the image(s). However, \citet{DingTheVisualComputer2014} and \citet{PiotraschkeCVPR2016} addressed the fitting procedure in a local manner considering subregions of the face, although in slightly different ways. The former \citep{DingTheVisualComputer2014} reconstructed each of the subregions locally from the same input image by ensuring that each of them was optimally fitted, without taking into account the other regions. In contrast, \citet{PiotraschkeCVPR2016} reconstructed global 3D faces from different input images, and then fused the reconstructions with a criterion based on the accuracy of each of the subregions.

Specifically, \citet{DingTheVisualComputer2014} obtained $K$ shape parameter vectors $\vec{\alpha}_k$, each of them individually optimised for one of the $K$ subregions. Then, the shape parameter vectors were linearly combined with blending weights that depended on the vertices: $\vec{x} = \overline{\vec{x}} + \sum_{k=1}^K\vec{w}_k \left(\vec{\Phi}\vec{\alpha}_k\right)$, with $\sum_{k=1}^K w_{kj} = 1, \forall \text{ vertex }j$. In this way, they increased the flexibility of the 3DMM and ensured that each of the regions were optimally fitted, obtaining more accurate reconstructions.

On the other hand, as stated above, \citet{PiotraschkeCVPR2016} used multiple in-the-wild pictures of a person and reconstructed individual 3D sha\-pes from each of them using \citep{BlanzVetterPAMI2003}. Each of them was evaluated locally, obtaining a quality measure for each of the subregions of each 3D face. These measures were used to obtain a single reconstruction of each subregion by computing a weighted li\-ne\-ar combination of the best ones, which were finally merged into a single 3D face. Thus, the main contribution of \citep{PiotraschkeCVPR2016} was the pipeline that provides a more accurate 3D face by combining several sub-optimal 3D facial reconstructions based on their quality, which was computed without the need of the ground truth face.

\subsection{Take-home message} \label{subsec:ModelFitting_conclusions}

Fitting a 3DMM to 2D images basically consists on finding the linear combination of the model bases that produce a 3D face that best resembles the person image of the input picture(s). To do so, the community has used different input modalities and proposed different approaches. Whereas the most widespread input is a single image, which is the most challenging scenario, some researchers have fitted a 3DMM to a collection of images from the same person or even video sequences, which provide more information by observing the subject from different poses and illuminations. Also, most of the proposed works used a set of corresponding landmarks to drive the fitting process, but other features such as texture edges \citep{BasACCV2016} or complex image features \citep{HuberICIP2015} have also been adopted.

We have identified three main approaches to fit a 3DMM to images: the optimisation of a non-linear cost function, which is the most widespread one, the linearisation of the cost function, and a probabilistic formulation. Although all three approaches are different in the way they optimise the model parameters, the fitting process is guided by the same main elements: the landmarks or corresponding 2D-3D points, the rendered image, and statistical regularisation. The set of 2D-3D correspondences guides the fitting by imposing the known 2D location of the 3D points when projected to the image plane. The rendered image does so by imposing some similarity measure between the rendered appearance and the input images. And, finally, plausible reconstructions are ensured by imposing the statistical regularisation defined by the 3DMM.

Whereas the optimisation of a non-linear cost function allows incorporating many different (complex) terms easily, it may get stuck on local minima or even diverge, and it can be slower than desired due to its computational cost. In contrast, the linearisation of the cost function avoids most of these issues given its simpler nature, but it is harder to incorporate complex terms to the optimisation problem, and usually strong non-realistic assumptions are required. Finally, probabilistic approaches favour the modelling of uncertainties thanks to the probabilistic inference framework, however, the process of drawing samples from a posterior distribution may be very slow.

On the other hand, even though a wide range of different approaches have been proposed, all of them are limited by the kind of information included in the fitted 3DMM. For example, approaches based on generating a synthetic image cannot be used to fit a 3DMM that does not model appearance, such as the FaceWarehouse \citep{CaoVCG2014_FaceWarehouseFM}. In such cases, the community has exploited the information of additional geometric features, such as contour landmarks and occluding boundaries. Similarly, if the fitted 3DMM does not model expressions, the accuracy of the method may be reduced given that expression-related shape variations can be misinterpreted as identity-related variations.

\section{Photometric Methods} \label{sec:Photometry}
Photometric 3D-from-2D face reconstruction methods estimate the lighting parameters and surface normals from a set of images usually assuming a Lambertian reflectance model, which defines each pixel $(x,y)$ in an image $\vec{I}$ as
\begin{equation}\label{eq:LambertianModel}
    \vec{I}(x,y) = \mathcal{A}(x,y) \vec{l}^{\text{T}} \vec{n}(x,y)
\end{equation}
where $\mathcal{A}(x,y)$ is the albedo of the pixel $(x,y)$, $\vec{n}$ its surface normal and $\vec{l}$ the light source vector. This model can be approximated using spherical harmonics basis functions, $\vec{\mathcal{Y}}(\vec{n})$, as
\begin{equation} \label{eq:LambertianModel_SH}
    \vec{I}(x,y) \approx \mathcal{A}(x,y)\vec{\gamma}^{\text{T}}\vec{\mathcal{Y}}(\vec{n})
\end{equation}
where $\vec{\gamma}$ are the coefficients of the spherical harmonics functions.

This approach to reconstruct the 3D geometry of a surface based on photometry was originally introduced by \citet{WoodhamOpEng1980} who proposed to estimate the surface normals from several 2D images by observing the object under different lighting conditions. However, Woodham's work assumed a rigid geometry of the object, fixed Lambertian reflectance, fixed camera pose, and uniform albedo. These assumptions have been relaxed by subsequent works to adapt to more realistic settings.

As stated above, the 3D-from-2D face reconstruction problem is ill-posed when considering a single image, and additional constraints are needed to adequately constrain the space of solutions. Some approaches, like \citet{WoodhamOpEng1980}, did so by using only a collection of images of the same subject \citep{KemelmacherICCV2011,SLiangECCV2016,SuwajanakornECCV2014,SnapeCVPR2015} (Section \ref{subsec:PS_Multiple}), whereas others also used template shapes \citep{RothCVPR2015,DZengImgVisCom2017} or 3D facial models \citep{RothCVPR2016} (Section \ref{subsec:PS_Single}). The reviewed works on 3D-from-2D face reconstruction based on photometry are summarised in Table \ref{tab:PS}, and main conclusions are outlined in Section \ref{subsec:PS_conclusions}.

\begin{table*}[bhp]
\caption{3D-from-2D face reconstruction approaches based on photometry.} \label{tab:PS}
\renewcommand\tabularxcolumn[1]{m{#1}}
\begin{tabularx}{\textwidth}{Xp{2cm}Xl}
\hline
Reference & Images & Prior & \begin{tabular}[c]{@{}l@{}}Illumination\\model\end{tabular} \\ \hline
\citet{KemelmacherICCV2011} & Multiple & Collection of images & Lambertian \\ \hline
\citet{KemelmacherPAMI2011} & Single & Template mesh & Lambertian \\ \hline
%\citet{KumarPAMI2011} & Single & Illumination-shape model & Non-Lambertian \\ \hline
\citet{MLeePR2011} & Single & Illumination-shape model & Lambertian \\ \hline
\citet{MLeePR2013} & Single & Illumination-shape model & Lambertian \\ \hline
\citet{MLeeCVIU2014} & Single & Illumination-shape model & Lambertian \\ \hline
\citet{SuwajanakornECCV2014} & Video & Collection of images & Lambertian \\ \hline
\citet{SnapeCVPR2015} & Multiple & \begin{tabular}[c]{@{}l@{}}Collection of images from\\ multiple subjects\end{tabular} & Lambertian \\ \hline
\citet{RothCVPR2015} & Multiple & \begin{tabular}[c]{@{}l@{}}Collection of images and\\ template mesh\end{tabular} & Lambertian \\ \hline
\citet{RothCVPR2016} & Multiple & Collection of images and 3DMM & Lambertian \\ \hline
\citet{SLiangECCV2016} & Multiple & Collection of images & Lambertian \\ \hline
\citet{DZengImgVisCom2017} & Three & Collection of reference meshes & Lambertian \\ \hline
\citet{XCaoCVPR2018} & Single & 3DMM fitting & Lambertian \\ \hline
\citet{YLiCVMP2018} & Single & 3DMM fitting & Lambertian \\ \hline
\citet{RotgerWSCG2019} & Single & 3DMM fitting & Lambertian \\ \hline
\end{tabularx}
\end{table*}

\subsection{Multiple image methods} \label{subsec:PS_Multiple}
\citet{KemelmacherICCV2011} reconstruc\-ted the 3D facial shape of a subject given a collection of in-the-wild photos. They proposed to decompose the matrix $\vec{M} \in \mathbb{R}^{n\times d}$ of the $n$ vectorised and frontalised images into a matrix $\vec{\Gamma} \in \mathbb{R}^{n \times 4}$ containing the lighting coefficients and a matrix $\vec{S} \in \mathbb{R}^{4 \times d}$ containing the albedo (first row) and the surface normals. The initial estimation of $\vec{\Gamma}$ and $\vec{S}$ was computed using singular value decomposition and taking the rank-4 approximation. However, this produced 3D facial reconstructions with insufficient details, thus they proposed an additional step to iteratively refine the matrix $\vec{S}$: for each pixel, they selected the images whose decomposition error was small in that pixel, and used them to recalculate the normal vector (the corresponding column of $\vec{S}$). Although with this approach, \citet{KemelmacherICCV2011} were able to reconstruct detailed 3D faces, a large amount of photos of the same person is needed, which is not always available.

The above work inspired many others \citep{SnapeCVPR2015,SuwajanakornECCV2014,SLiangECCV2016,RothCVPR2015}. Specifically,  \citep{SLiangECCV2016,SuwajanakornECCV2014} used the method proposed by \citep{KemelmacherICCV2011} as a part of their reconstruction pipeline: the former \citep{SLiangECCV2016} to reconstruct the whole head of a person from a collection of photos, and the latter \citep{SuwajanakornECCV2014} to build a personalised reference shape, which was later refined. To reconstruct the whole head from a set of images, \citet{SLiangECCV2016} clustered it according to the yaw angle of the face, and used the frontal cluster to initialise the reconstruction, which covered a limited part of the face. Thus, to recover the rest of the head, the remaining clusters were used to progressively extend the reconstruction. On the other hand, the goal of \citet{SuwajanakornECCV2014} was to reconstruct a 3D shape for each frame of a video sequence, hence, they deformed the personalised template shape build with \citep{KemelmacherICCV2011} to match each frame. These deformations were estimated via a proposed 3D optical flow approach combined with shading cues. This 3D optical flow algorithm computed dense correspondences between the 3D personalised shape and each frame, making possible to deform the mesh to fit the image.

In contrast to \citep{SLiangECCV2016,SuwajanakornECCV2014}, \citet{SnapeCVPR2015} and \citet{RothCVPR2015} extended the work proposed by \citep{KemelmacherICCV2011}: the former \citep{SnapeCVPR2015} by including several identities in the input photo collection, and the latter \citep{RothCVPR2015} by using non-frontal images. In particular, \citet{SnapeCVPR2015} proposed to decompose a matrix formed by vectorised images but not restricting the collection of input photos to a single subject. Thus, they proposed to decompose $\vec{M}$ into $(\vec{\Gamma} * \vec{C})\vec{S}$, where $*$ denotes the Khatri-Rao product and $\vec{C}$ is a matrix containing the shape coefficients, related to identity. This allowed them to recover normals for multiple subjects at the same time, unlike \citep{KemelmacherICCV2011}, who only recovered the facial geometry of one individual. On the other hand, \citet{RothCVPR2015} argued that non-frontal images are very useful for 3D face reconstruction, in contrast to \citep{KemelmacherICCV2011}, who used near-frontal images. To obtain facial regions for each of the input images, they wrapped a template mesh to each of them, obtaining individual projection parameters. Then, the matrix $\vec{M}$ built from the facial regions was decomposed into $\vec{\Gamma}$ and $\vec{S}$ following \citep{KemelmacherICCV2011}. With this approach, they obtained less noisy faces than \citep{KemelmacherICCV2011} while capturing more accurate fine details. However, in their subsequent publication \citep{RothCVPR2016}, Roth et al. noticed that their former approach \citep{RothCVPR2015} had two limitations: on the one hand, the facial template used has a specific ethnicity and thus may fit poorly to other ethnicities; and, on the other hand, their method fails when the number of images is small and limited pose variations are included. They overcame these shortcomings by building a personalised template using 3DMM fitting as \citep{XZhuCVPR2015}. Also, instead of following \citep{KemelmacherICCV2011} to reconstruct the 3D face, they proposed to minimise an energy function consisting of a term that penalised the difference between each image pixel and its estimated value (according to eq. \eqref{eq:LambertianModel}), and another one that penalised the distance between the ground truth and the estimated surface normals. Finally, they used a coarse-to-fine scheme to first fit the overall face shape and later adapt the coarse estimation to the details present in the collection. With this modifications, they were able to reduce the reconstruction errors with respect to \citep{RothCVPR2015}.

Differently from the works above, \citet{DZengImgVisCom2017} proposed to use only three images (one frontal and two profile ones) and a set of reference meshes as prior knowledge. First, an initial estimation of the shape was computed using a single-image reconstruction method \citep{KemelmacherPAMI2011}. This initialisation was refined by minimising an energy function with three terms: a shading term that penalised dissimilarity between the initial and the new estimates; a multi-view consistency term that forced projections of the same 3D point onto neighbouring views to be similar for all the reference shapes; and finally a smoothness term that ensured smooth transitions in depth. According to \citet{DZengImgVisCom2017}, using a set of reference meshes, contrary to using a single template mesh, helps increasing the probability of finding the most similar candidate for the input face.

\subsection{Single image methods} \label{subsec:PS_Single}

Differently from the works studied above, \citep{KemelmacherPAMI2011,MLeePR2011,MLeePR2013,MLeeCVIU2014,RotgerWSCG2019,XCaoCVPR2018,YLiCVMP2018} recovered the 3D face of a person from a single image, thus additional prior knowledge has to be included in the reconstruction process. We have identified three different ways of incorporating such prior information: the first one is using a pre-designed template mesh that is fitted to the input image \citep{KemelmacherPAMI2011}, the second one is by training a model that integrates illumination and shape information \citep{MLeePR2011,MLeePR2013,MLeeCVIU2014}, and the last one is by fitting a 3DMM to obtain a coarse 3D face estimation that is later refined \citep{RotgerWSCG2019,XCaoCVPR2018,YLiCVMP2018}.

In the first category, \citet{KemelmacherPAMI2011} used a template face, for which the albedo, surface normals, and depth map were known; and recovered each of the elements individually by fixing the rest. More precisely, the proposed reconstruction scheme consisted in: 1) recovering the spherical harmonic coefficients $\vec{\gamma}$ by fitting the reference shape to the input image and fixing the normals and albedo of the template, 2) estimating the depth map for the input image given the estimated $\vec{\gamma}$ and the albedo of the template, and 3) recovering the albedo with the estimated spherical harmonic coefficients and depth map. However, this method is highly dependent on the image and the template used, obtaining geometries that vary significantly among reconstructions of the same subject.

The works on the second category \citet{MLeePR2011,MLeePR2013,MLeeCVIU2014} reformulated the estimation of the 3D shape as a photometry-based model fitting problem. Whereas the above approaches vectorised the images to build 2D matrices, Lee an Choi kept the 2D nature of the images and worked with tensors. They assumed a Lambertian reflectance model and approximated it using spherical harmonics such that
\begin{equation*}
    \vec{I} = \vec{F} \times_3 \vec{l}^\text{T}
\end{equation*}
where $\vec{I} \in \mathbb{R}^{n_x\times n_y}$ is an image, $\vec{F} \in \mathbb{R}^{n_x\times n_y \times n_l}$ is the 3rd-order tensor related to surface characteristics (albedo, normals), and $\vec{l} \in \mathbb{R}^{n_l}$ is the light source vector. In \citep{MLeePR2011}, Lee et al. proposed to parametrise $\vec{F}$ as a function of a personal identity vector $\vec{\iota}$ by decomposing it similarly to 3DMM (see Section \ref{subsec:Build3Dmodel}) and using $N$-mode singular value decomposition so that $\vec{F} = \vec{\bar{F}} + \vec{T} \times_4 \vec{\iota}$, where $\vec{\bar{F}} \in \mathbb{R}^{n_x \times n_y \times n_l}$ is the mean of $\vec{F}$, $\vec{T} \in \mathbb{R}^{n_x \times n_y \times n_l \times n_{id}}$ is formed by the bases functions, and $\vec{\iota} \in \mathbb{R}^{n_{id}}$. Thus, an image $\vec{I}_\text{new}$ is reconstructed by finding the $\vec{l}$ and $\vec{\iota}$ vectors that minimise the reconstruction error between $\vec{I}_\text{new}$ and $(\vec{\bar{F}} + \vec{T} \times_4 \vec{\iota}) \times_3 \vec{l}^\text{T}$.

In \citep{MLeePR2013}, Lee et al. proposed a me\-thod to estimate an image depth by fitting a model that considers cast shadows. Similarly to \citep{MLeePR2011}, they first reduced the dimension of the training images by building a subspace with tensor decomposition techniques. The representations of the training examples in that subspace were transformed to hyperspherical coordinates to simplify the problem. Finally, a linear mapping was learnt to estimate the depth from these hyperspherical coordinates. Although with this approach they obtained slightly less accurate reconstructions than with their previous work \citep{MLeePR2011}, the computational cost was reduced about 10 times. The contributions presented in \citep{MLeePR2011} and \citep{MLeePR2013} was compiled in \citep{MLeeCVIU2014}.

% KumarPAMI2011 --> requires at least nine input images with known illumination directions
%\citet{KumarPAMI2011}, different from all the works above, did not assume a Lambertian model but used apparent bidirectional reflectance distribution functions to model the illumination of faces. They proposed to approximate the field of such functions using \textit{tensor splines}, which are combinations of Cartesian tensors and B-splines. However, whereas the Lambertian model (see eq. \eqref{eq:LambertianModel}) explicitly includes the surface normals in the formula, \textit{tensor splines} do not; thus, recovering the surface normals from the illumination in the image is a nontrivial task. To solve this problem, Kumar \emph{et al.} assumed local homogeneity of the apparent bidirectional reflectance distribution functions field, i.e. neighbouring pixels have the same shape, only differing by a rotation, which allowed them to derive a surface normal at a pixel by rotating the known normal at another pixel.

Finally, in the third category, \citep{RotgerWSCG2019,YLiCVMP2018,XCaoCVPR2018} included a 3DMM to ensure global plausibility of the reconstructions. In fact, they only used photometry-based methods to refine a coarse 3D face that was estimated by fitting a 3DMM. The idea behind these works is to refine the coarse 3D face estimating per-vertex displacements, which result from the minimisation of an image error
\begin{equation*}
    \mathbb{E}_{\mathbf{I}} = \norm{ \Iin - \mathcal{A}\vec{\gamma}^\text{T}\vec{\mathcal{Y}} }_2^2
\end{equation*}
where $\mathcal{A}$, $\vec{\gamma}$, and $\vec{\mathcal{Y}}$ are as in eq. \eqref{eq:LambertianModel_SH}.

\citet{RotgerWSCG2019} estimated per-vertex displacement $\Delta \vec{v}^i$ to account for wrinkles in $\Iin$, which were detected as changes of the facial texture according to the partial derivatives in both horizontal and vertical directions. Thus, the updated vertices $\vec{v}_i + \Delta\vec{v}_i$ were obtained by minimising
\begin{equation*}
    \mathbb{E}_{\mathbf{I}} = \sum_{i=1}^{n} \norm{ \mathbf{I}_\text{in}(i) - \mathcal{A}_i\vec{\gamma}^\text{T}\vec{\mathcal{Y}}(\vec{n}[\vec{v}_i+\Delta\vec{v}_i])}_2^2
\end{equation*}
where $\vec{n}[\vec{v}_i+\Delta\vec{v}_i]$ denotes the normal vector estimated at the updated vertex $\vec{v}_i+\Delta\vec{v}_i$, $\vec{\mathcal{Y}}(\vec{n})$ denotes the spherical harmonic basis functions computed for $\vec{n}$, and $n$ is the number of pixels in $\Iin$. Although this method is able to recover fine details from the input image, it fails in dealing with facial tattoos and beards because of the way the wrinkles are modelled, and also the reconstructions can be noisy.

In contrast, \citet{YLiCVMP2018} and \citet{XCaoCVPR2018} directly updated the normals, instead of estimating displacements for each vertex. The former \citep{YLiCVMP2018} added the shadows, $ \mathcal{B}$, that cannot be explained by a Lambertian model to their image formation equation:
\begin{equation}\label{eq:Imod_YLi}
    \mathbf{I} = \mathcal{A}\vec{\gamma}^\text{T}\vec{\mathcal{Y}} + \mathcal{B}.
\end{equation}
They iteratively estimated $\mathcal{A}$ and $\mathcal{B}$ by minimising the difference between both sides of eq. \eqref{eq:Imod_YLi}, keeping the surface normals fixed to those from the coarse 3D reconstruction. In a following step, they fixed $\mathcal{A}$ and $\mathcal{B}$, and estimated the normals. Thus, by adopting a more complex reflectance model and an iterative procedure, \citet{YLiCVMP2018} were able to recover facial details very accurately, enhancing the results from \citep{RotgerWSCG2019}, however, their reconstructions were still noisy and facial hair was misinterpreted as shape variations. 

Differently from \citep{RotgerWSCG2019,YLiCVMP2018}, \citet{XCaoCVPR2018} considered a near point light model, instead of distant directional light sources, whose computational cost is lower since it only requires a subset of key pixels and light sources. They modelled the near point light model by changing the light source vector $\vec{l}$ in eq. \eqref{eq:LambertianModel} for the scaled directions $\vec{e}$, which depend on the light positions and brightness. Thus, they jointly estimated the albedo $\mathcal{A}$, the normals update, and the light positions and brightness, by minimising the difference between the input and the modelled image. However, unlike \citep{RotgerWSCG2019,YLiCVMP2018}, they added a final step in the reconstruction process to denoise hairy regions, which allowed them to obtain more realistic and less noisy 3D faces than \citet{YLiCVMP2018}.

\subsection{Take-home message} \label{subsec:PS_conclusions}
3D-from-2D face reconstruction methods based on photometry try to recover the surface normals from one or more 2D images, generally, by assuming a Lambertian illumination model that decomposes an RGB image into albedo, lighting and normals. However, since the images are unconstrained, the light source is unknown and so is the pure albedo. Therefore, additional prior knowledge has to be added to the reconstruction problem in order to constrain it.

The type of prior knowledge used in the proposed works clearly separates them into two groups: the ones that constrain the reconstruction by taking a collection of images from the same subject under different lighting conditions, poses, expressions, etc., and the ones that use a single image but add 3D constraints in the form of 3DMMs or 3D facial templates.

Essentially, the approaches that use a collection of images are based on the decomposition of the matrix formed by the vectorised images into two matrices, one containing the lighting information and the other one containing the albedo and surface normals. Although these methods are able to capture fine details, the amount of images needed is generally large and the resulting reconstructions are rather noisy, since they lack a geometric prior to constrain the solution to be a plausible face.

On the other hand, the single-image approaches are more heterogeneous. For example, some works constructed an illumination-shape model that is then fitted to the image, whereas others refined the coarse 3D face obtained with 3DMM-fitting by minimising the difference between the input image and the image obtained with the Lambertian model using the estimated refined normals. This last category of methods based on 3DMM is the most promising since, by combining the prior knowledge of the global facial shape encoded in the 3DMMs and the fine details that can be captured by photometry-based approaches, they are able to reconstruct realistic 3D faces that are highly detailed without neglecting global plausibility. 

\section{Deep Learning Methods} \label{sec:DL}
% In the first two approaches for 3D-from-2D face reconstruction (statistical model fitting and photometric stereo), the prior knowledge needed to constrain the ill-posed problem is encoded in a 3D facial model or in a predefined template, or, as an alternative, a large collection of photos is used. In contrast, deep learning is able to learn the solutions space given a sufficiently large training set containing pairs of 2D images and their corresponding 3D faces.

The 3D-from-2D face reconstruction methods from Section \ref{sec:ModelFittingMethods} and \ref{sec:Photometry} use models to embody prior knowledge: statistical model fitting methods include a geometry (and usually texture) model, and photometric methods model the reflectance of the face. In contrast, deep learning methods directly learn the mapping between the 2D image and the 3D face, encoding prior knowledge in the weights of the trained network.

Although deep learning has shown to be a very powerful tool in many different applications, its direct application in 3D-from-2D face reconstruction is hampered by the lack of ground truth 3D facial scans. However, researchers have proposed different approaches to generate and learn from realistic representative training data, circumventing the obstacle of the lack of ground truth data.

In this section, we present and compare the most relevant works in 3D-from-2D face reconstruction that use deep learning as the main tool. Among the many elements that are involved in the learning process, we consider three representative ones, namely, 1) the training set used to train the network, 2) the learning framework, and 3) the training criterion. We organise this section according to these items.

\begin{table*}[thp]
\begin{threeparttable}[b]
\caption{3D-from-2D face reconstruction approaches based on deep learning - part I.} \label{tab:DL1}
\renewcommand\tabularxcolumn[1]{m{#1}}
\begin{tabularx}{\textwidth}{l|>{\centering\arraybackslash}m{0.5cm}>{\centering\arraybackslash}m{1cm}|m{1.6cm}m{2.2cm}XX|YYY}%{XllXXXX}%{Xlm{1.2cm}m{2.45cm}m{2.65cm}m{2.58cm}X}
\hline
\multirow{2}{*}{Reference} & \multicolumn{2}{l|}{\begin{tabular}[l]{@{}l@{}}Synthetic\\training\end{tabular}} & \multicolumn{4}{l|}{Learning framework} & \multicolumn{3}{l}{Training criterion}\\ \cline{2-10}
& Fit & Render & Network type & \# Layers & Skip connect. & I\-te\-ra\-ti\-ve & Param. space & 3D space &2D space\\ \hline

\multirow{2}{*}{\citet{SongTIP2012}} & \multicolumn{2}{c|}{\multirow{2}{*}{$-$}} & Encoder & 3 & No & No & \multirow{2}{*}{x} & \multirow{2}{*}{} & \multirow{2}{*}{} \\
 & \multicolumn{2}{l|}{} & Decoder & 3 & No & No & & & \\ \hline

\citet{XZhuCVPR2016} & x & x & CNN & 6 & No & Yes & x & & \\ \hline

\citet{Richardson3DV2016} & & x & Residual & 18 & Yes & Yes & & x & \\ \hline

\citet{JourablooCVPR2016} & x & & CNN & 6 & No & Yes & x & & \\ \hline

\multirow{2}{*}{\citet{JourablooIJCV2017}} & \multirow{2}{*}{x} & \multirow{2}{*}{} & CNN & 5 & No & No & \multirow{2}{*}{x} & \multirow{2}{*}{} & \multirow{2}{*}{x}\\
 & & & CNN & 6 & No & Yes & & &  \\ \hline

\multirow{2}{*}{\citet{BhagavatulaICCV2017}} & \multirow{2}{*}{x} & \multirow{2}{*}{x} & CNN & 8 (16)\tnote{1} & No & No & \multirow{2}{*}{} & \multirow{2}{*}{x} & \multirow{2}{*}{} \\
 & & & FC & \#lmks//4//4 & No & No & & & \\ \hline

\multirow{2}{*}{\citet{DouCVPR2017}} & \multirow{2}{*}{} & \multirow{2}{*}{x} & CNN & 16 & No & No & \multirow{2}{*}{} & \multirow{2}{*}{x} & \multirow{2}{*}{} \\
 & & & CNN & 19 & Yes & No & & & \\ \hline

\multirow{2}{*}{\citet{GulerCVPR2017}} & \multirow{2}{*}{x} & \multirow{2}{*}{x} & Residual & 101 & Yes & No & \multirow{2}{*}{x} & \multirow{2}{*}{} & \multirow{2}{*}{}\\
& & & FC & $4\times 1$ & No & No & & & \\ \hline

\multirow{2}{*}{\citet{JacksonICCV2017}} & \multirow{2}{*}{x} & \multirow{2}{*}{x} & Encoder & 6 & Yes & No & \multirow{2}{*}{} & \multirow{2}{*}{x} & \multirow{2}{*}{} \\
 & & & Decoder & 5 & Yes & No & & & \\ \hline

\multirow{2}{*}{\citet{RichardsonCVPR2017}} & \multirow{2}{*}{} & \multirow{2}{*}{x} & Residual & 19 & Yes & Yes & \multirow{2}{*}{} & \multirow{2}{*}{x} & \multirow{2}{*}{x} \\
 & & & CNN & 16 & No & No & & & \\ \hline

\multirow{2}{*}{\citet{SelaICCV2017}} & \multirow{2}{*}{} & \multirow{2}{*}{x} & Encoder & 8 & Yes & No & \multirow{2}{*}{} & \multirow{2}{*}{x} & \multirow{2}{*}{} \\
 & & & Decoder & 8 & Yes & No & & & \\ \hline

\citet{TewariICCV2017}& x & x & CNN & 8 & No & No & & & x  \\ \hline

\citet{JourablooICCV2017}& x & & CNN & 24 & Yes & No & x & & \\ \hline

\citet{TrigeorgisCVPR2017}& x & x & Residual & 50 & Yes & No & & x & \\ \hline

\citet{TranCVPR2017} & x & & Residual & 101 & Yes & No & x & & \\ \hline

\multirow{3}{*}{\citet{TranCVPR2018}} & \multirow{3}{*}{x} & \multirow{3}{*}{} & Residual & 101 & Yes & No & \multirow{3}{*}{x} & \multirow{3}{*}{} & \multirow{3}{*}{} \\
 & & & Encoder & 8 & Yes & No & & & \\
 & & & Decoder & 8 & Yes & No & & & \\ \hline

 \multirow{2}{*}{\citet{FLiuCVPR2018}} & \multirow{2}{*}{x} & \multirow{2}{*}{} & Encoder & 21 & No & No & \multirow{2}{*}{} & \multirow{2}{*}{x} & \multirow{2}{*}{} \\
 & & & Decoder & $2\times1$ & No & No & & & \\ \hline
 
 \multirow{2}{*}{\citet{FLiuICCV2019}} & \multirow{2}{*}{} & \multirow{2}{*}{x} & Encoder & 21 & No & No & \multirow{2}{*}{} & \multirow{2}{*}{x} & \multirow{2}{*}{} \\
 & & & Decoder & $2\times1$ & No & No & & & \\ \hline

\citet{KimCVPR2018} & x & x & CNN & 8 & No & Yes & x & & \\ \hline

\multirow{2}{*}{\citet{LTranCVPR2018}} & \multirow{2}{*}{x} & \multirow{2}{*}{x} & Encoder & 14 & No & No & \multirow{2}{*}{} & \multirow{2}{*}{} & \multirow{2}{*}{x} \\
 & & & Decoder & 15 // 17 & No & No & & & \\ \hline

\multirow{3}{*}{\citet{SenguptaCVPR2018}} & \multirow{3}{*}{} & \multirow{3}{*}{x} & CNN & 3 & No & No & \multirow{3}{*}{x} & \multirow{3}{*}{} & \multirow{3}{*}{x} \\
 & & & Residual & $2\times10$ & Yes & No & & & \\
 & & & CNN & ($2\times3$) // 2 & No & No & & & \\ \hline

\citet{TewariCVPR2018} & x & x & CNN & 8 & No & No & & & x  \\ \hline

\multirow{2}{*}{\citet{YFengECCV2018}} & \multirow{2}{*}{x} & \multirow{2}{*}{x} & Encoder & 21 & Yes & No & \multirow{2}{*}{} & \multirow{2}{*}{x} & \multirow{2}{*}{} \\
& & & Decoder & 17 & No & No & & & \\ \hline
 
\multirow{3}{*}{\citet{YGuoPAMI2018}} & \multirow{3}{*}{} & \multirow{3}{*}{x} & Residual & 18 & Yes & No & \multirow{3}{*}{} & \multirow{3}{*}{} & \multirow{3}{*}{x} \\
& & & Encoder & 10 & Yes & No & & & \\
& & & Decoder & 9 & Yes & No & & & \\ \hline
 
\multirow{2}{*}{\citet{GenovaCVPR2018}} & \multirow{2}{*}{} & \multirow{2}{*}{x} & Encoder & 22 & No & Yes & \multirow{2}{*}{x} & \multirow{2}{*}{} & \multirow{2}{*}{} \\
 & & & Decoder & 3 & No & Yes & & & \\ \hline
 
\citet{TewariPAMI2020} & x & x & CNN & 8 & No & No & & & x \\ \hline

\citet{XZhuPAMI2019} & x & x & CNN & 6 // 7 & No & Yes & x &  &  \\ \hline

\multirow{2}{*}{\citet{LTranPAMI2019}} & \multirow{2}{*}{x} & \multirow{2}{*}{x} & Encoder & 14 & No & No & \multirow{2}{*}{} & \multirow{2}{*}{} & \multirow{2}{*}{x} \\
 & & & Decoder & 15 // 17 & No & No & & & \\ \hline

\end{tabularx}
\begin{tablenotes}
\vspace{0.5em}
\item [1] \footnotesize{\citet{BhagavatulaICCV2017} compare two different CNNs, one with 8 layers and the other with 16.}
\end{tablenotes}
\end{threeparttable}
\end{table*}

\begin{table*}[thp]
\begin{threeparttable}[b]
\caption{3D-from-2D face reconstruction approaches based on deep learning - part II.} \label{tab:DL2}
\renewcommand\tabularxcolumn[1]{m{#1}}
% \begin{tabularx}{\textwidth}{l|>{\centering\arraybackslash}m{0.4cm}Y|m{1.6cm}m{2.2cm}XX|YYY}
\begin{tabularx}{\textwidth}{l|>{\centering\arraybackslash}m{0.5cm}>{\centering\arraybackslash}m{1cm}|m{2cm}m{2.6cm}XX|YYY}
%{XllXXXX}%{Xlm{1.2cm}m{2.45cm}m{2.65cm}m{2.58cm}X}
\hline
\multirow{2}{*}{Reference} & \multicolumn{2}{l|}{\begin{tabular}[l]{@{}l@{}}Synthetic\\training\end{tabular}} & \multicolumn{4}{l|}{Learning framework} & \multicolumn{3}{l}{Training criterion}\\ \cline{2-10}
& Fit & Render & Network type & \# Layers & Skip connect. & I\-te\-ra\-ti\-ve & Param. space & 3D space &2D space\\ \hline

\multirow{2}{*}{\citet{LTranCVPR2019}} & \multirow{2}{*}{x} & \multirow{2}{*}{x} & Encoder & 14 & No & No & \multirow{2}{*}{} & \multirow{2}{*}{} & \multirow{2}{*}{x} \\
 & & & Decoder & 15 // 17 & No & No & & & \\ \hline

\multirow{2}{*}{\citet{YZhouCVPR2019}} & \multirow{2}{*}{x} & \multirow{2}{*}{x} & Encoder & 11 // 4 & No & No &  & \multirow{2}{*}{x} & \multirow{2}{*}{x} \\
& & & Decoder & 4 & No & No & & & \\ \hline

\multirow{2}{*}{\citet{SanyalCVPR2019}} & \multicolumn{2}{c|}{\multirow{2}{*}{$-$}} & Residual & 50 & Yes & No & \multirow{2}{*}{x} &  & \multirow{2}{*}{x} \\ 
& & & FC & 3 & No & Yes & & & \\ \hline

\multirow{2}{*}{\citet{GalteriCVPRW2019}} & \multirow{2}{*}{x} & \multirow{2}{*}{x} & Generator & 19(enc)+19(dec)\tnote{1} & No & No &  & \multirow{2}{*}{x} &  \\
& & & Discriminator & 19 & No & No & & & \\ \hline

\multirow{2}{*}{\citet{GalteriCVIU2019}} & \multirow{2}{*}{x} & \multirow{2}{*}{x} & Generator & 19(enc)+19(dec)\tnote{1} & No & No &  & \multirow{2}{*}{x} &  \\
& & & Discriminator & 19 & No & No & & & \\ \hline

\citet{YDengCVPRW2019} & x & x & Residual & 50 & Yes & No & x &  & x \\ \hline

\multirow{2}{*}{\citet{FWuCVPR2019}} & \multirow{2}{*}{x} & \multirow{2}{*}{x} & CNN  & 19 & No & No & \multirow{2}{*}{x} &  & \multirow{2}{*}{x} \\
& & & FC & $2\times2$ & No & No & & & \\ \hline

\multirow{4}{*}{\citet{HYiCVPR2019}} & \multirow{4}{*}{x} & \multirow{4}{*}{x} & Encoder\tnote{2} & 6 & Yes & No &  & \multirow{4}{*}{x} & \multirow{4}{*}{x} \\
& & & Decoder\tnote{2} & 5 & Yes & No & & & \\
& & & CNN & 5 & No & No & & & \\
& & & FC & $3\times1$ & No & No & & & \\ \hline

\multirow{4}{*}{\citet{JPiaoICCV2019}} &  & \multirow{4}{*}{x} & Generator\tnote{3} & 8(enc)+8(dec)\tnote{3}& Yes & No &  & \multirow{4}{*}{x} & \multirow{4}{*}{x} \\
& & & Discriminator\tnote{3} & 6 & No & No & & & \\
& & & Encoder & 20 & No & No & & & \\
& & & Decoder & 5 & No & No & & & \\ \hline

\multirow{4}{*}{\citet{JSYoonCVPR2019}} & \multirow{4}{*}{x} &  & Encoder\tnote{4} & 19 & No & No & \multirow{4}{*}{x} &  & \multirow{4}{*}{x} \\
& & & Encoder\tnote{4} & 5(enc)+6(dec)\tnote{5} & Yes & No & & & \\
& & & FC & $2\times4$ & No & No & & & \\
& & & Decoder & 10 & No & No & & & \\ \hline

\citet{PWangICIP2019} & \multicolumn{2}{c|}{$-$} & CNN & 28 & No & No &  &  & x \\ \hline

\citet{SavovICCVW2019} & \multicolumn{2}{c|}{$-$} & CNN & 7 & No & No &  &  & x \\ \hline

\multirow{2}{*}{\citet{RamonICCVW2019}} & \multicolumn{2}{c|}{\multirow{2}{*}{$-$}} & CNN & 16 & No & No &  & \multirow{2}{*}{x} &  \\
& & & MLP & 4 & No & No & & & \\ \hline

\multirow{3}{*}{\citet{XZengICCV2019}} & \multirow{3}{*}{x} & \multirow{3}{*}{x} & Encoder & 10 & Yes & No &  & \multirow{3}{*}{x} & \multirow{3}{*}{x} \\
& & & Decoder & 9 & Yes & No & & & \\
& & & CNN & 9+16 \tnote{6} & Yes & No & & & \\ \hline

\multirow{2}{*}{\citet{BaiCVPR2020}} & & \multirow{2}{*}{x} & Residual & 49 & Yes & Yes &  & \multirow{2}{*}{x} & \multirow{2}{*}{x} \\
 & & & Residual & 16 & Yes & Yes &  &  &  \\\hline

\citet{ChinaevECCVW2018} & x & & CNN & 28 & No & No &  & x & x \\ \hline

\multirow{2}{*}{\citet{ChaudhuriCVPR2019}} & \multirow{2}{*}{x} & & CNN & 25 & No & No &  & \multirow{2}{*}{x} & \multirow{2}{*}{x} \\
& & & FC & $3\times1$ & No & No & & & \\ \hline

\multirow{3}{*}{\citet{ChaudhuriECCV2020}} & \multicolumn{2}{c|}{\multirow{3}{*}{$-$}} & Residual & 18 & Yes & No & \multirow{3}{*}{x} & \multirow{3}{*}{x} & \multirow{3}{*}{x} \\
& & & Encoder & 8 & Yes & No & & & \\
& & & Decoder & 8 & Yes & No & & & \\ \hline

\multirow{2}{*}{\citet{GZhangFG2018}} & \multirow{2}{*}{x} & \multirow{2}{*}{x} & CNN & 4 & No & No &  & \multirow{2}{*}{x} & \multirow{2}{*}{x} \\
& & & FC & $3\times3$ & No & No & & & \\ \hline

\end{tabularx}
\begin{tablenotes}
\vspace{0.5em}
\item [1] \footnotesize{\citet{GalteriCVPRW2019,GalteriCVIU2019} proposed a generator with an encoder-decoder architecture, with 19 layers each.}
\item [2] \footnotesize{\citet{HYiCVPR2019} stacked two encoder-decoder networks, followed by two CNNs.}
\item [3] \footnotesize{\citet{JPiaoICCV2019} considered a CycleGAN composed of two stacked GANs. The generators are encoder-decoder networks with 8 layers each. }
\item [4] \footnotesize{\citet{JSYoonCVPR2019} used two parallel encoders for the same decoder.}
\item [5] \footnotesize{One of the encoders used by \citet{JSYoonCVPR2019} is an hourglass network, which has an encoder-decoder architecture with 5 and 6 layers, respectively.}
\item [6] \footnotesize{\citet{XZengICCV2019} trained separately two CNN with skip connections, one that refines the output of the other.}
\end{tablenotes}
\end{threeparttable}
\end{table*}

\begin{table*}[thp]
\begin{threeparttable}[b]
\caption{3D-from-2D face reconstruction approaches based on deep learning - part III.} \label{tab:DL3}
\renewcommand\tabularxcolumn[1]{m{#1}}
% \begin{tabularx}{\textwidth}{l|>{\centering\arraybackslash}m{0.4cm}Y|m{1.6cm}m{2.2cm}XX|YYY}
\begin{tabularx}{\textwidth}{l|>{\centering\arraybackslash}m{0.5cm}>{\centering\arraybackslash}m{1cm}|m{2cm}m{2.6cm}XX|YYY}
%{XllXXXX}%{Xlm{1.2cm}m{2.45cm}m{2.65cm}m{2.58cm}X}
\hline
\multirow{2}{*}{Reference} & \multicolumn{2}{l|}{\begin{tabular}[l]{@{}l@{}}Synthetic\\training\end{tabular}} & \multicolumn{4}{l|}{Learning framework} & \multicolumn{3}{l}{Training criterion}\\ \cline{2-10}
& Fit & Render & Network type & \# Layers & Skip connect. & I\-te\-ra\-ti\-ve & Param. space & 3D space &2D space\\ \hline

\multirow{2}{*}{\citet{JGuoECCV2020}} & \multirow{2}{*}{x} & \multirow{2}{*}{x} & CNN & 28 & No & No & \multirow{2}{*}{x} & \multirow{2}{*}{x} & \multirow{2}{*}{x} \\
& & & FC & $2\times1$ & No & No & & & \\ \hline

\multirow{2}{*}{\citet{JLinCVPR2020}} & \multicolumn{2}{c|}{\multirow{2}{*}{$-$}} & Generator & Mixed\tnote{1} & Mixed\tnote{1} & No & \multirow{2}{*}{x} & \multirow{2}{*}{x} & \multirow{2}{*}{x} \\ 
& & & Discriminator & 6 & No & No & & & \\ \hline

% \multirow{5}{*}{\citet{JLinCVPR2020}} & \multicolumn{2}{c|}{\multirow{5}{*}{$-$}} & CNN (Gen) & 22 & No & No & \multirow{5}{*}{x} & \multirow{5}{*}{x} & \multirow{5}{*}{x} \\ 
% & & & Residual (Gen) & 50 & Yes & No & & & \\
% & & & GCN (Gen) & $2\times8$ & Yes & No & & & \\
% & & & GCN (Gen) & 1 & No & No & & & \\
% & & & CNN (Disc) & 6 & No & No & & & \\ \hline

\citet{JShangECCV2020} & x & x & Residual & 50 & Yes & No &  & x & x \\ \hline

\multirow{2}{*}{\citet{KoizumiECCV2020}} & \multicolumn{2}{c|}{\multirow{2}{*}{$-$}} & Encoder & 8 & Yes & No & \multirow{2}{*}{x} & \multirow{2}{*}{x} & \multirow{2}{*}{} \\
 & & & Decoder & 8 & Yes & No & & & \\ \hline

\multirow{5}{*}{\citet{LattasCVPR2020}} & \multicolumn{2}{c|}{\multirow{5}{*}{$-$}} & Encoder\tnote{2} & 19 & No & No & \multicolumn{3}{c}{\multirow{2}{*}{$-$}} \\
 & & & Decoder\tnote{2} & 19 & No & No & & &\\
 & & & CNN & 1013 & Yes & Yes & &  & \multirow{3}{*}{x} \\
 & & & Generator & 24 + (3 $\times$ 24) & Yes & No &  & &  \\
 & & & Discriminator\tnote{2} & $3\times4$ & No & No & & &\\\hline
% \multirow{4}{*}{\citet{LattasCVPR2020}} & \multicolumn{2}{c|}{\multirow{4}{*}{$-$}} & Encoder\tnote{2} & 19 & No & No & \multicolumn{3}{c}{\multirow{2}{*}{$-$}} \\
%  & & & Decoder\tnote{2} & 19 & No & No & & &\\
%  & & & CNN & 1013 & Yes & No & \multicolumn{3}{c}{\multirow{2}{*}{N/A}}\\
%  & & & Residual & 24 + (3 $\times$ 24) & Yes & No &  & &  \\\hline

\multirow{3}{*}{\citet{PWangICME2020}} & \multirow{3}{*}{x} & \multirow{3}{*}{x} & Residual & 18 & Yes & No &  &  & x \\
 & & & Enc-Dec & 4(enc)+4(dec)\tnote{3} & Yes & No & & & x \\
 & & & Enc-Dec & 4(enc)+4(dec)\tnote{3} & Yes & No & & x & x\\\hline

\multirow{3}{*}{\citet{TewariCVPR2019}} & \multicolumn{2}{c|}{\multirow{3}{*}{$-$}} & CNN & 5 & No & No &  &  & \multirow{3}{*}{x} \\
 & & & CNN & 5 & No & No & & & \\
 & & & CNN & 7 & No & No & & & \\\hline
 
\multirow{5}{*}{\citet{XChaiICME2020}} & \multirow{5}{*}{x} & \multirow{5}{*}{x} & Encoder & 16 & No & No & \multirow{5}{*}{x} &  & \multirow{5}{*}{x} \\
 & & & Decoder & $2\times5$ & No & No & & & \\
 & & & FC & 2 & No & No & & & \\
 & & & CNN & 16 & No & No & & & \\
 & & & CNN & 16 & Yes & No & & & \\\hline
 
\multirow{2}{*}{\citet{XFanToM2020}} & \multirow{2}{*}{x} & \multirow{2}{*}{} & CNN & 52 & Yes & No & \multirow{2}{*}{} & \multirow{2}{*}{x} & \multirow{2}{*}{} \\
 & & & MLP & $\#\text{verts}\times1$ & No & No & & & \\\hline

\multirow{2}{*}{\citet{XLiICASSP2020}} & \multirow{2}{*}{x} & \multirow{2}{*}{} & Encoder & 18 // 50 & Yes & No & \multirow{2}{*}{} & \multirow{2}{*}{x} & \multirow{2}{*}{} \\
 & & & Decoder & N/A & No & No & & & \\\hline

\multirow{2}{*}{\citet{XTuToM2020}} & \multirow{2}{*}{x} & \multirow{2}{*}{x} & Generator & 50 & Yes & No & \multirow{2}{*}{} & \multirow{2}{*}{x} & \multirow{2}{*}{} \\
& & & Discriminator & 6(CNN)+4(FC) & No & No & & & \\ \hline
 
\multirow{3}{*}{\citet{XWangCVPR2020}} & \multirow{3}{*}{x} & \multirow{3}{*}{x} & Residual & 18 & Yes & No & \multirow{3}{*}{} & \multirow{3}{*}{x} & \multirow{3}{*}{x} \\
& & & Encoder & (\#images+1)$\times 4$ & No & No & & & \\
& & & Decoder & 5 & No & No & & & \\\hline

\multirow{2}{*}{\citet{XZhuECCV2020}} & \multirow{2}{*}{x} & \multirow{2}{*}{x} & Encoder & 21 & Yes & No & \multirow{2}{*}{} & \multirow{2}{*}{x} & \multirow{2}{*}{} \\
& & & Decoder & 17 & No & No & & & \\ \hline

\multirow{3}{*}{\citet{YChenTIP2020}} & \multicolumn{2}{c|}{\multirow{3}{*}{$-$}} & CNN & 16 & No & No & x &  & x \\
& & & Encoder & 8 & Yes & No & \multirow{2}{*}{} & \multirow{2}{*}{x} & \multirow{2}{*}{x} \\
& & & Decoder & 8 & Yes & No & & & \\\hline

\multirow{2}{*}{\citet{ZGaoCVPRW2020}} & \multirow{2}{*}{x} & \multirow{2}{*}{x} & Generator & Mixed\tnote{4} & Mixed\tnote{4} & No & & \multirow{2}{*}{x} & \multirow{2}{*}{x} \\
 & & & Discriminator & N/A & No & No & & & \\\hline

% \multirow{5}{*}{\citet{ZGaoCVPRW2020}} & \multirow{5}{*}{x} & \multirow{5}{*}{x} & Residual (Gen) & 50 & Yes & No & & \multirow{5}{*}{x} & \multirow{5}{*}{x} \\
%  & & & FC (Gen) & 4$\times$1 & No & No & & & \\
%  & & & GCN (Gen) & $2\times7$ & No & No & & & \\
%  & & & CNN (Gen) & N/A & Yes & No & & & \\
%  & & & GCN (Disc) & N/A & No & No & & & \\\hline

\multirow{3}{*}{\citet{ZShuFG2020}} & \multirow{3}{*}{x} & \multirow{3}{*}{x} & CNN & N/A & Yes & No & x & & \\
 & & & Encoder & 8 & Yes & No & & \multirow{2}{*}{x} & \multirow{2}{*}{x} \\
 & & & Decoder & 8 & Yes & No & & & \\\hline

\multirow{3}{*}{\citet{GHLeeCVPR2020}} & \multirow{3}{*}{x} & \multirow{3}{*}{x} & Encoder & 64 & Yes & No & \multirow{3}{*}{x} & & \multirow{3}{*}{x} \\
 & & & Decoder & 8 & Yes & No & & &  \\
 & & & Decoder & N/A & Yes & No & & &  \\\hline

\end{tabularx}

\begin{tablenotes}
\vspace{0.5em}
\item [1] \footnotesize{\citet{JLinCVPR2020} used a mixed generator composed of a CNN, a residual network and three GCN, two of them with skip connections.}
\item [2] \footnotesize{\citet{LattasCVPR2020} used as generator the pre-trained network from \citet{GecerCVPR2019}. They also adopted the GAN architecture from \citet{TCWangCVPR2018_pix2pixHD} that has three discriminators.}
\item [3] \footnotesize{\citet{PWangICME2020} trained two encoder-decoder networks separately for different tasks.}
\item [4] \footnotesize{The generator used by \citet{ZGaoCVPRW2020} is composed of a residual network, four parallel fully connected layers and three GCN, one of them with skip connections.}
\end{tablenotes}

\end{threeparttable}
\end{table*}

Tables \ref{tab:DL1}, \ref{tab:DL2}, and \ref{tab:DL3} summarise the main characteristics of each of the reviewed deep learning works according to the above items. The \emph{training set} column indicates whether the network was built by fitting a 3DMM to real images or by rendering synthetic images from a 3D face (or both). In the learning framework, four attributes are specified: the network type, the number of layers, whether skip connections are used, and whether the learning process is iterative. The number of layers is indicated as $A\times B$ if the network consists of $A$ networks of $B$ layers each, and $B_1//B_2//\cdots//B_n$ if there are $n$ networks arranged in parallel with $B_1, \dots, B_n$ layers respectively. Finally, the training criterion column indicates if the loss function is computed in a parameters space, the 3D space, and/or the 2D space.

%%%%%%%%%%%%%%%%%%%%%%%%%%%%%%%%%%%%%%%%%%%%%%%%%%%%%%%%%%%%%%%%%%%%%%%%%%%

\subsection{Training data set} \label{subsec:DL_training_set}
As we have mentioned above, the biggest obstacle when applying deep learning to 3D-from-2D face reconstruction is the lack of training data, since obtaining a huge number of 3D facial scans together with their corresponding 2D pictures required by deep learning algorithms is impractical. To overcome this limitation, researchers have proposed techniques for building synthetic training sets, taking advantage of pre-built 3DMMs to obtain 3D faces in a much more accessible way.

We can distinguish three main strategies to build synthetic training sets. We refer to the first one as \textit{Fit\&Render}, and it consists in fitting a 3DMM to real images and then rendering synthetic images using the estimated 3D faces. The second one, \textit{Generate\&Render}, is generating 3D faces by randomly sampling from a 3DMM and then, again, rendering synthetic images using the generated 3D faces. On the other hand, a new strategy is raising in the last few years that consists in self-supervised training, avoiding the need of paired 2D-3D data and thus the need of building synthetic datasets. The proposed approaches based on these three main strategies are reviewed in Sections \ref{subsubsec:DL_FitRender}, \ref{subsubsec:DL_GenerateRender} and \ref{subsubsec:DL_SelfSupervisionTS}, respectively. However, there have been other works that used synthetic training dataset built with both \textit{Fit\&Render} and \textit{Generate\&Render}, or that used real data. These works are summarised in Section \ref{subsubsec:DL_OtherTrainingDS}.

\subsubsection{\textit{Fit\&Render} strategy} \label{subsubsec:DL_FitRender}
As stated above, the \textit{Fit\&Render} strategy is based on fitting a 3DMM to real images. Then, to enlarge the variability of the set of 2D pictures corresponding to a 3D face, synthetic images are created using the estimated 3D face. Even though this strategy allows having realistic 2D images in the training set, which helps the network to perform better at test time, it is not without weaknesses. One drawback is that the accuracy of the trained deep learning method is highly determined by that of the 3DMM-fitting algorithm used to reconstruct the ground truth 3D faces. In other words, since the deep learning method is taught to reconstruct the estimated 3D faces, it will learn to reproduce the results obtained with the 3DMM-fitting algorithm. Also, one of the advantages of deep learning, which is its ability to learn non-linearities from the training data, is restricted by the linearly-modelled 3D data with which it is trained. Some works have tried to overcome these obstacles by, for example, including real datasets, reconstructing from multiple images, or by refining the coarse 3D facial reconstruction.

The \textit{Fit\&Render} strategy was first introduced by \citet{XZhuCVPR2016}. They proposed a face profiling technique that is used to generate images across larger poses, creating the 300W-LP (300W large poses) database. They first estimated a 3D mesh over the given face image by fitting a 3DMM using \citep{RomdhaniCVPR2005} and \citep{XZhuCVPR2015} over the background. Then, the 3D mesh was rotated and projected to the image to generate a synthetic image similar to the original one but with a larger pose. This 300W-LP database has been used by many other authors \citep{BhagavatulaICCV2017,YFengECCV2018,GulerCVPR2017,JacksonICCV2017,LTranCVPR2018,LTranPAMI2019,LTranCVPR2019,HYiCVPR2019,XZhuPAMI2019,GZhangFG2018,JGuoECCV2020,JShangECCV2020,XTuToM2020,ZGaoCVPRW2020,ZShuFG2020,GHLeeCVPR2020} since it includes realistic and challenging facial images with the ground truth 3DMM and projection parameters. \citet{GalteriCVPRW2019,GalteriCVIU2019} also followed the \textit{Fit\&Render} strategy to augment the FRGC dataset \citep{PhillipsCVPR2005_FRGC} by generating images with novel poses similarly to \citet{XZhuCVPR2016}.

\citet{YGuoPAMI2018} noticed the shortcomings mentioned above and proposed a pipeline for constructing a training set with detailed 3D faces and photo-realistic 2D images. To do so, they estimated per-vertex displacements from the coarse facial shape obtained with a 3DMM-fitting method, and then blended the albedo estimated by the 3DMM so as to obtain a rendered image as close to the original image as possible. Once all the rendering components are estimated, they varied them to generate more realistic 2D images.

Closely related to the \textit{Fit\&Render} strategy, some other works proposed to fit a 3DMM to images without rendering afterwards \citep{JourablooCVPR2016,JourablooIJCV2017, JourablooICCV2017,FLiuCVPR2018,TranCVPR2017,TranCVPR2018,JSYoonCVPR2019,ChinaevECCVW2018,ChaudhuriCVPR2019,XLiICASSP2020}. \citet{JourablooCVPR2016,JourablooIJCV2017}, \citet{JourablooICCV2017}, \citet{ChinaevECCVW2018}, \citet{ChaudhuriCVPR2019} and \citet{XLiICASSP2020} estimated the 3DMM parameters independently for each training image, whereas \citet{FLiuCVPR2018} and \citet{TranCVPR2017,TranCVPR2018} combined parameters extracted from multiple images of the same person. \citet{FLiuCVPR2018} fitted a 3DMM to multiple images enforcing common shape parameters to all the images of the same subject and estimating independently for each image the shape deformation due to expressions. In contrast, \citet{TranCVPR2017,TranCVPR2018} computed a weighted average of the shape and texture parameters estimated from the pictures of the same person. The resulting parameters were considered the ground truth for all the images of that subject. Differently from all the other works, \citet{JSYoonCVPR2019} used the non-linear face model created by \citep{LombardiToG2018_DAM}, who used an autoencoder network to learn a non-linear representation of textured 3D faces, and took as training set the images fitted by \citep{LombardiToG2018_DAM} in their training phase. This training dataset consists of 2D videos (i.e. consecutive 2D images) from different angles and the corresponding model parameters.

\subsubsection{\textit{Generate\&Render}}\label{subsubsec:DL_GenerateRender}

The \textit{Generate\&Render} strategy consists in sampling from a 3DMM to obtain ground truth 3D faces and then creating corresponding 2D images by rendering the 3D faces under different conditions (poses, lighting, etc.). This strategy avoids using an auxiliary 3DMM-fitting algorithm, and thus the network's learning capability is not limited to that of the reconstruction algorithm. However, unlike in the \textit{Fit\&Render} strategy, the 2D images are not realistic because the rendering process is fully synthetic, with synthetic backgrounds, lighting conditions, projection parameters, etc. Also, this strategy does not overcome the shortcoming of learning from linearly-modelled data, since still a 3DMM is used to create the ground truth 3D faces. Aware of these drawbacks, the works that follow the \textit{Generate\&Render} strategy propose different approaches to overcome them, such as including real data, adding synthetic deformations to the 3D faces, or adopting more complex training frameworks.

The \textit{Generate\&Render strategy} strategy was used by \citet{Richardson3DV2016}, proposing a weak perspective projection and the Phong reflectance model \citep{Phong1975} to render the synthetic images. This approach was strictly followed by some \citep{DouCVPR2017,RichardsonCVPR2017,SelaICCV2017}. \citet{RichardsonCVPR2017} and \citet{SelaICCV2017} trained their networks using only the dataset generated by \citep{Richardson3DV2016}, unlike \citet{DouCVPR2017}, who also used publicly available 3D face databases, namely, the FRGC2 database \citep{PhillipsPAMI2009_FRGC2_DB}, the BU-3DFE database \citep{YinFG2006_BU-3DFE_DB}, and the UHDB31 data\-base \citep{WuISBA2016_UHDB31_DB}.

Similarly to \citet{Richardson3DV2016}, \citet{JPiaoICCV2019} followed the \textit{Generate\&Render} strategy to build their training set. However, they added small free-form deformations to the nose and chin of some of the generated 3D faces. According to the authors, this allowed them to synthesise more realistic facial shapes, since real faces may not be completely captured by a linear model.

\citet{SenguptaCVPR2018} and \citet{GenovaCVPR2018} trained their networks in a two stage manner. \citet{SenguptaCVPR2018}, in the first stage, trained a simple network over synthetic data extracted using the \textit{Generate\&Render} technique. In the second stage, they used the pre-trained simple network to obtain normal, albedo and lighting estimates for a set of real images. The training data set was composed of synthetic 3D-2D data and 2D real images with the estimated normals, albedo and lighting, whose goal was to prevent the network to produce trivial solutions. In contrast, \citet{GenovaCVPR2018} trained the network with different training sets in each stage. In the first stage, they used synthetic 3D-2D data generated following the \textit{Generate\&Render} strategy, whereas in the second stage, they used only unlabelled images within an autoencoder architecture.

%in the first stage, pre-trained their face reconstruction network with synthetic 3D-2D data (generated following the \textit{Generate\&Render} strategy). In the second stage, they used unlabelled real images.

\subsubsection{Self-Supervision} \label{subsubsec:DL_SelfSupervisionTS}

Given the lack of real ground truth 3D-2D paired data and the drawbacks of training with synthetic sets, a new strategy -self-supervision- is gaining attention. The main idea is that the data itself provide the supervision by adding adding a rendering layer at the end of the network \citep{TewariICCV2017,TewariCVPR2018,TewariPAMI2020,LTranCVPR2018,LTranPAMI2019,LTranCVPR2019,YDengCVPRW2019,YZhouCVPR2019,SavovICCVW2019,FWuCVPR2019,JLinCVPR2020,ChaudhuriECCV2020,KoizumiECCV2020,PWangICME2020,TewariCVPR2019,XChaiICME2020,YChenTIP2020,GHLeeCVPR2020}. This rendering layers takes the textured 3D face and the rendering parameters estimated by the main network and renders a synthetic image. This enables to train the network end-to-end without the need of ground truth 3D faces by minimising the difference between the input and the rendered images. Some works \citep{TewariICCV2017,TewariCVPR2018,TewariPAMI2020,LTranCVPR2018,LTranPAMI2019,LTranCVPR2019,YDengCVPRW2019,YZhouCVPR2019,FWuCVPR2019,PWangICME2020,XChaiICME2020} trained their network with synthetically modified images to enlarge the variability in illumination, pose, etc., whereas others only used real images \citep{SavovICCVW2019,JLinCVPR2020,ChaudhuriECCV2020,KoizumiECCV2020,TewariCVPR2019,YChenTIP2020,GHLeeCVPR2020}.

In addition, \citet{YZhouCVPR2019} used a set of real 3D facial scans to train an autoencoder network, whose decoder is used in the image-to-mesh encoder-decoder network, and \citet{FWuCVPR2019} pre-trained their network in a fully supervised manner using the 300W-LP dataset generated by \citet{XZhuCVPR2016}. \citet{SanyalCVPR2019} and \citet{PWangICIP2019} also trained their network in a self-supervised manner, but instead of rendering a synthetic image, they projected only the landmarks position onto the image plane.

\citet{FLiuICCV2019} and \citet{XTuToM2020} used self-supervised training without a endering layer. \citet{FLiuICCV2019} trained a 3D-to-3D autoencoder network to use the decoder in the 3D face reconstruction process, similarly to \citep{YZhouCVPR2019}, but they also included synthetic 3D faces generated using a 3DMM to also train it in a supervised manner (having dense correspondence). On the other hand, \citet{XTuToM2020} combined images with ground truth 3DMM and pose parameters, with images without supervision. In this way, the network estimates parameters for the images without ground truth from the conditional distribution learnt from the images with ground truth.

\subsubsection{Others} \label{subsubsec:DL_OtherTrainingDS}
The two main strategies, \textit{Fit\-\&\-Ren\-der} and \textit{Ge\-ne\-ra\-te\-\&\-Ren\-der}, were combined by \citet{TrigeorgisCVPR2017}, \citet{XZengICCV2019} and \citet{KimCVPR2018}. \citet{TrigeorgisCVPR2017} constructed their training dataset using the ICT-3DRFE database \citep{StratouFG2011_3DRFE_DB} and the Photoface database \citep{ZafeiriouCVPR2011_PhotofaceDB} to generate synthetic images varying the illumination. They followed the \textit{Fit\&Render} strategy using the LSFM \citep{BoothIJCV2017_LSFM} as 3DMM and adding expressions with the FaceWarehouse model \citep{CaoVCG2014_FaceWarehouseFM}. With the same models, they randomly generated synthetic 3D faces, which were aligned to in-the-wild images to provide realistic backgrounds to the rendering. \citet{XZengICCV2019} built their training set in a similar way, using both strategies, although they also added images from the CACD dataset \citep{BCChenECCV2014_CACD_DB} without corresponding 3D faces to train their network in a self-supervised manner. In contrast, \citet{KimCVPR2018} built an initial training set using \textit{Generate\&Render}, which was used to pre-train the network. Then, during iterative training, they followed the \textit{Fit\&Render} strategy using the pre-trained network to estimate the model parameters from real images. Gaussian noise was added to these estimated parameters before rendering synthetic images, which were used to train the network in the next iteration.

Although most of the works built training sets following one of the three main strategies, there are others that followed none of them. \citet{SongTIP2012},  \citet{BaiCVPR2020}, \citet{XFanToM2020}, \citet{RamonICCVW2019}, \citet{LattasCVPR2020}, \citet{XWangCVPR2020} and \citet{XZhuECCV2020} trained their network using only real data. The first authors obtained the 2D-3D pairs from the BU-3DFE database \citep{YinFG2006_BU-3DFE_DB}, whereas \citet{BaiCVPR2020}, and \citet{XFanToM2020} used the Stirling/ESRC 3D face database and the FRGC v2.0 database, respectively. The first authors enlarged the training set by rendering 2D images from the textured 3D scans, whereas \citep{XFanToM2020} fitted a 3DMM to 2D images and then non-rigidly registered them to the triangulated depth images to obtain ground truth 3D faces with the same triangulation. In a similar way, \citet{XZhuECCV2020} also fitted a 3DMM but to RGB-D images and the non-rigid registration was conducted by finding correspondences based on geometry and texture. Differently, \citet{RamonICCVW2019}, \citet{LattasCVPR2020}, and \citet{XWangCVPR2020} built the training set themselves. The first \citep{RamonICCVW2019} captured 2D images and 3D faces from more than $6,000$ subjects, and the second \citep{LattasCVPR2020} captured 3D faces and reflectance maps of over $200$ individuals under 7 different expressions. Since \citet{XWangCVPR2020} estimated the 3DMM parameters, they constructed a dataset by refining, with known photometric parameters, the coarse shape extracted by fitting a 3DMM to three images captured under different lighting conditions. In addition, they enlarged the training set by following the \emph{Generate\&Render} approach but transferring the albedo from the captured images. They also rendered synthetic images from 3D faces obtained by non-rigidly registering the mean of the 3DMM to the 3D faces in the Light Stage \citep{WCMa2007_LightStage} database and, again, transferring the albedos from the captured images.

%%%%%%%%%%%%%%%%%%%%%%%%%%%%%%%%%%%%%%%%%%%%%%%%%%%%%%%%%%%%%%%%%%%%%%%%%%%

\subsection{Learning framework}
The core part of a deep leaning algorithm is the network itself, in terms of its architecture and how it learns its weights. The most straightforward learning framework is a single neural network that is trained in a single pass (Section \ref{subsubsec:DL_SingleNet}). Alternatively, some authors trained their networks in an iterative manner (Section \ref{subsubsec:DL_Iterative}) and/or exploited the potential of more complex architectures composed of multiple networks, such as encoder-decoder architectures (Section \ref{subsubsec:DL_EncDec}) or generative adversarial networks (Section \ref{subsubsec:DL_GANs}). Also, some researchers trained each of the multiple networks to perform specific sub-tasks (Section \ref{subsubsec:DL_SpecificTasks}).

Even though the most widespread way of representing 3D facial data is with a triangular mesh, most works rely on other representations, such as depth maps and 3DMM parameters. This is due to the inability of the classical 2D convolution-based networks to process non-Euclidean data, such as meshes. Nevertheless, a recent research field named geometric deep learning studies how to extend convolutional networks to non-Euclidean inputs, allowing to directly dealing with 3D facial meshes. In Section \ref{subsubsec:DL_GCN}, we include a brief technical background on this field and summarise how it has been used to solve the 3D-from-2D face reconstruction problem.

\subsubsection{Single Network with Single-Pass Training} \label{subsubsec:DL_SingleNet}

Convolutional neural networks (CNNs) have shown impressive results when dealing with 2D images, encouraging researchers to use CNNs to reconstruct the 3D face from uncalibrated 2D images \citep{ChinaevECCVW2018,JGuoECCV2020,PWangICIP2019,ChaudhuriCVPR2019,RamonICCVW2019,TewariICCV2017,TewariCVPR2018,TewariPAMI2020,SavovICCVW2019,PWangICIP2019,FWuCVPR2019}.

\citet{TewariICCV2017,TewariCVPR2018,TewariPAMI2020} and \citet{SavovICCVW2019} trained a CNN based on the AlexNet \citep{KrizhevskyNIPS2012_AlexNet} to regress the 3DMM parameters (shape, expression, and texture) together with the rendering parameters (projection and illumination parameters) from a single image. In Tewari's et al. early work \citep{TewariICCV2017}, the resulting reconstructions were coarse, and facial details were not captured. Later, \citet{TewariCVPR2018,TewariPAMI2020} refined the coarse face resulting from the AlexNet by estimating displacements for each vertex as coefficients from a generative model that is learnt directly from the training data. Besides, \citet{SavovICCVW2019} used the 3D face reconstruction to obtain a robust age estimation adding another AlexNet to regress the age of the person. They employed a soft-weights sharing strategy \citep{MisraCVPR2016} between both networks, which consists in adding layers that linearly combine activations from analogous layers from both networks. In the training stage, the weights of the linking layers are learnt jointly to the rest of both AlexNets, and thus the network learns which weights to share.

In contrast, \citet{FWuCVPR2019} and \citet{RamonICCVW2019} used the CNN to extract image features that were then processed by several parallel layers to regress the 3DMM and projection parameters separately. They used the VGGNet \citep{SimonyanICLR2015_VGG2} as feature extractor, which is a deeper network than the AlexNet \citep{KrizhevskyNIPS2012_AlexNet} and thus is able to extract more meaningful features, but it is also slower. In addition, both works considered multiple images as input with the aim of obtaining more faithful 3D faces, given that multi-view reconstruction is more accurate than single-view reconstruction. \citet{FWuCVPR2019} extracted features from three images of the same subject (frontal, left, and right views) and processed them by two separated fully-connected layers: one of them regressed the projection parameters for each image, and the other one regressed common 3DMM shape and expression parameters of the subject. \citet{RamonICCVW2019} proposed a very similar approach, however, they aimed at building a network that could reconstruct a 3D face from both single and multiple images. Therefore, instead of regressing a single shape from the VGG-features extracted from the input images, they estimated all parameters individually for each of the input images using a 3-layer perceptron as the regressor. Then, the resulting shape parameters are combined with another multilayer perceptron to output a single shape.

Similarly to \citep{TewariICCV2017,TewariCVPR2018,TewariPAMI2020,SavovICCVW2019}, \citet{ChinaevECCVW2018}, \citet{JGuoECCV2020} and \citet{PWangICIP2019} directly regressed all the parameters (3DMM, rendering and projection) with a CNN. However, they used the MobileNet \citep{HowardArXiv2017_MobileNets}  which, according to all the authors, is a fast and compact network, and therefore allows for real-time processing of the input images, even on mobile devices. Whereas \citet{ChinaevECCVW2018} only estimated the above mentioned parameters, \citep{JGuoECCV2020} and \citep{PWangICIP2019} noticed that it was not sufficient and tried to improve the accuracy obtained with the MobileNet by estimating additional features. \citet{JGuoECCV2020} found that estimating the 2D landmarks locations as a separated regression task, instead of using the projection of the estimated 3D landmarks, improved the results. On the other hand, \citet{PWangICIP2019} compensated the limitations of the 3DMM by training the MobileNet to also output middle-level corrections, which were represented as a 3D deformation field modelled by linear manifold harmonics basis functions \citep{ValletCGF2008}. These corrections allowed them to obtain lower error than \citep{JGuoECCV2020} in the landmarks on the AFLW2000-3D dataset \citep{XZhuCVPR2016}.

% \citet{ChinaevECCVW2018}, \citet{JGuoECCV2020} and \citet{PWangICIP2019} used the MobileNet \citep{HowardArXiv2017_MobileNets} to estimate the parameters of a 3DMM, the rendering and projection parameters. According to all the authors, the MobileNet is a fast and compact network that allows for real-time processing of the input images even on mobile devices. Unlike \citep{ChinaevECCVW2018}, \citep{JGuoECCV2020} and \citep{PWangICIP2019} tried to improve the accuracy obtained with the MobileNet. \citet{JGuoECCV2020} regressed the 2D landmarks locations separately to the above mentioned parameters (with two separated fully-connected layers), which served as extra regularisation for the reconstruction process. Instead, \citet{PWangICIP2019} trained the MobileNet to also output middle-level corrections. These corrections are represented as a deformation 3D field modelled by linear manifold harmonics basis functions \citep{ValletCGF2008}.

With the same idea of real-time applications, \citet{ChaudhuriCVPR2019} built a CNN with Fire Modules from the SqueezeNet \citep{IandolaArXiv2016_Squeezenet} and squeeze-and-excite modules \citep{JHuCVPR2018_SqueezeExciteNets}, which were designed with the aim of reducing the model size and complexity without compromising the accuracy of the results. With this architecture, they achieved comparable results to those of \citet{PWangICIP2019} on the AFLW2000-3D dataset \citep{XZhuCVPR2016}.

Differently from the above works, \citep{JourablooICCV2017,TranCVPR2017,YDengCVPRW2019,JShangECCV2020,TrigeorgisCVPR2017} used residual CNNs. These architectures make it possible to train very deep CNNs by adding skip connections (connections between non-consecutive layers). This increase in the complexity of the model, with respect to non-residual CNNs, allows learning more representative features, although it also requires much more time to train.

\citet{JourablooICCV2017} aiming a a better pose estimation for face alignment, added two fully-connected layers at the end of each residual block to estimate a visualisation mask, which measured how much each surface normal was pointing towards the camera. This visualisation mask was passed to the next residual block through an additional input channel so that the next block can correct the errors of the previous one.

On the other hand, \citep{TrigeorgisCVPR2017,TranCVPR2017,YDengCVPRW2019,JShangECCV2020} used the ResNet architecture \citep{HeCVPR2016_ResNet} without modifying it, the first one with 101 layers and the latter three with 50 layers. The work presented by \citet{TrigeorgisCVPR2017} was based on the classical shape-from-shading, where the facial normals are estimated by assuming, generally, a Lambertian reflectance model (eq. \eqref{eq:LambertianModel_SH}). They estimated the facial normals with the ResNet and then retrieved the 3D facial surface using \citep{FrankotPAMI1988}.

\citet{TranCVPR2017}, \citet{YDengCVPRW2019} and \citet{JShangECCV2020} exploited multiple views similarly to \citep{FWuCVPR2019,RamonICCVW2019}. The former aimed at face recognition and thus trained the ResNet to be discriminative by using a training set with the same 3D face corresponding to several images of the same subject. \citet{YDengCVPRW2019} recovered the final reconstruction by linearly combining the single-image reconstructions according to confidence scores estimated by a second network. This way, a more accurate reconstruction contributed more to the final reconstruction, which resulted in better results compared with averaging the shapes. Different from \citep{YDengCVPRW2019}, \citet{JShangECCV2020} exploited the multiple views to further drive the reconstruction process. They reconstructed the 3D face corresponding to a target image from two adjacent views by imposing multi-view consistency, which allowed them to tune the final reconstruction using the three images at the same time. However, this approach forces the authors to train their network with adjacent images, whereas \citet{YDengCVPRW2019} could train their networks with ``non-related'' set of images.

\subsubsection{Iterative Training} \label{subsubsec:DL_Iterative}
In contrast to the single pass training adopted by the works in the previous section, other authors proposed to train their networks in an iterative manner. We can distinguish two different approaches: one is based on iteratively improving the synthetic training set, and the other on iteratively refining the result of the previous iteration.

\citet{KimCVPR2018} followed the first approach. They updated the training set in each iteration by adding synthetic images that were rendered using parameters (after adding a Gaussian noise) estimated from real images by the train\-ed network up to the cur\-rent iteration. Thus, in each iteration, the network was trained over a training set generated by the ``previous'' network. In this way, the training data distribution is more similar to the real distribution, alleviating the bias effect of the synthetic data in the training process. 

The second approach follows a similar idea to that of a cascaded regressor: each regressor estimates an update of the input parameters, which are estimated by the previous regressor, such that they are closer to the ground truth. Most of the proposed works used the same architecture for all the iterations \citep{Richardson3DV2016,SanyalCVPR2019,XZhuCVPR2016,XZhuPAMI2019,GZhangFG2018,BaiCVPR2020}, but \citep{JourablooCVPR2016,JourablooIJCV2017} alternated regressors that estimated updates for different parameters.

\citet{Richardson3DV2016} and \citet{SanyalCVPR2019} proposed a network based on the ResNet architecture \citep{HeCVPR2016_ResNet}. The former trained the ResNet to estimate the 3DMM parameters (the pose is pre-computed using \citep{KazemiCVPR2014}) and then refined the resulting 3D face by reconstructing fine details using \citep{OrElCVPR2015}. In contrast, \citet{SanyalCVPR2019} estimated the pose together with the 3DMM parameters. They trained their network in a multi-view manner by imposing the distance between shape parameters of different subjects to be larger than the distance between parameters of the same person.

\citet{XZhuCVPR2016,XZhuPAMI2019} and \citet{GZhangFG2018} focused their work on large-pose face alignment, i.e. the detection of facial landmarks, by 3D face reconstruction. \citet{XZhuCVPR2016,XZhuPAMI2019} trained iteratively a 6-layers CNN to output the update of the pose, shape and expression parameters. In their early work \citep{XZhuCVPR2016}, they fed the network with the 2D image and a feature representing the current estimated 3D face, namely, the projected normalised coordinate code. This feature is computed as the projected 3D face with the normalised coordinates (to $[0,1]$) as its colourmap, and allowed them to 1) pass to the next iteration the estimation of the previous one, 2) reflect the fitting accuracy of the current estimate, and 3) apply the convolution operation. In their extended work \citep{XZhuPAMI2019}, they proposed a two-stream network whose outputs are combined by a fully-connected layer. One of the streams is that of \citep{XZhuCVPR2016} and the other one inputs the location of the projection of sampled 3D points. This architecture allowed them to separately process two different sources of information that complement each other, thus improving the results from \citep{XZhuCVPR2016}. In contrast, \citet{GZhangFG2018} used as additional input the patches of the image centred at the projection of the 3D landmarks. They highlighted that these features are much faster than the projected normalised coordinate code proposed by \citep{XZhuCVPR2016,XZhuPAMI2019} or the projection of 3D points used by \citep{JourablooCVPR2016}.

\citet{BaiCVPR2020} estimated an update of the previously regressed 3D face, instead of estimating the parameters update. The authors argued that existing 3DMMs are limited in their representation power and thus proposed to learn an \emph{adaptive basis} $\vec{\Phi}_\text{adap}$ at each iteration to model the updates. Then, the final reconstruction is computed as $\vec{x} = \overline{\vec{x}} + \vec{\Phi}\vec{\alpha} + \sum_{i = 1}^3\vec{\Phi}^i_\text{adap}\vec{\alpha}^i_\text{adap}$.

Differently from all the above works, \citet{JourablooCVPR2016,JourablooIJCV2017} alternated two different regressors: one that estimated the 3DMM parameters updates, and another that estimated the update of the projection parameters. Whereas in \citep{JourablooCVPR2016} they used a single CNN, in \citep{JourablooIJCV2017} they used a Siamese network \citep{BromleyNIPS1993_SiameseNet}, which has two branches with shared weights to process at the same time the input image and its mirrored image. With this architecture they could enforce the update of the 2D landmarks (projection of the 3D landmarks) for the input image and the mirrored image to be similar, which further constrains the face alignment problem, leading to better reconstructions.
% also focused on large-pose face alignment, however, instead of using a single network to estimate all the parameters update like in \citep{GZhangFG2018,XZhuCVPR2016,XZhuPAMI2019},

\subsubsection{Encoder-Decoder} \label{subsubsec:DL_EncDec}
The encoder-decoder architecture partitions the network in two parts: the encoder, which extracts representative features of reduced dimensionality from the input face image, and the decoder, which estimates a reconstruction of the input data from the features estimated by the encoder. This reconstruction may be a 3D face whose geometry corresponds to that of the face from the input image, parameters of a facial model with which the 3D face can be recovered, or a synthetic 2D image resembling the input one, among others.

Although all the works presented in this section used an encoder-decoder architecture to recover the 3D facial geometry from 2D images, we have categorised them in 3 groups according to how they apply this architecture. Firstly, there are works \citep{SelaICCV2017,YFengECCV2018,XZhuECCV2020,JacksonICCV2017,HYiCVPR2019,KoizumiECCV2020} that considered the encoder and the decoder as a single network, without paying attention to the output of the encoder. Others, in a second category, \citep{SongTIP2012,ChaudhuriECCV2020,JSYoonCVPR2019,GenovaCVPR2018,YZhouCVPR2019,GHLeeCVPR2020} used the encoder to extract meaningful image features from the 2D picture that help the decoder to estimate the 3D representation. And, finally, the third category consists of works \citep{LTranCVPR2018,LTranPAMI2019,LTranCVPR2019,XLiICASSP2020,FLiuCVPR2018,FLiuICCV2019} in which the encoder was trained to estimate descriptive parameters, such as parameters related to shape (identity), expression, texture, etc., which were then used by the decoder to estimate the 3D face.

Within the first category we find the works by \citet{SelaICCV2017} and \citet{KoizumiECCV2020} who adopted the U-Net \citep{RonnebergerMICCAI2015_UNet} with skip connections between corresponding layers in the encoder and the decoder. Both works used this network to estimate a depth map and a correspondence map from each pixel of the image to a vertex of a reference mesh. \citet{SelaICCV2017} used as reference a template mesh, which they non-rigidly registered to the estimated depth map using the correspondence map. They refined the resulting coarse mesh by estimating a displacement map from local variations in the image colours \citep{BeelerToG2010}. With this method, they obtained very detail 3D faces, although could not cope with low resolution images. On the other hand, \citet{KoizumiECCV2020} used the mean shape of a 3DMM as reference mesh, thus, with the depth and correspondence maps, they solved a least squares problem to estimate the 3DMM shape and texture parameters and the rendering parameters. Unlike \citep{SelaICCV2017}, they trained the network end-to-end by adding the least squares layer and a rendering layer (see Section \ref{subsec:DL_training_set}), which allowed unsupervised learning.

Differently from the other works in the first category, \citet{JacksonICCV2017} stacked two hourglass networks \citep{NewellECCV2016_HourglassNet} to refine the output of the first network with the second one. The hourglass network is an encoder-decoder that uses residual modules and skip connections between corresponding layers in the encoder and the decoder, similarly to the U-Net \citep{RonnebergerMICCAI2015_UNet}. However, the skip connections in the hourglass network further process the information with a residual module, which allows capturing information at every scale. With this architecture, \citet{JacksonICCV2017} reconstructed a volumentric representation of the face that is defined as a 3D binary discrete volume. According to the authors, this representation simplifies the learning since it can be treated as the probability of a voxel containing part of the face, and the training 3D faces do not need to have a fixed number of vertices. However, the resulting reconstructions are of low resolution. In the same line, \citet{HYiCVPR2019} used \citep{JacksonICCV2017} to obtain a first estimation of the 3D facial geometry, but constrained it using a 3DMM. More specifically, they estimated the projection parameters and the 3DMM shape and expression parameters from the volumetric representation using a CNN. Since both networks were jointly trained, the entire framework benefited from purely geometric supervision and from the 3DMM.

Still within the first category, \citet{YFengECCV2018} proposed a different 3D representation: they estimated a UV position map containing the XYZ coordinates of the estimated shape in the UV space. This representation allowed them to infer the invisible parts of the face, thus directly predicting the complete 3D face. However, unlike the depth map \citep{SelaICCV2017,KoizumiECCV2020} or the volumetric representation \citep{JacksonICCV2017,HYiCVPR2019}, the UV coordinates in \citep{YFengECCV2018} need to be based on a 3DMM to maintain the semantic meaning of points within the UV map. This work was extended by \citet{XZhuECCV2020}, who employed the 3D representation and the network from \citep{YFengECCV2018} to estimate a shape update from an initial shape estimation $\vec{x}^0$ obtained with \citep{XZhuPAMI2019}. More specifically, they input the vertex positions and the texture of $\vec{x}^0$ as UV-maps (like \citep{YFengECCV2018}). The texture UV-map is weighted by an attention map extracted from the visibility of the vertices in the input image and the final reconstruction is obtained by adding the estimated update to $\vec{x}_0$. This approach allowed \citet{XZhuECCV2020} to obtain more detailed 3D faces than \citep{XZhuPAMI2019} by estimating a shape update, and more accurate results than \citep{YFengECCV2018} by constraining the coarse shape using the 3DMM-based approach from \citep{XZhuPAMI2019}.

As stated above, a second category of works used the encoder as a complex feature extractor. \citet{GenovaCVPR2018} used the FaceNet network \citep{SchroffCVPR2015_FaceNet} and passed its output to a decoder that estimated the 3DMM parameters. Their core contribution was a training method that avoided the need of 2D-3D data pairs and, unlike many other works, did not rely on a rendering layer to minimise the difference between the synthetic and original image (see Section \ref{subsubsec:DL_ParamsLoss} for more details). \citet{JSYoonCVPR2019} used the VGGNet \citep{SimonyanICLR2015_VGG2} and the hourglass network \citep{NewellECCV2016_HourglassNet} to extract features from the input image, but, unlike \citep{GenovaCVPR2018}, they used the features extracted by intermediate layers from both encoders. They concatenated these features and processed them by two separate networks \citep{JSYoonICCV2017} that estimated separately the pose parameters and the parameters of the deep appearance face model of \citet{LombardiToG2018_DAM}. This model was constructed as a 3D autoencoder, thus, to recover the textured 3D face from the model parameters, they used the pre-trained decoder from \citet{LombardiToG2018_DAM}. However, this method requires a pre-built person-specific deep appearance face model to be used as input.

Unlike the above works that focus on estimating parameters of a pre-built model, \citet{ChaudhuriECCV2020} approached the 3D face reconstruction problem as the construction of personalised models. To do so, they estimated shape and albedo corrections from a pre-built model taking as input a collection of frames from a video of a subject. They jointly trained a residual network that estimated the pose, expression, and illumination coefficients for each of the input images, and an encoder-decoder that estimated shape and albedo corrections with respect to the input model. The encoder extracted features from the frames, and two separate decoders processed them to estimate the shape and the albedo corrections, respectively. In this way, they could alleviate the constraints of pre-build facial models by estimating personalised models with more accurate basis, and thus more accurate 3D faces.

% Unlike the above works that estimate model parameters, \citet{ChaudhuriECCV2020} aimed at constructing personalised models. To do so, they estimated shape and albedo corrections from a pre-built model taking as input a collection of frames from a video of a subject. They jointly trained a residual network that estimated the pose, expression, and illumination coefficients for each of the input images, and an encoder-decoder that estimated shape and albedo corrections with respect to the input model. The encoder extracted features from the frames, and two separate decoders processed them to estimate the shape and the albedo corrections, respectively.

\citet{SongTIP2012} and \citet{YZhouCVPR2019} used the 3D face as input to drive the reconstruction process. The former \citep{SongTIP2012} trained two radial basis function networks to extract the same intrinsic features from the 2D image and the 3D geometry, respectively. In this way, a 3D face can be reconstructed from a 2D image by estimating the features with the \textit{2D-network} (encoder) and processing them by the inverted \textit{3D-network} (decoder). In contrast, \citet{YZhouCVPR2019} jointly trained an image-to-mesh encoder-decoder and a 3D autoencoder, which share the decoder, using recent graph convolutional networks (GCNs) (see Section \ref{subsubsec:DL_GCN}). This allowed them to jointly train the decoder 1) in a fully supervised manner for the mesh-to-mesh stream; and 2) in a self-supervised manner for the image-to-mesh stream by adding a final rendering layer that creates a synthetic image from the resulting 3D face.

\citet{GHLeeCVPR2020} also used a GCN to directly regress a 3D facial mesh from image features (see Section \ref{subsubsec:DL_GCN}), although their main contribution is the non-deterministic feature-extractor. They argued that the 3D face should be reconstructed with high specificity from confident features and with high generalisation ability from uncertain features. Therefore, to account for such objectives, they pre-trained a network to estimate the uncertainty of a feature by maximising the log-likelihood of features extracted from images of the same subject. Then, the estimated feature distributions were passed to a GCN to estimate the shape and to the network proposed by \citep{BrockICLR2019} to estimate a UV texture map. According to the authors, adding uncertainty to the image feature acts as a regulariser, controlling generality for global shape and specificity for details.

Finally, the third category of works trained their encoder to estimate descriptive parameters. Most of them \citep{FLiuCVPR2018,FLiuICCV2019,LTranCVPR2018,LTranPAMI2019,LTranCVPR2019} followed a similar approach: they extracted identity and non-identity features from the input image with a single encoder, and decoded each of them with separate networks. \citet{FLiuCVPR2018} trained the decoders to output identity-related $\Delta \vec{x}_\text{id}$ and expression-related $\Delta \vec{x}_{\text{exp}}$ updates from a template 3D face $\vec{x}_\text{T}$, thus reconstructing the final 3D face as $\vec{x} = \vec{x}_\text{T} + \Delta \vec{x}_\text{id} + \Delta \vec{x}_{\text{exp}}$. In a subsequent work, \citet{FLiuICCV2019} followed a similar idea to that of \citep{YZhouCVPR2019} and used pre-trained decoders. These decoders were trained within a mesh-to-mesh autoencoder architecture, which can be used as a dense correspondence network, in a supervised manner. In contrast, \citet{LTranCVPR2018} trained the encoder to estimate shape (including identity and expression), texture and projection parameters, and directly regressed the 3D face and the texture with the decoders. In their subsequent works \citep{LTranPAMI2019,LTranCVPR2019}, they estimated separately the lighting from the albedo, instead of estimating jointly all the texture-related features, and revised the regularisation scheme (see Section \ref{subsubsec:DL_SelfSupervisedTraining}) to obtain more plausible and detailed textured 3D faces.

Differently from the above works, \citet{XLiICASSP2020} argued that expression and identity features extraction are two separate tasks and thus cannot be carried out with the same encoder. Therefore, they trained two parallel encoder-decoder networks in an end-to-end manner to estimate separately identity and expression features by taking global and local perspectives, respectively. To do so, the identity network inputs the cropped face from the input image, whereas the expression network inputs different local regions of the input facial image, both detected using \citep{KZhangSPL2016}. According to the authors, decomposing identity and expression attributes, allowed them to reconstruct facial details and micro-expressions that are not captured by other works, obtaining much lower reconstruction errors.

% Like \citep{FLiuCVPR2018,FLiuICCV2019}, \citet{XLiICASSP2020} estimated identity and expression deformations with respect to a template mesh.

\subsubsection{Generative Adversarial Networks} \label{subsubsec:DL_GANs}
Generative adversarial networks (GANs) allow learning the distribution of the ground truth 3D faces by jointly training two separate networks, the generator and the discriminator, to compete with each other. The generator $G: \mathcal{X} \to \mathcal{Y}$ maps the input data $\mathbf{x}\in\mathcal{X}$ to a (possibly) different space $\mathbf{y}\in\mathcal{Y}$, $G(\mathbf{x}) = \mathbf{y}$. The discriminator estimates the probability of a data point $\mathbf{y}\in\mathcal{Y}$ being ground truth data and not being synthetically generated by $G$. In this framework, the discriminator is trained to correctly distinguish between real and synthetic data, whereas the generator is trained to fool the discriminator by generating data resembling real data. This approach has been used to refine a coarse 3D face \citep{JLinCVPR2020,GalteriCVPRW2019,GalteriCVIU2019} and to obtain more accurate 3D faces \citep{XTuToM2020,ZGaoCVPRW2020,JPiaoICCV2019}. 

\citet{JLinCVPR2020} and \citet{GalteriCVPRW2019,GalteriCVIU2019} pointed out the limited power of the 3DMMs to recover fine details. Whereas \citet{JLinCVPR2020} focused on refining the texture obtained by a 3DMM-based reconstruction network \citep{YDengCVPRW2019}, \citet{GalteriCVPRW2019,GalteriCVIU2019} aimed at recovering fine geometric details to refine 3D shapes obtained with \citep{FerrariToM2017,TranCVPR2017}.

Specifically, \citet{JLinCVPR2020} refined the per-vertex texture with a GCN \citep{HammondACHA2011_GCN_spectral,DefferrardNIPS2016_GCN_spectral} and extracted facial details from the input image using FaceNet \citep{SchroffCVPR2015_FaceNet}, which were decoded with another GCN. The information obtained with the GCNs was combined with a single graph convolutional layer to obtain a 3D face with high-fidelity texture.

On the other hand, \citet{GalteriCVPRW2019,GalteriCVIU2019} refined the coarse shape using a conditional GAN, taking the input coarse shape as a condition in the estimation of the probability. This conditional GAN was trained using the procedure proposed by \citet{KarrasICLR2018}, which is based on progressively and simultaneously growing the dimension of the generator and the discriminator. According to the authors, this way of training the GAN model allows learning finer details progressively as the resolution of the input increases. Also, since processing 3D meshes is harder than processing 2D images, \citet{GalteriCVPRW2019,GalteriCVIU2019} converted the 3D faces into 2D images by projecting them to the image plane but, instead of projecting the $(x,y,z)$ coordinates, they extracted 3 features from each vertex. \citet{GalteriCVPRW2019} used the depth (i.e. $z$ coordinate) and the azimuth and elevation angles of the normal vector. In a subsequent work, \citet{GalteriCVIU2019} experimentally found that the azimuth angle did not add relevant information, and thus replaced it by the mean curvature of the surface at the given vertex, which encodes the local variability of the face. However, since the coarse reconstruction process is not included in the framework, the architecture has to be trained from scratch if the input coarse face is changed.

Differently from the above works, \citet{XTuToM2020} and \citet{ZGaoCVPRW2020} used the GAN architecture to regress more realistic reconstructions: the former focused on regressing more accurate 3DMM parameters, whereas \citep{ZGaoCVPRW2020} used the discriminator to train a decoder network to output more realistic 3D facial geometries. Specifically, \citet{XTuToM2020} proposed a conditional GAN where the discriminator determines whether the estimated 3DMM parameters, given the input image, follow the conditional distribution learnt from images with ground truth parameters. In other words, the discriminator determines if the estimated parameters relate to the input image in the same way as ground truth parameters relate to their corresponding 2D images. In this way, according to the authors, the generator is trained to regress more precise and robust parameters. However, this approach cannot recover facial deformations not captured by the 3DMM, and thus, the reconstruction is limited by the facial model obtaining facial surfaces that are excessively smooth to be realistic.

\citet{ZGaoCVPRW2020} proposed an encoder-decoder architecture very similar to that of \citep{LTranCVPR2019,LTranPAMI2019}, where the encoder learns the pose and lighting parameters and identity- and expression-related features. Unlike \citet{XTuToM2020}, the identity and expression are not modelled by a 3DMM, but decoded by 2 GCNs: one that estimated the 3D face from both features and the other one that used only the identity to estimate the albedo. As mentioned above, \citet{ZGaoCVPRW2020} used the discriminator to force the ``shape decoder'' to produce faces that follow the distribution of real 3D faces. However, this distribution is learnt from 3D faces sampled from a 3DMM, which limits the reconstruction ability to that of the 3DMM, as in \citep{XTuToM2020}.

In contrast, \citet{JPiaoICCV2019} argued that, on the one hand, training the network directly on 2D-3D synthetic pairs hinders dealing with real images. On the other hand, training the network in a self-supervised manner is likely to overfit. To overcome this, \citet{JPiaoICCV2019} used a cycle-consistent GAN (CycleGAN) \citep{JYZhuICCV2017} to transform realistic 2D images to synthetic-style images. The CycleGAN consists of two GANs, $\{G_\text{r},D_\text{r}\}$ and $\{G_\text{s},D_\text{s}\}$, that are concatenated forming a cycle:
\begin{equation} \label{eq:cycleGAN}
    \mathbf{I}_\text{r} \xrightarrows[\textit{G}\textsubscript{s}][\textit{G}\textsubscript{r}] \mathbf{I}_\text{s}.
\end{equation}
$G_\text{r}: \mathcal{S} \to \mathcal{R}$ maps a synthetic image $\mathbf{I}_\text{s}$ to a real image $\mathbf{I}_\text{r}$, and $G_\text{s}:\mathcal{R}\to \mathcal{S}$ is its inverse. Then, an encoder-decoder network inputs the synthetic-style image and estimates position UV-maps of the 3D face, similarly to \citep{YFengECCV2018} (see Section \ref{subsubsec:DL_EncDec}). Therefore, the whole end-to-end trainable network is:
\begin{equation}
    \mathbf{I}_\text{r} \xrightarrows[\textit{G}\textsubscript{s}][\textit{G}\textsubscript{r}] \mathbf{I}_\text{s} \xrightarrow{\text{\normalsize Enc-Dec}} \text{Facial position UV-map}.
\end{equation}
According to the authors, mapping the realistic images to a synthetic-style facilitates the shape estimation process of the 3D face reconstruction network. However, they performed similarly to other state-of-the-art methods, and the reconstructions were not sufficiently detailed.

\subsubsection{Multiple Networks for Different Tasks} \label{subsubsec:DL_SpecificTasks}
In this section, we group the approaches that use multiple networks, where each one is trained for different tasks. These approaches can be divided into three main blocks: \emph{1)} multi-stage systems, \emph{2)} coarse-to-fine systems, and \emph{3)} networks with multiple-branches.

\emph{1)} In a multi-stage system, multiple networks are trained to estimate different parameters in a sequential manner. For example, \citet{SenguptaCVPR2018}, \citet{BhagavatulaICCV2017} and \citet{TewariCVPR2019} used a CNN to extract features from the input image(s) and considered multiple networks to estimate different parameters from the CNN's output. \citet{SenguptaCVPR2018} used two different residual blocks to learn to separate low and high frequency variations into normals and albedo. Then, the output of the two residual blocks and the input image were jointly used to learn the lighting parameters. \citet{BhagavatulaICCV2017} estimated the warping parameters from a template to the input image and the projection parameters using different fully connected networks. Since they aimed at face alignment, they refined the projection of the resulting 3D landmarks into the image with an additional network. On the other hand, \citet{TewariCVPR2019} used a Siamese CNN to extract features jointly from sets of four video frames. These features are fed into another CNN that averages them and estimates the shape and albedo parameters, common to all the frames. The estimate for each of the input frames is further refined by computing its own pose, illumination, and expression parameters, using yet another Siamese CNN. However, the main contribution of \citep{TewariCVPR2019} is that the shape and albedo models were learnt in the training stage from in-the-wild Internet videos of over 6,000 subjects, obtaining in this way facial models that generalise well. Then, in the test step, only the model parameters are regressed.

Unlike the above works, where each network regresses different features from the input image, \citet{ZShuFG2020} proposed a main network that estimated a depth map and two additional networks that helped regularise the output of the main network. Given that estimating depth from a single image is an ambiguous task, they used a pre-trained 3DMM-fitting network that helped constrain the space of possible solutions by taking depth maps similar to the 3D face obtained by the 3DMM. Also, aiming for a good generalisation to unseen real images, they trained the main network not only with labelled images (in which the depth map is known), but also with pairs of unlabelled images taken at the same time instance from different viewpoints. However, according to the authors, ensuring photo-consistency directly on the multi-view images is not possible since the same anatomical point may have different pixel colours in both images due to lighting variations. For this reason, they trained another network to predict the suitability of each pixel to be used in the multi-view photo-consistency loss, giving more weight to those pixels whose colours do not change much between the images.

\emph{2)} A coarse-to-fine system is typically composed by two networks connected in cascade and trained to estimate a coarse 3D face and the fine-level details, respectively. This approach was followed by \citep{XFanToM2020,RichardsonCVPR2017,YGuoPAMI2018,TranCVPR2018,XZengICCV2019,LattasCVPR2020,PWangICME2020,XWangCVPR2020,YChenTIP2020}. \citet{XFanToM2020} first estimated a personalised template and then refined it by estimating an update using a network based on the 50-layer ResNet \citep{HeCVPR2016_ResNet}. The personalised template is initially estimated by fitting a 3DMM to the input image and then each vertex is refined by different 1-layer multi-layer perceptron (MLP) networks. These MLPs take into consideration not only the vertex in question but also a neighbourhood, together with their colours extracted by projecting the template into the input image.

\citet{XWangCVPR2020}, \citet{RichardsonCVPR2017}, \citet{YGuoPAMI2018}, \citet{TranCVPR2018} and \citet{YChenTIP2020} estimated the coarse 3D face by regressing parameters of a 3DMM, and refined it by adopting different alternative representation of the coarse faces.

% The coarse networks of the first three works \citep{XWangCVPR2020,RichardsonCVPR2017,TranCVPR2018} were based in the ResNet network \citep{HeCVPR2016_ResNet}, whereas the latter \citep{YChenTIP2020} trained the VGG-Face \citep{ParkhiBMVC2015_VGGFace} architecture. Unlike the other works, \citet{RichardsonCVPR2017} estimated the 3DMM and rendering parameters in the fine network iteratively by adding a feedback channel.

For the fine network, \citet{XWangCVPR2020} converted the 3DMM parameters into a coarse normal map that was refined using the features extracted from it and from the input image(s). These two sets of features were decoded to obtain a refined normal map. In contrast, \citet{RichardsonCVPR2017} converted the 3DMM parameters into a depth map that was refined by a network based on the hypercolumn architecture \citep{HariharanCVPR2015_HypercolumnNet}. According to the authors, this architecture generates a pixel-wise feature map that incorporates structural and semantic data, and it is able to capture fine details in the input image. This work inspired \citet{YGuoPAMI2018} to apply the coarse-to-fine framework to videos instead of single images. To do so, they included a coarse tracking network that predicted the next frame's coarse parameters. On the other hand, \citet{TranCVPR2018} and \citet{YChenTIP2020} represented the fine details as a displacement map from the coarse face, and both used the image-translation network proposed by \citet{IsolaCVPR2017_Img2Img}. \citet{TranCVPR2018} trained their fine network with a supervised setting, thus, they required ground truth displacements maps, which they borrowed from \citet{OrElCVPR2015}. The drawback of this strategy is that reconstruction power of their fine network is inherently limited by the accuracy achievable with \citet{OrElCVPR2015}. To avoid this, \citet{YChenTIP2020} added a rendering layer to train the network in a self-supervised manner. More specifically, the fine network from \citep{YChenTIP2020} inputs two UV-texture maps, one extracted from the input image and the other extracted from the albedo estimated with the coarse network, thus obtaining a displacement map.

Differently from the above works, based on two networks, \citet{XZengICCV2019}, \citet{PWangICME2020} and \citet{LattasCVPR2020} stacked three different networks. \citet{XZengICCV2019} trained each of them to refine the depth map estimated by the previous network by taking into account also the input image. The first two networks were trained in a fully supervised manner and, thus, their performance was limited by the synthetically generated data. The third network was trained in a self-supervised manner and designed so that the input image is integrated not only in the input layer but also in several intermediate layers, which, according to the authors, helps to gradually correct the depth map locally.

Instead of refining directly the geometry, \citet{PWangICME2020} and \citet{LattasCVPR2020} reconstructed photo-realistic 3D faces by refining the texture estimated by a 3DMM fitting-based approach. 

\citet{PWangICME2020} used the first two networks to refine the 3DMM-fitted texture, inputed as a UV map, and the third network to refine the 3D geometry using the refined texture. Specifically, the first network separated the input texture into albedo and lighting, which were used to train the second network to translate the texture UV map into a refined albedo UV map. Finally, the third network estimated per-vertex displacements using the refined UV albedo map. Similarly, \citet{LattasCVPR2020} refined the output obtained with the 3DMM-fitting method presented by \citet{GecerCVPR2019} by inferring the diffuse and specular albedo and normals, which are then used to improve the reconstructed 3D face. First, the texture UV map estimated by \citep{GecerCVPR2019} is enhanced with a super resolution network \citep{YZhangECCV2018_Rcan}, then an image-to-image network \citep{TCWangCVPR2018_pix2pixHD} is trained to estimate the diffuse albedo by removing the illumination from the enhanced UV texture map. Finally, another 3 image-to-image networks \citep{TCWangCVPR2018_pix2pixHD} are trained to infer the specular albedo, and the specular and diffuse normals.

\emph{3)} Finally, the third block is composed of approaches whose architectures have multiple parallel networks that process the input image to estimate different attributes. For example, \citet{GulerCVPR2017} and \citet{DouCVPR2017} attached different branches to a main network where each of them estimates different parameters. \citet{GulerCVPR2017} proposed to add bran\-ches at the end of a ResNet-based network \citep{HeCVPR2016_ResNet} to separately estimate the U- and V-coordina\-tes of an UV-map. In contrast, the network proposed by \citet{DouCVPR2017} has a ``sub-CNN'' that combines features from intermediate layers of the VGG-Face network \citep{ParkhiBMVC2015_VGGFace}. This sub-CNN estimates ex\-pression-specific features while the main network, the VGG-Face, estimates the identity-specific features. Both types of features are then combined to form a 3D face.

On the other hand, \citet{XChaiICME2020} proposed to process the input image by two streams that separately estimate the shape and the texture. The shape stream is composed of an encoder that extracts image features, and two separate decoders that estimate identity and expression attributes, which are combined to construct the final 3D shape. In addition, the features extracted by the encoder are passed through two fully-connected layers to estimate the pose, illumination, and 3DMM albedo parameters. The texture stream basically consists of a style transfer network, where the style of the input image is transferred to the estimated 3DMM albedo in order to obtain a realistic skin-style texture map.

\subsubsection{Geometric Deep Learning} \label{subsubsec:DL_GCN}
The most common way of representing a 3D facial surface is with a mesh, i.e., an undirected graph whose vertices lie in $\mathbb{R}^3$. However, classical convolutions cannot deal with non-Euclidean domains (such as meshes, graphs, manifolds, etc.), which is why many researchers have used alternative representations compliant with the Euclidean setting. As we have already mentioned above, some have used depth maps \citep{SelaICCV2017,ZShuFG2020,KoizumiECCV2020} or volumetric representations \citep{JacksonICCV2017,HYiCVPR2019}, and others have projected the 3D face into a 2D space, such as the input image space \citep{JourablooCVPR2016,XZhuCVPR2016,XZhuPAMI2019} or the UV space \citep{YFengECCV2018,XZhuECCV2020,GulerCVPR2017}. Still, the parametric representation provided by the 3DMMs is the most widespread \citep{TewariICCV2017,TewariCVPR2018,TewariPAMI2020,SavovICCVW2019,FWuCVPR2019,RamonICCVW2019,Richardson3DV2016,XChaiICME2020,XTuToM2020,GenovaCVPR2018}, since the 3DMMs model the shape variations with an Euclidean subspace, in which a 3D face is described by the model parameters.

An emerging recent research field called geometric deep learning \citep{BronsteinSPM2017_GeomDL} has focused on generalising deep learning methods to non-Euclidean data. In particular, \citet{BrunaICLR2014} proposed a technique to extend classical convolutions to graphs using spectral filters. More specifically, given a mesh with a set of $N$ vertices $\mathcal{V}\subset\mathbb{R}^3$ and an adjacency matrix $\mathbf{A}\in\{0,1\}^{N\times N}$, its normalised graph Laplacian is defined as
\[
\mathbf{L} = \mathbf{I}_N - \mathbf{D}^{-\frac{1}{2}}\mathbf{A}\mathbf{D}^{-\frac{1}{2}}
\]
where $\mathbf{D}$ is a diagonal matrix whose entries are the degree of each vertex, $\mathbf{D}_{ii} = \sum_{j=1}^N \mathbf{A}_{ij}$, and $\mathbf{I}_N$ is the $N\times N$ identity matrix. Since $\mathbf{L}$ is real symmetric, there exists an orthonormal matrix $\mathbf{U}\in\mathbb{R}^{N\times N}$ and a diagonal matrix $\mathbf{\Lambda}\in\mathbb{R}^{N\times N}$ such that $\mathbf{L} = \mathbf{U}\mathbf{\Lambda}\mathbf{U}^\text{T}$. The eigenvectors of $\mathbf{L}$, i.e., the columns of $\mathbf{U}$, form an orthonormal basis of $\mathbb{R}^N$ analogous to the Fourier basis in the Euclidean space, and the projection of a signal $\vec{x}\in\mathbb{R}^N$ into this space, $\mathcal{F}(\vec{x}) = \mathbf{U}^\text{T}\vec{x}\in\mathbb{R}^N$ is called its graph Fourier transform. In this new basis, the graph convolution operation $*_\mathcal{G}$ of a signal $\vec{x}\in\mathbb{R}^{N}$ with a filter $\vec{g}$ is defined as a product in the spectral domain:
\begin{align*}
\vec{x}*_\mathcal{G}\vec{g} &= \mathcal{F}^{-1}\left( \mathcal{F}(\vec{x}) \odot \mathcal{F}(\vec{g}) \right)\\
&= \mathbf{U}\left( \mathbf{U}^\text{T}\vec{x} \odot \mathbf{U}^\text{T}\vec{g} \right)
\end{align*}
where $\odot$ is the Hadamard product, and $\mathcal{F}^{-1}(\vec{x}) = \vec{Ux}$ is the inverse graph Fourier transform (due to the orthonormal property of $\mathbf{U}$). Therefore, if we denote $\vec{g}_{\vec{w}} = \text{diag}(\vec{U}^\text{T}\vec{g})\in\mathbb{R}^{N\times N}$, which can be interpreted as the coefficients of a filter in the spectral domain, the graph convolution is simplified as
\begin{equation}\label{eq:graphconv}
\vec{x}*_\mathcal{G}\vec{g}_\vec{w} = \mathbf{U}\left( \vec{g}_\vec{w}\mathbf{U}^\text{T}\vec{x}\right).
\end{equation}

Thus, the convolution on the non-Euclidean graph domain on the left-hand side of eq. \eqref{eq:graphconv} becomes a straight-forward product in the spectral domain, which is related to the graph domain by simple matrix operations. This construction of the spectral graph convolution is followed by all the works that extended the successful CNNs to undirected graphs, called graph convolutional networks (GCNs), differing in the choice of the filter $\vec{g}_\vec{w}$. We refer to a recently published survey on GCNs presented by \citet{ZWuNNLS2021} for a more detailed overview of these networks. Nevertheless, we outline the most used filter among the works presented in this section, which is parametrised by a truncated expansion of the Chebyshev polynomials:
\[
\vec{g}_\vec{w} = \sum_{k=0}^K w_k T_k(\widetilde{\mathbf{\Lambda}})
\]
where $T_k(\cdot)$ is the Chebyshev polynomial of order $k$, $\vec{w} = (w^0,\dots,w^K)\in\mathbb{R}^{K+1}$ are the weights that are learnt by the network in the training phase, and $\widetilde{\vec{\Lambda}} = 2\mathbf{\Lambda}/\lambda_\text{max} - \vec{I}_N$, with $\lambda_\text{max}$ the maximum eigenvalue of $\vec{L}$. Therefore, the spectral graph convolution (eq. \eqref{eq:graphconv}) is written as
\[
\vec{x}*_\mathcal{G}\vec{g}_\vec{w} = \mathbf{U}\left(\left( \sum_{k=0}^K w_k T_k(\widetilde{\mathbf{\Lambda}})\right)\mathbf{U}^\text{T}\vec{x}\right).
\]

Denoting $\widetilde{\mathbf{L}} = \vec{U}\widetilde{\mathbf{\Lambda}}\vec{U}^\text{T} = 2\mathbf{L}/\lambda_\text{max} - \vec{I}_N$, it is easy to prove that $T_k(\widetilde{\mathbf{L}}) = \vec{U}T_k(\widetilde{\mathbf{\Lambda}})\vec{U}^\text{T}$, thus the spectral graph convolution becomes
\[
\vec{x}*_\mathcal{G}\vec{g}_\vec{w} = \sum_{k=0}^K w_k T_k(\widetilde{\mathbf{L}})\vec{x},
\]
which simplifies the computations since the Chebyshev polynomials are defined recursively, and the eigendecomposition of the graph Laplacian $\vec{L}$ is avoided.

The introduction of GCNs allows directly processing the 3D facial meshes, without the need of intermediate representations, thus keeping its topological structure. In addition, the number of parameters in GCNs is small, which translates to reduced computational complexity.

Motivated by the advantages of GCNs, in the past few years there have been researchers that have used this type of networks to build non-linear 3D facial models \citep{RanjanECCV2018,BouritsasICCV2019,ZJiangCVPR2019}. By training a mesh-to-mesh autoencoder, they are able to learn a low-dimensional latent space that models facial shape and expression variations. Then, after training, the encoder serves as a model fitting system and the decoder as a 3D facial mesh regressor.

Also, very recently, some of the proposed works on 3D face reconstruction from uncalibrated 2D images have used GCNs to regress directly a 3D facial mesh. With a very similar idea to those works that construct non-linear 3D facial models \citep{RanjanECCV2018,BouritsasICCV2019,ZJiangCVPR2019}, \citet{YZhouCVPR2019} trained an architecture composed of a mesh autoencoder and an additional encoder that extracts features also from 2D images (see Section \ref{subsubsec:DL_EncDec}). By jointly training the whole architecture, both encoders learn the same latent representation from both images and meshes, which are then directly decoded into a 3D facial mesh. \citet{GHLeeCVPR2020}, \citet{ZGaoCVPRW2020} and \citet{JLinCVPR2020} used GCNs to decode image features into 3D facial meshes. Whereas the former regressed the 3D shape from uncertainty-aware features extracted from the input image (see Section \ref{subsubsec:DL_EncDec}), \citet{ZGaoCVPRW2020} extracted identity- and expression-related features and used two GNCs to separately regress the 3D facial shape and the per-vertex texture. In contrast, \citet{JLinCVPR2020} used GCNs to refine the per-vertex texture obtained with a 3DMM-fitting scheme. Therefore, the features $\vec{x}_i$ in eq. \eqref{eq:graphconv} are the RGB colours at each vertex, instead of their XYZ coordinates.

%%%%%%%%%%%%%%%%%%%%%%%%%%%%%%%%%%%%%%%%%%%%%%%%%%%%%%%%%%%%%%%%%%%%%%%%%%%

\subsection{Training criterion}
Another decisive element for the learning is the training criterion, i.e the loss function (also called cost or error function) minimised by the network in the training process. Evidently, this loss function depends directly on the output of the network since, roughly, the weights of the network are learnt to minimise the error between the estimated output and the ground truth. Accordingly, we group the errors computed in the parameters space in Section \ref{subsubsec:DL_ParamsLoss}, and the ones computed in the 3D space in Section \ref{subsubsec:DL_lossIn3D}. Also, as stated above, many works have trained their networks with unlabelled data, thus we review the losses used to carry out self-supervised training in Section \ref{subsubsec:DL_SelfSupervisedTraining}. In Section \ref{subsubsec:AdversarialLoss}, we show the losses used to train the proposed generative adversarial networks, and, in Section \ref{subsubsec:DL_otherLosses}, losses that cannot be categorised in any of the above groups.

% PARAMS ERROR
% Leave Jourabloo at the end so it links with LMKS ERROR with Bhagavatula (3D FACES ERROR)

\subsubsection{Loss Functions in the Parameters Space} \label{subsubsec:DL_ParamsLoss}
Most of the presented works estimate the parameters of a 3DMM from the input image. Among them, a popular term included in the cost function is the error between the estimated and the ground truth parameters, $\vec{\theta}$ and $\vec{\theta}^\text{gt}$, respectively. Some of these works \citep{FWuCVPR2019,JSYoonCVPR2019,JourablooCVPR2016,JourablooIJCV2017,SenguptaCVPR2018,ChaudhuriCVPR2019,ZShuFG2020,XChaiICME2020} simply computed the $L_p$ norm of the difference between $\vec{\theta}$ and $\vec{\theta}^\text{gt}$ (raised to $p$th power),
\begin{equation} \label{eq:error_pnorm}
 \mathbb{E}_{p\text{-norm}}(\vec{\theta},\vec{\theta}^\text{gt}) = \norm{\vec{\theta} - \vec{\theta}^\text{gt}}_p^p = \sum_{i=1}^K | \theta_i - \theta^\text{gt}_i |^p,
\end{equation}
whereas others \citep{KimCVPR2018,JourablooICCV2017,XZhuCVPR2016,XZhuPAMI2019,XTuToM2020,JGuoECCV2020,GenovaCVPR2018} used a \textit{weighted parameters distance cost}, which computes the Euclidean distance between $\vec{\theta}$ and $\vec{\theta}^{\text{gt}}$, but with weighting factors for each parameter to balance their global importance,
\begin{equation}\label{eq:error_wpdc}
 \mathbb{E}_{\text{wpdc}}(\vec{\theta},\vec{\theta}^\text{gt}) = (\vec{\theta} - \vec{\theta}^\text{gt})^\text{T} \vec{W} (\vec{\theta} - \vec{\theta}^\text{gt}).
\end{equation}

% \subsubsection{$L_p$ Norms in the Parameters Space}

All the works that used $L_p$ norms to compute the parameters error either used the $L_1$ \citep{ChaudhuriCVPR2019}, or the $L_2$ \citep{XChaiICME2020,FWuCVPR2019,JSYoonCVPR2019,JourablooCVPR2016,JourablooIJCV2017,ZShuFG2020}. In addition, to regularise the reconstruction, \citet{FWuCVPR2019}, \citet{JSYoonCVPR2019} and \citet{ZShuFG2020} minimised the distance between the landmarks on the input image and the projection of the landmarks of the estimated 3D face. Moreover, \citet{ZShuFG2020} ensured consistency in the parameters estimated from images taken simultaneously but from different viewpoints by minimising their difference. \citet{JourablooIJCV2017} used the mirrored version of the input image to regularise the results by including a term that ensured that the landmarks on the input face were similar to those on the mirrored image (with a reordering), thus minimising  $L_2$ norm of the difference between the projected landmarks over the original image and over the mirrored image.

Differently from the above works, \citet{SenguptaCVPR2018} used both the $L_1$ and the $L_2$ norms, depending on the parameters. The error over the lighting parameters $\vec{\gamma}$ was computed using a $L_2$ norm, whereas the error over the surface normals $\vec{n}$, the error over the 3DMM texture parameters $\vec{\beta}$, and the image error were computed using the $L_1$ norm: %{\color{red}REMOVE FORMULAS?}
\begin{align}
    & \mathbb{E}_{\vec{\gamma}}(\vec{\gamma},\vec{\gamma}^\text{gt}) = \norm{\vec{\gamma} - \vec{\gamma}^\text{gt}}_2, \nonumber\\
    & \mathbb{E}_{\vec{n}}(\vec{n},\vec{n}^\text{gt}) = \norm{\vec{n} - \vec{n}^\text{gt}}_1, \label{eq:errorN}\\
    & \mathbb{E}_{\vec{\beta}}(\vec{\beta},\vec{\beta}^\text{gt}) = \norm{\vec{\beta} - \vec{\beta}^\text{gt}}_1, \nonumber%\\
  %  & \errorI(\Imod,\Iin) = \norm{\Imod - \Iin}_1. \label{eq:errorI_L1}
\end{align}
% Their loss function was composed of four terms $\mathbb{E} = w_{\vec{n}} \mathbb{E}_{\vec{n}} + w_{\vec{\beta}}\mathbb{E}_{\vec{\beta}} + w_{\vec{I}}\errorI + w_{\vec{\gamma}} \mathbb{E}_{\vec{\gamma}}$.

% IMAGE ERROR (unsupervised?)
% Introduced with Richardson, continue with YGuo and Sengupta.
% Leave Tewari and LTran at the end since they are unsupervised.

% \subsubsection{Weighted Parameters Errors} \label{subsubsec:DL_WeightParamsErr}

On the other hand, the works that used the \textit{weighted parameters distance cost} (eq. \eqref{eq:error_wpdc}) have proposed different weighting matrices, which is the main difference between them. The weight matrix used by \citet{JourablooICCV2017} is a diagonal matrix with the inverse of the standard deviations of each parameter, $\vec{W} = \text{diag}\left(\frac{1}{\sigma_1},\dots,\frac{1}{\sigma_K}\right)$, which normalises the importance of each element in the loss function.

As described in Section \ref{subsubsec:DL_Iterative}, \citet{XZhuCVPR2016,XZhuPAMI2019} used a cascaded regressor approach to iteratively estimate the parameters update $\vec{\Delta} \vec{\theta}^k$ to move the initial parameters $\vec{\theta}^0$ closer to the $\vec{\theta}^\text{gt}$. Thus, the cost function in eq. \eqref{eq:error_wpdc} is slightly modified to
\begin{equation*}
 \mathbb{E}_{\text{wpdc}}(\vec{\Delta}\vec{\theta},\vec{\Delta}\vec{\theta}^\text{gt}) = \left(\vec{\Delta}\vec{\theta} - \vec{\Delta}\vec{\theta}^\text{gt}\right)^\text{T} \vec{W} \left(\vec{\Delta}\vec{\theta} - \vec{\Delta}\vec{\theta}^\text{gt}\right)
\end{equation*}
where $\vec{\Delta}\vec{\theta}^\text{gt} = \vec{\theta}^\text{gt} - \vec{\theta}^0$. In both works, the weight matrix is a diagonal matrix $W = \text{diag}(\vec{w})$, $\vec{w} = (w_1,\dots,w_K)$. \citet{XZhuCVPR2016} defined the weights $w_i$, $i = 1,\dots,K$ proportional to the squared error of model projections on the 2D image:
\begin{equation}\label{eq:error_wpdc_XZhuCVPR2016}
 w_i = \frac{\norm{\mathcal{P}\left(\vec{x}\left(\tilde{\vec{\theta}}^{[i]}\right),\tilde{\vec{\theta}}^{[i]}\right) - \mathcal{P}\big(\vec{x}\left(\vec{\theta}^\text{gt}\right),\vec{\theta}^\text{gt}\big)}_2}{\sum_{k=1}^K w_k}.
\end{equation}
Here, $\tilde{\vec{\theta}}^{[i]}$ defines the vector of parameters where the $i$th element comes from the estimated parameter $\tilde{\theta}^{[i]}_i = \theta_i$, whereas the rest come from the ground truth $\tilde{\theta}^{[i]}_j = \theta_j^\text{gt}$, $\forall j\neq i$. $\vec{x}(\vec{\theta})$ defines the 3D facial shape built with the shape and expression parameters from $\vec{\theta}$, and $\mathcal{P}(\vec{x}(\vec{\theta}),\vec{\theta})$ defines the projection of $\vec{x}(\vec{\theta})$ according to the projection parameters in $\vec{\theta}$. In a subsequent work, \citet{XZhuPAMI2019} enhanced this formulation to optimise the weights at each iteration
\begin{align*}
    \vec{w}^* = &\argmin_{\vec{w}} \norm{ \mathcal{P} \big( \vec{x}( \check{\vec{\theta}} ), \check{\vec{\theta}} \big) - \mathcal{P}\big( \vec{x}(\vec{\theta}^\text{gt}), \vec{\theta}^\text{gt} \big)}_2^2 \\
    & + \lambda \norm{ \text{diag}(\vec{w}) ( \vec{\theta} - \vec{\theta}^\text{gt})}_2^2 \\
    &s.t. \ \ 0 \leq w_i \leq 1, \ \ \forall i
\end{align*}
where $\check{\vec{\theta}} = \vec{\theta} + \text{diag}(\vec{w})( \vec{\theta} - \vec{\theta}^\text{gt})$ and $\lambda \in \mathbb{R}_{\geq0}$. According to the authors, adding the weighted updated $\text{diag}(\vec{w})( \vec{\theta} - \vec{\theta}^\text{gt})$ to the current estimated parameters $\vec{\theta}$ allows estimating a new face closer to the ground truth face, thus improving the final result.

A similar approach was used by \citet{XTuToM2020} and \citet{JGuoECCV2020}, who proposed weighting matrices based on that from \citep{XZhuCVPR2016} (eq. \ref{eq:error_wpdc_XZhuCVPR2016}). Whereas the former \citep{XTuToM2020} only computed the projection of the landmarks, rather than projecting the whole facial reconstruction $\vec{x}(\vec{\theta})$, \citet{JGuoECCV2020} computed the distance between the 3D faces directly and normalised the weights by their maximum value:
\begin{equation*}
     w_i = \frac{\norm{\vec{x}\left(\tilde{\vec{\theta}}^{[i]}\right) - \vec{x}\left(\vec{\theta}^\text{gt}\right)}_2}{\max \ \{w_k\}_{k=1}^K}.
\end{equation*}

In contrast, \citet{GenovaCVPR2018} and \citet{KimCVPR2018} used a common weight for all the parameters of the same space, i.e., one weight for the shape parameters and another one for the texture parameters. Additionally, \citet{GecerCVPR2019} used other two terms: identity loss and batch distribution loss. The first one ensures that the identity features $\vec{f}$ extracted from a real input image and from the image rendered using the parameters estimated by the decoder are similar, and it is computed based on cosine similarity: $\mathbb{E}_{\text{id}}(\vec{f}_i,\vec{f}_j) = \vec{f}_i \cdot \vec{f}_j$. The batch distribution loss is a regularisation term applied at a batch level rather than to the whole training set, which enforces that the estimated parameters match the zero-mean standard normal distribution, assumed by the 3DMM.

Finally, \citet{TranCVPR2017,TranCVPR2018} and \citet{SanyalCVPR2019} also used losses in the parameters space, but adopted none of those explained above. \citet{TranCVPR2017,TranCVPR2018} highlighted that the Euclidean loss, i.e. $L_2$, favours reconstructions closer to the mean, resulting in less detailed faces. To avoid this, they used the \textit{asymmetric Euclidean loss}, which decouples the over-estimation and under-estimation errors (first and second terms in eq. \eqref{eq:error_asym}, respectively), favouring reconstructions further away from the mean:
\begin{align} \label{eq:error_asym}
 \mathbb{E}_{\text{asym}}&(\vec{\theta},\vec{\theta}^\text{gt}) = \nonumber \\
 & = w_1 \norm{\text{abs}(\vec{\theta}^\text{gt}) - \max(\text{abs}(\vec{\theta}),\text{abs}(\vec{\theta}^\text{gt}))}^2_2 + \nonumber \\
 & \quad \ w_2 \norm{\text{abs}(\vec{\theta}) - \max(\text{abs}(\vec{\theta}),\text{abs}(\vec{\theta}^\text{gt}))}_2^2,
\end{align}
where $\text{abs}(\cdot)$ and $\max(\cdot)$ are defined for each element of $\vec{\theta}$ and $\vec{\theta}^\text{gt}$. The weights $w_1, w_2 \in \mathbb{R}_{\geq0}$ control the trade-off between the over- and under-estimation errors. In addition, in their subsequent work \citep{TranCVPR2018}, they included a loss function to learn to predict the bump map $\vec{\Gamma}(x,y)$, defined as a displacement for each pixel $(x,y)$ in the input image, which refines the 3D face estimated by \citep{TranCVPR2017}. This \textit{bump loss} computes the error of the estimated bump map, including the difference between the gradient of the estimated and ground truth bump maps to favour smoother surfaces
\begin{align*}
 \mathbb{E}_{\text{bump}}(\vec{\Gamma},\vec{\Gamma}^\text{gt}) &= \norm{\vec{\Gamma} - \vec{\Gamma}^\text{gt}}_1 + \norm{\frac{\partial \vec{\Gamma}}{\partial x} - \frac{\partial \vec{\Gamma}^\text{gt}}{\partial x}}_1 + \\ 
 & \quad \norm{\frac{\partial \vec{\Gamma}}{\partial y} - \frac{\partial \vec{\Gamma}^\text{gt}}{\partial y}}_1.
\end{align*}

On the other hand, \citet{SanyalCVPR2019} proposed to feed the network with subsets of $R$ images during training, where $R-1$ of these images were of the same person and the $R$th image belonged to a different person. This subset of images allowed the authors to impose shape consistency among the reconstructions of the same person, while enforcing them to be different from the reconstruction of the ``other person", basically following a similar idea to the popular triplet loss \citep{SchroffCVPR2015_FaceNet}. This shape consistency constraint was mathematically expressed as
\begin{equation} \label{eq:shape_consistency}
\norm{\vec{\alpha}_i - \vec{\alpha}_j}_2^2 + \eta \leq \norm{\vec{\alpha}_i - \vec{\alpha}_R}_2^2    
\end{equation}
for all $i,j = 1,\dots,R-1$, $i\neq j$ and some $\eta \in \mathbb{R}_{\geq0}$, where $\vec{\alpha}_i$ are the estimated 3DMM shape parameters for the $i$th image. They translated eq. \eqref{eq:shape_consistency} into the loss function:
\begin{equation*}
    \mathbb{E} = \sum_{i,j = 1}^{R-1} \max \Big( 0, \norm{\vec{\alpha}_i - \vec{\alpha}_j}_2^2 + \eta - \norm{\vec{\alpha}_i - \vec{\alpha}_R}_2^2\Big).
\end{equation*}

\subsubsection{Loss Functions in 3D Space} \label{subsubsec:DL_lossIn3D}

Another loss function that has been used is the Euclidean distance between the reconstructed and the ground truth 3D faces. Although, as mentioned above, several representations have been used and thus the loss function may differ across representations. Among the works that have computed the error in 3D space, some regressed directly the 3D coordinates of the 3D face \citep{BhagavatulaICCV2017,FLiuCVPR2018,GZhangFG2018,BaiCVPR2020,XFanToM2020,XZhuECCV2020,XLiICASSP2020,GHLeeCVPR2020,FLiuICCV2019}, others estimated 3DMM parameters \citep{DouCVPR2017,Richardson3DV2016,RichardsonCVPR2017,ChinaevECCVW2018,JGuoECCV2020,HYiCVPR2019,XWangCVPR2020,ZGaoCVPRW2020}, or a depth map \citep{XZengICCV2019,JShangECCV2020,ZShuFG2020,SelaICCV2017}, and \citep{JacksonICCV2017,HYiCVPR2019} used a volumetric representation.

% 3D face directly
The works that estimated the 3D face directly, simply compute the $L_p$ norm ($p = 1$ or $2$) of the difference between the estimated 3D face, $\vec{x}$, and the ground truth face, $\vec{x}^\text{gt}$,
\[
\mathbb{E}_\text{rec}\left(\vec{x},\vec{x}^\text{gt}\right) = \norm{\vec{x} - \vec{x}^\text{gt}}_p.
\]
Specifically, \citet{XFanToM2020} and \citet{XZhuECCV2020} estimated an update $\Delta \vec{x}$ from a template face $\vec{x}_\text{T}$, therefore they minimised $\norm{\Delta \vec{x} - \Delta \vec{x}^\text{gt}}_2$, where $\Delta \vec{x}^\text{gt} = \vec{x}^\text{gt} - \vec{x}_\text{T}$. \citet{FLiuICCV2019} and \citet{BaiCVPR2020} included two additional losses: a cosine distance between normals of the estimated 3D face $\vec{n}_i$, and normals of the ground truth $\vec{n}_{i,\text{gt}}$:
\begin{equation}\label{eq:errorN_cosDist}
\mathbb{E}_\vec{n}\left(\vec{n},\vec{n}^\text{gt}\right) = \sum_{i} \left( 1 - \cos\left(\vec{n}_i\cdot\vec{n}_i^\text{gt}\right)\right).
\end{equation}
and an edge loss that forces each pair of adjacent vertices $(\vec{x}_i, \vec{x}_j)$ to be at the same distance as the corresponding vertices in the ground truth 3D face, $(\vec{x}_{i,\text{gt}}, \vec{x}_{j,\text{gt}})$:
\[
\mathbb{E}_\text{edge}\left(\vec{x},\vec{x}^\text{gt}\right) = \frac{1}{\abs{\mathcal{E}}} \sum_{(i,j)\in\mathcal{E}} \abs{\frac{\norm{\vec{x}_i - \vec{x}_j}_2}{\norm{\vec{x}_{i,\text{gt}} - \vec{x}_{j,\text{gt}}}_2} - 1}.
\]
where $\mathcal{E}$ is the set of edges in the estimated 3D face. According to \citep{BaiCVPR2020}, these two losses improve the smoothness of the surface while preserving high-frequency details.

To also train the network on face recognition, \citet{FLiuCVPR2018} added a softmax loss that served as a face identification error. Similarly, \citet{XLiICASSP2020} added the cross-entropy loss to obtain more robust identity features. Finally, \citet{BhagavatulaICCV2017} aimed at face alignment, thus added an Euclidean loss over the 2D regressed landmarks so that the network also learned to estimate the 2D landmarks over the input image.

Other works \citep{DouCVPR2017,Richardson3DV2016,RichardsonCVPR2017,ChinaevECCVW2018,JGuoECCV2020,HYiCVPR2019,XWangCVPR2020,ZGaoCVPRW2020} estimated the parameters of a 3DMM, thus, to apply the Euclidean loss to the 3D faces, first the 3D geometry (and texture in some cases) had to be recovered by multiplying the model basis by the estimated parameters,
\[
\mathbb{E}_\text{rec}\left(\vec{\theta},\vec{\theta}^\text{gt}\right) = \norm{\vec{x}\left(\vec{\theta}\right) - \vec{x}\left(\vec{\theta}^\text{gt}\right)}_2.
\]
where $\vec{\theta}$ are the corresponding model parameters, and $\vec{x}(\vec{\theta})$ is the recovered face from the 3DMM. In addition, \citet{ChinaevECCVW2018} included the $L_2$ norm between the projection to the image plane of the ground truth and the reconstructed 3D faces, and \citet{XWangCVPR2020} added the normal cosine distances (eq. \eqref{eq:errorN_cosDist}).

Unlike the other works, \citet{HYiCVPR2019} separated the influence of the shape parameters $\vec{\alpha}$ and the expression parameters $\vec{\delta}$ to the final facial geometry, $\vec{x}(\vec{\alpha},\vec{\delta})$,
\begin{align*}
    \mathbb{E}_{\text{rec}} =&\ w_{\vec{\alpha}}\norm{\vec{x}(\vec{\alpha},\vec{\delta}^\text{gt}) - \vec{x}(\vec{\alpha}^\text{gt},\vec{\delta}^\text{gt})}_2^2 +\\ & \ w_{\vec{\delta}}\norm{\vec{x}(\vec{\alpha}^\text{gt},\vec{\delta}) - \vec{x}(\vec{\alpha}^\text{gt},\vec{\delta}^\text{gt})}_2^2,
\end{align*}
where $\vec{\alpha}^\text{gt}$ and $\vec{\delta}^\text{gt}$ are the ground truth shape and expression parameters, respectively. Similarly, \citet{ZGaoCVPRW2020} added a term to train the network to disentangle the identity features from the expression features. To do so, they minimised the difference between the 3D face reconstructed from known 3DMM parameters, $\vec{x}(\vec{\alpha}^\text{gt},\vec{\delta}^\text{gt})$, and the corresponding 3D face regressed by the decoder from identity and expression features, $\vec{f}_\text{id},\vec{f}_\text{exp}$, estimated by the encoder, $\vec{x}(\vec{f}_\text{id},\vec{f}_\text{exp})$. Specifically, this term swaps the identity and the expression of two input images $\mathbf{I}_\text{A}$ and $\mathbf{I}_\text{B}$ from different subjects,
\begin{equation*}
    \mathbb{E} = \norm{\vec{x}(\vec{\alpha}^{\text{A},\text{gt}},\vec{\delta}^{\text{B},\text{gt}}) - \vec{x}(\vec{f}^\text{A}_\text{id},\vec{f}^\text{B}_\text{exp})}_1
\end{equation*}
where $\vec{\alpha}^{\text{gt},\text{A}}$ are the known shape parameters from $\mathbf{I}_\text{A}$, $\vec{\delta}^{\text{B},\text{gt}}$ the known expression parameters from $\mathbf{I}_\text{B}$, and similarly, $\vec{f}^\text{A}_\text{id}$ and $\vec{f}^\text{B}_\text{exp}$ the identity and expression features estimated by the encoder from $\mathbf{I}_\text{A}$ and $\mathbf{I}_\text{B}$, respectively. Also, to ensure smooth reconstructions, the difference between each vertex in the reconstruction $\vec{x}_i$ and its first order neighbours $\mathcal{N}_i$ is compared to that of a template $\vec{x}_{i,\text{T}}$ (the mean of the 3DMM):
\begin{align*}
    \mathbb{E}_\text{smooth}(\vec{x},\vec{x}_\text{T}) = \sum_i &\left\lvert \left( \vec{x}_i  - \frac{1}{\abs{\mathcal{N}_i}} \sum_{j\in\mathcal{N}_i}\vec{x}_j\right)\right. - \\
    &\left.\left( \vec{x}_{i,\text{T}}  - \frac{1}{\abs{\mathcal{N}_i}} \sum_{j\in\mathcal{N}_i}\vec{x}_{j,\text{T}}\right)\right\rvert.
\end{align*}

Other works \citep{XZengICCV2019,SelaICCV2017,ZShuFG2020,JShangECCV2020} estimated a depth map from the input 2D image and, therefore, their reconstruction loss is the difference between the predicted and the ground truth depths maps. \citet{XZengICCV2019} and \citet{SelaICCV2017} also included the $L_1$ error between the normals as in eq. \eqref{eq:errorN} to avoid over-sharpened estimations produced by the $L_1$ loss in the depth error function. \citet{ZShuFG2020} also added a normals error but further regularised the output depth map with a 3DMM by minimising the difference between the estimated depth map and the depth map obtained from a 3DMM-reconstruction estimated with a pre-trained network. In contrast, \citet{JShangECCV2020} used the depth error as an additional multi-view consistency loss to ensure depth consistency across the three input images. %The designed depth consistency loss is very similar to the image consistency loss (eq. \eqref{eq:pcConsistency_JShangECCV2020}) but the synthesised depth map from the adjacent view are scaled to rectify the difference of scale in both depth maps.

\citet{JacksonICCV2017} and \citet{HYiCVPR2019} represented the reconstructed 3D faces as binary volumes (see Section \ref{subsubsec:DL_EncDec}), thus they minimised the sigmoid cross-entropy loss,

\begin{align*}
\mathbb{E}_\text{vol}
\left(V,V^\text{gt}\right)= \sum_{w}\sum_{h}\sum_{d} \ &V^\text{gt}_{whd} \log \left(V_{whd}\right) + \\
&\left(1-V^\text{gt}_{whd}\right) \cdot \log \left(1-V_{whd}\right),
\end{align*}
where $V_{whd}$ is the estimated value of voxel $(w,h,d)$ and $V_{whd}^\text{gt}$ its ground truth value.

Differently to all the works presented above, \citet{TrigeorgisCVPR2017} based their facial representation on the surface normals, thus they minimised the cosine distance between the estimated and the ground truth normals, as in eq. \eqref{eq:errorN_cosDist}.

% Also differently from the previous works, \citet{XTuToM2020} proposed a set of three landmark errors to ensure consistency in 3D and in the image plane. The first is the distance between the projected 3D landmarks (from the reconstructed 3D face), $\LmkProj$, and the ground truth 2D landmarks, $\lmk$. The second is the distance between the 3D landmarks of the estimated 3D face, $\Lmk$, and the back-projection of $\LmkProj$ estimated in the backward pass, $\vec{\hat{\hat{\ell}}}{}^\text{ 3D}$. Finally, to ensure the consistency of the cycle formed by the forward and the backward pass, they computed the distance between the ground truth 2D landmarks, $\lmk$, and the projection of $\vec{\hat{\hat{\ell}}}{}^\text{ 3D}$ into the image plane, $\vec{\hat{\hat{\ell}}}{}^\text{ 2D}$.

\subsubsection{Self-Supervised Training} \label{subsubsec:DL_SelfSupervisedTraining}

Recently, comparing the synthetic image that results from rendering the estimated 3D textured face, $\Imod$, with the input image $\Iin$ has allowed many researchers to train neural networks without the need of 2D-3D data pairs (see Section \ref{subsubsec:DL_SelfSupervisionTS}). Usually, the rendered image $\Imod$ is modelled by the Lambertian reflectance model, which is approximated using spherical harmonics as basis functions $\vec{\mathcal{Y}}(\vec{n})$,
\[
\Imod(x,y) = \vec{\mathcal{A}}(x,y) \gamma^\text{T}\vec{\mathcal{Y}}(\vec{n})
\]
where $(x,y)$ are pixel locations, $\vec{\mathcal{A}}$ is the albedo, $\vec{n}$ are the surface normals, and $\vec{\gamma}$ the spherical harmonics coefficients.

Many of the proposed works \citep{LTranCVPR2018,LTranPAMI2019,LTranCVPR2019,YDengCVPRW2019,YZhouCVPR2019,JLinCVPR2020,ChaudhuriECCV2020,PWangICME2020,XChaiICME2020,YChenTIP2020,RichardsonCVPR2017,YZhouCVPR2019} drove the learning in a self-supervised manner by minimising the $L_p$ norm ($p = 1$ or $2$) of the difference between $\Iin$ and $\Imod$:
\begin{equation} \label{eq:errorI_selfsup}
    \errorI = \norm{\Iin - \Imod}_p.
\end{equation}
However, this loss alone is not capable of accurately reconstructing 3D faces and thus additional losses are included to further constrain the training. One of the most used losses exploits the known correspondence of landmarks by forcing the projection of the 3D landmarks, $\LmkProj$, to match the 2D landmarks of the input image, $\lmk$:
\begin{equation}\label{eq:errorlmks_DL}
    \errorlmk = \sum_{i = 1}^L \norm{ \lmk_i - \LmkProj_i }_2^2.
\end{equation}

Another loss that is widely used \citep{LTranPAMI2019,LTranCVPR2019,YDengCVPRW2019,JLinCVPR2020,PWangICME2020,XChaiICME2020,YChenTIP2020} is the perceptual loss that enforces similar feature representations of both the rendered and the input images as computed by a feature extraction function $\varphi$. Some works computed this similarity as the $L_2$ norm of the difference between features extracted from $\Iin$ and $\Imod$ \citep{LTranPAMI2019,LTranCVPR2019,PWangICME2020,YChenTIP2020}:
\begin{equation} \label{eq:errorI_featuresL2}
    \mathbb{E}_{\mathbf{I},\varphi}(\Imod,\Iin) = \frac{1}{|\mathcal{C}|} \sum_{j \in \mathcal{C}} \frac{\norm{\varphi_j (\Imod) - \varphi_j(\Iin)}_2^2}{W_jH_jC_j},
\end{equation}
where $\varphi_j(\mathbf{I})$ is the activation of the $j$th layer of the feature-extractor network when processing $\mathbf{I}$ with dimension $W_j \times H_j \times C_j$ (number of rows, columns and channels, respectively), and $\mathcal{C}$ is the set of layers used to compute the similarity. Alternatively, other works \citep{YDengCVPRW2019,JLinCVPR2020,XChaiICME2020} computed the cosine distance between the feature vectors:
\begin{equation}\label{eq:errorI_featuresCos}
    \mathbb{E}_{\mathbf{I},\varphi} (\Imod, \Iin) = 1 - \frac{\varphi(\Iin)^\text{T} \varphi(\Imod)}{\norm{\varphi(\Iin)}_2\norm{\varphi(\Imod)}_2}.
\end{equation}
As feature extractor $\varphi$, \citet{LTranPAMI2019,LTranCVPR2019} and \citet{YChenTIP2020} used the VGG-Face network \citep{ParkhiBMVC2015_VGGFace}, \citet{YDengCVPRW2019} and \citet{JLinCVPR2020} used the FaceNet \citep{SchroffCVPR2015_FaceNet}, \citet{PWangICME2020} used the VGG-16 network \citep{SimonyanICLR2015_VGG2} and \citep{XChaiICME2020} the ArcFace \citep{DengCVPR2019_arcface}.

In addition, \citet{LTranPAMI2019} used three regularisation terms: albedo symmetry, albedo constancy and shape smoothness within a neighbourhood. In their latter work \citep{LTranCVPR2019}, to promote high-level details, they proposed to transfer this strong regularisation to auxiliary shapes and albedos, and apply a weak regularisation to the original ones by maximising similarity between them. On the other hand, since \citet{PWangICME2020}, \citet{YChenTIP2020}, and \citet{RichardsonCVPR2017} refined the resulting reconstruction by estimating per-vertex displacements, such displacements were also regularised to avoid artefacts and distortions. The former minimised the $L_1$ norm of the difference between the gradient of the input and the rendered images, claiming that using image gradients forces the network to focus on geometric details, and regularised them with two terms: one that ensured small values and another one that ensured similarity between displacements of neighbouring vertices. These two terms were also used by \citet{YChenTIP2020} to regularise their reconstructions, but only the image error was used to estimate the fine details. \citet{RichardsonCVPR2017} also used only the image error but, instead of ensuring similar displacements in a neighbourhood, enforced the refined shape to be similar to the coarse one.

\citet{XChaiICME2020} trained their network to transfer the style of the face region in the input image to the texture modelled by a 3DMM, obtaining a realistic skin texture $\mathbf{T}_\text{out}$. To ensure the style transfer, they included another perceptual loss, which penalises differences in content, and a style loss, which penalises differences in style, that is, differences in colours, textures, etc. This second perceptual loss was computed as in eq. \eqref{eq:errorI_featuresL2} but between the 3DMM texture and the output texture, and the style loss was computed as
\begin{equation*}
    \mathbb{E}_{\text{style}}\left(\mathbf{T}_\text{in}^\text{face}, \mathbf{T}_\text{out}\right) = \frac{1}{\abs{\mathcal{C}}} \sum_{j\in\mathcal{C}} \norm{\mathbf{G}_j^\varphi (\mathbf{T}_\text{in}^\text{face}) - \mathbf{G}_j^\varphi (\mathbf{T}_\text{out})}_\text{F}^2,
\end{equation*}
where $\mathbf{T}_\text{in}^\text{face}$ is the visible facial region of the input image, $\norm{\cdot}_\text{F}$ is the Frobenius norm, and $\mathbf{G}_j^\varphi(\mathbf{I})$ is the Gram matrix defined as
\begin{equation*}
    \mathbf{G}_j^\varphi(\mathbf{I}) = \frac{\hat{\varphi}_j(\mathbf{I})^\text{T} \hat{\varphi}_j(\mathbf{I})}{W_j H_j C_j},
\end{equation*}
with $\hat{\varphi}_j(\mathbf{I})$ the activation of the $j$th layer of $\varphi$ (whose shape is $W_j \times H_j \times C_j$) reshaped into a 2D matrix of dimension $W_jH_j \times C_j$ for an input image $\mathbf{I}$.

Although most of the researchers adopted the image error as in eq. \eqref{eq:errorI_selfsup}, there are others that followed a slightly different approach \citep{TewariICCV2017,TewariCVPR2018,TewariPAMI2020,TewariCVPR2019,SavovICCVW2019,KoizumiECCV2020,FWuCVPR2019}. Instead of making a pixel-wise comparison between the input and the rendered images, they compared the input image and the estimated colour for every vertex:
\begin{equation*}
    \errorI = \sum_{i = 1}^N \norm{ \Iin( \hat{\vec{x}}_i ) - \vec{c}_i }_2
\end{equation*}
where $\vec{c}_i$ is the estimated colour of a vertex $\vec{x}_i\in\mathbb{R}^3$, $\hat{\vec{x}}_i \in \mathbb{R}^2$ is its projection onto the image plane, and $\mathbf{I}(x,y)$ the colour of the pixel $(x,y)$ in image $\mathbf{I}$.

Whereas most of them \citep{TewariICCV2017,TewariCVPR2018,TewariPAMI2020,TewariCVPR2019,SavovICCVW2019,FWuCVPR2019} also included the landmarks error (eq. \eqref{eq:errorlmks_DL}) and some regularisation, \citet{KoizumiECCV2020} proposed a landmark-free approach by estimating a correspondences map and a depth map (see Section \ref{subsubsec:DL_EncDec}) that were used to estimate 3DMM parameters. Therefore, to be able to train the network end-to-end, they included the least squared residuals and some regularisation terms for the 3DMM and the camera parameters.

Unlike the other works, \citet{FWuCVPR2019} reconstructed the 3D face of a subject from three images (frontal, left, and right views), thus they had to adapt the image error to take into account the difference between several pairs of images:
\begin{align} \label{eq:errorI_FWu3imgs}
    \errorI =& \norm{\mathbf{I}_\text{in}^B - \mathbf{I}_\text{mod}^{A\to B}}_2^2 + \norm{\mathbf{I}_\text{in}^B - \mathbf{I}_\text{mod}^{C\to B}}_2^2 +\\ &\norm{\mathbf{I}_\text{in}^A - \mathbf{I}_\text{mod}^{B\to A}}_2^2 + \norm{\mathbf{I}_\text{in}^C - \mathbf{I}_\text{mod}^{B\to C}}_2^2 \nonumber
\end{align}
where $\mathbf{I}_\text{in}^A$, $\mathbf{I}_\text{in}^B$, and $\mathbf{I}_\text{in}^C$ are the three input images (right, frontal, and left, respetively), and $\mathbf{I}_\text{mod}^{i\to j}$ represents the estimated 3D face coloured with image $\mathbf{I}_\text{in}^i$ projected into $\mathbf{I}_\text{in}^j$. However, the authors reported that using only the image error does not ensure good alignment between the input and the rendered images, thus they proposed an alignment term that minimised the optical flow between them
\begin{align*}
    \mathbb{E}_\text{align}\left(\mathbf{I}_\text{in}^i, \mathbf{I}_\text{mod}^{j\to i}\right) = \sum_{\mathbf{p}} \bigg[ &\norm{ \mathcal{F}_o\Big(\mathbf{I}_\text{mod}^{j\to i}(\mathbf{p}), \ \mathbf{I}_\text{in}^{i}(\mathbf{p})\Big) }_2^2 +\\
    &\norm{\mathcal{F}_o\Big(\mathbf{I}_\text{in}^{i}(\mathbf{p}), \ \mathbf{I}_\text{mod}^{j\to i}(\mathbf{p})\Big)}_2^2 \bigg]
\end{align*}
where $\mathcal{F}_o(\mathbf{I}^i(\mathbf{p}),\mathbf{I}^j(\mathbf{p}))$ is the optical flow from $\mathbf{I}^i$ to $\mathbf{I}^j$ at pixel $\mathbf{p}$ estimated using \citep{SunCVPR2018}. This alignment error was computed for the same pairs of images as in the image error (eq. \eqref{eq:errorI_FWu3imgs}).

Even though the image error has been widely used to drive the 3D facial reconstruction in a self-supervised manner, there have been works that included the image error as an additional constraint even though the training was supervised \citep{SenguptaCVPR2018,XZengICCV2019,ZGaoCVPRW2020,GHLeeCVPR2020,JShangECCV2020,ZShuFG2020}. In addition, \citet{ZGaoCVPRW2020} used a loss that minimises the difference in the gradients of the images, similarly to \citep{PWangICME2020}, and \citet{GHLeeCVPR2020} computed the cosine feature loss (eq. \eqref{eq:errorI_featuresCos}) between the input image and rendered images with random projection parameters. \citet{GHLeeCVPR2020} claimed that using random projections made their method robust to different viewpoints.

On the other hand, \citet{JShangECCV2020} and \citet{ZShuFG2020} used the image loss to ensure photometric consistency across the multi-view input images similarly to \citet{FWuCVPR2019}. \citet{JShangECCV2020} used two adjacent images, $\Iin^{\text{A}_1}$ and $\Iin^{\text{A}_2}$, to improve the quality of the reconstruction obtained with the input target image $\Iin^\text{T}$, by minimising
\begin{equation} \label{eq:mviewCons_JShangECCV2020}
    \mathbb{E}(\Imod^{\text{A}_j\to\text{T}}, \Iin^\text{T}) = \mathbf{M}\odot\norm{ \Iin^\text{T} - \Imod^{\text{A}_j\to\text{T}}}_1
\end{equation}
where $\Imod^{\text{A}_j\to\text{T}}$ is the synthesised target image from the adjacent image $\Iin^{\text{A}_j}$, $\mathbf{M}$ is the occlusion mask to account for the common visible face region, and $\odot$ the Hadamard product. \citet{ZShuFG2020} highlighted that the same anatomical point may not have the same pixel colour across the different images, thus, to account for this issue, they multiplied the loss in eq. \eqref{eq:mviewCons_JShangECCV2020} by a suitability map $\mathbf{S}(\mathbf{I})$ that encodes the suitability of each pixel to be used for computing the photo-consistency loss. This suitability map was estimated from $\mathbf{I}$ by another network.

\subsubsection{Adversarial loss} \label{subsubsec:AdversarialLoss}
As explained in Section \ref{subsubsec:DL_GANs}, \citep{XTuToM2020,GalteriCVPRW2019,GalteriCVIU2019,ZGaoCVPRW2020,JPiaoICCV2019,JLinCVPR2020} used GANs, which consist of a generator $G$ and a discriminator $D$ that are jointly trained for $D$ to discriminate between real and fake data and for $G$ to fool $D$. Therefore, $D$ is trained to maximise $\log(D(\mathbf{y}^\text{gt})) \in (-\inf,0]$, while $G$ is trained to minimise $\log(1 - D(G(\mathbf{x}))) \in (-\inf,0]$. This is expressed mathematically as the following minimax problem
\begin{equation*}
    \min_G \max_D \ \mathcal{L}_\text{adv}(G,D)
\end{equation*}
where
\begin{align}\label{eq:L_adv}
    \mathcal{L}_\text{adv}(G,D) =& \ \mathbb{E}\Big[\log\big(D(\mathbf{y}_\text{gt})\big)\Big] +\\ & \ \mathbb{E}\Big[\log\big(1 - D(G(\mathbf{x}))\big)\Big]. \nonumber
\end{align}
$\mathbb{E}[X]$ expresses the expected value of a random variable $X$.

In particular, \citet{XTuToM2020} used the GAN architecture to improve the estimation of 3DMM parameters by evaluating the consistency of the predicted 3DMM parameters with the input image. To do so, they took pairs consisting of a latent representation of an image and the corresponding 3DMM parameters. Therefore, eq. \eqref{eq:L_adv} becomes
\begin{align*}
    \mathcal{L}_\text{adv}(G,D) =& \ \mathbb{E}\Big[\log\big(D([\mathbf{f},\vec{\theta}])\big)\Big] +\\ & \ \mathbb{E}\Big[\log\big(1 - D([\widetilde{\mathbf{f}},G(\mathbf{I})])\big)\Big] \nonumber
\end{align*}
where $\mathbf{f}$ is the latent representation of an image with ground truth 3DMM parameters $\vec{\theta}$ and $\widetilde{\mathbf{f}}$ is the latent representation of an image $\mathbf{I}$ with parameters $G(\mathbf{I})$ synthesised by the generator.

\citet{GalteriCVPRW2019,GalteriCVIU2019}, \citet{ZGaoCVPRW2020} and \citet{JLinCVPR2020} trained a GAN where the discriminator had to distinguish real, $\Iin$, from synthetic images, $\Imod$. They adopted the paradigm of Improved Wasserstein GAN \citep{GulrajaniNICP2017_WassersteinGANs}, which, according to the authors, allows for a more stable training. In this paradigm, the adversarial loss is defined as
\begin{align}\label{eq:adversarial_Wass}
    \mathcal{L}_\text{adv}(G,D) =& \ \mathbb{E}\Big[D(\Iin)\Big] - \mathbb{E}\Big[D(\Imod)\Big] +\\ & \
    + \lambda \mathbb{E}\Big[\norm{\nabla_{\vec{\hat{I}}} D(\vec{\hat{I}})}\Big] \nonumber
\end{align}
where $\vec{\hat{I}}$ is uniformly sampled between pairs of points from the distribution of real and generated images. Since \citet{ZGaoCVPRW2020} used the discriminator to distinguish real from fake 3D faces and not images, in eq. \eqref{eq:adversarial_Wass}, $\Iin$ is the real 3D face, $\Imod$ is the estimated 3D face, and $\mathbf{\hat{I}}$ are 3D faces uniformly sampled between pairs of 3D faces from the real and fake distributions.

In contrast, \citet{JPiaoICCV2019} used a CycleGAN to map the real images to synthetic-style images and vice-versa. The CycleGAN is composed of two GANs concatenated forming a cycle (see eq. \eqref{eq:cycleGAN}) and it is trained following the same idea of the simple GANs but, additionally, a cycle-consistency loss has to be taken into account. This loss function enforces both generators to be able to recover the original image after performing a cycle. That is, let $G_\text{r}: \mathcal{S} \to \mathcal{R}$ and $G_\text{s}:\mathcal{R}\to\mathcal{S}$ be both generators, with $\mathcal{R}$ and $\mathcal{S}$ the space of the real and the synthetic images, respectively; then, the cycle-consistency loss is defined as
\begin{align*}
    \mathcal{L}_\text{cyc} (G_\text{s},G_\text{r}) = & \ \mathbb{E}\Big[ \norm{ G_\text{r}\big(G_\text{s}(\mathbf{I}_\text{r})\big) - \mathbf{I}_\text{r}}_1 \Big] +\\ &\ \mathbb{E}\Big[ \norm{ G_\text{s}\big(G_\text{r}(\mathbf{I}_\text{s})\big) - \mathbf{I}_\text{s}}_1 \Big]
\end{align*}
where $\mathbf{I}_\text{r}, G_\text{r}(\mathbf{I}_\text{s}) \in \mathcal{R}$ and $\mathbf{I}_\text{s}, G_\text{s}(\mathbf{I}_\text{r})\in\mathcal{S}$. Also, \citet{JPiaoICCV2019} noticed that the adversarial losses (eq. \eqref{eq:L_adv}) for each GAN and the cycle-consistency loss are not sufficient to avoid artefacts and large deformations in the image domain-transfer results. Therefore, to further constrain the image mapping between the real and synthetic image spaces, they considered a landmarks error defined as
\begin{align*}
    \mathcal{L}_\text{lmks}(G_1,G_2) = & \ \mathbb{E}\Big[ \norm{ \vec{\ell}^{\text{2D}}_{\mathbf{I}_\text{s}} - \vec{\ell}^{\text{2D}}_{G_\text{s}(\mathbf{I}_\text{r})} }_2 \Big] + \\ & \ \mathbb{E}\Big[ \norm{ \vec{\ell}^{\text{2D}}_{\mathbf{I}_\text{r}} - \vec{\ell}^{\text{2D}}_{G_\text{r}(\mathbf{I}_\text{s})} }_2 \Big]
\end{align*}
where $\vec{\ell}^\text{2D}_\mathbf{I}$ denotes the landmark coordinates estimated in the image $\mathbf{I}$. %Finally, the 3D face reconstruction network, which inputs a synthetic-style image and recovers the position UV-map (see eq. \eqref{eq:cycleGAN}), was trained by minimising three loss functions: one that computed the distance between the resulting and the ground truth position UV-maps, another one that computed the scalar product between the ground truth normals and the vectors formed by adjacent vertices, and a landmarks distance term.

\subsubsection{Other Losses} \label{subsubsec:DL_otherLosses}

Whereas most of the used losses can be classified into one of the above categories, other losses have been proposed. \citet{PWangICIP2019} used exclusively the landmarks error (eq. \eqref{eq:errorlmks_DL}) but benefited from the variety of landmark databases to increase the number of anchor points by developing a dataset fusion method. \citet{RamonICCVW2019} and \citet{YGuoPAMI2018} extended the widely used landmarks error to all the vertices in the 3D face, computing the distance between the projection of the estimated and the ground truth 3D faces:
\begin{equation}\label{eq:error_proj3D}
\mathbb{E}_\text{proj 3D}(\vec{\alpha},\vec{\zeta}) = \norm{ \mathcal{P}\left(\vec{x}(\vec{\alpha}), \vec{\zeta}) - \mathcal{P}(\vec{x}_\text{gt},\vec{\zeta}_\text{gt}\right) }_2^2
\end{equation}
where $\mathcal{P}(\vec{x},\vec{\zeta})$ represents the projection of a 3D face $\vec{x}$ according to the projection parameters $\vec{\zeta}$, and $\vec{x}(\vec{\alpha})$ is the recovered 3D face from the estimated 3DMM shape parameters $\vec{\alpha}$. Since \citet{RamonICCVW2019} used multiple images, their loss function was the sum of the losses for each of the images. On the other hand, \citet{YGuoPAMI2018} separated the error produced by incorrect geometries and by incorrect projection parameters to be able to control their effect individually, and also for better convergence:
\[
\mathbb{E} = w \mathbb{E}_\text{proj} + (1-w)\mathbb{E}_\text{geo}
\]
where
\begin{align*}
&\mathbb{E}_\text{proj} = \norm{\mathcal{P}(\vec{x}(\vec{\theta}^\text{geo}_\text{gt}),\vec{\zeta}) - \mathcal{P}(\vec{x}(\vec{\theta}^\text{geo}_\text{gt}),\vec{\zeta}_\text{gt})}^2_2 \text{ and}\\
&\mathbb{E}_\text{geo} = \norm{\mathcal{P}(\vec{x}(\vec{\theta}^\text{geo}),\vec{\zeta}_\text{gt}) - \mathcal{P}(\vec{x}(\vec{\theta}^\text{geo}_\text{gt}),\vec{\zeta}_\text{gt})}^2_2.
\end{align*}
with $\vec{\theta}_\text{geo}$ the parameters corresponding to the facial geometry.

\citet{YFengECCV2018} and \citet{GulerCVPR2017} adopted an UV-based representation of the 3D face, and thus used an according loss function. Specifically, the former \citep{YFengECCV2018} estimated position UV-maps (see Section \ref{subsubsec:DL_EncDec}) by minimising the Euclidean distance between the ground truth and the estimation, adding a mask that assigns a weight to each pixel in the map to penalise errors on the facial regions that have more discriminative features. In contrast, \citet{GulerCVPR2017} discretised each dimension of the UV-space, therefore dividing each ground truth $(u,v)$ location into a discrete part, which is defined as the nearest lower integer, and a residual part, which is defined as the difference between the discrete part and the ground truth value. In this way, they combined a classification problem to estimate the discrete part, and a regression problem to estimate the residual part. For the former, they used the softmax cross-entropy loss, and for the latter the smoothed version of the $L_1$ norm proposed by \citep{GirshickICCV2015_smoothL1}.

As explained in Section \ref{subsubsec:DL_EncDec}, \citet{SongTIP2012} trained two encoders to extract the same features from 2D images and 3D faces, and then used the ``reversed'' 3D-encoder to obtain a decoder that regresses a 3D face from the features estimated by the 2D-encoder. Therefore, they used a reconstruction loss that ensures that both networks can be stacked together by minimising the difference between the input of one and the output of the other. Also, to reduce the space of possible solutions, they added a smoothness regularisation term and a feature consistency term.

Finally, \citet{JShangECCV2020} proposed an epipolar loss to account for multi-view consistency. This loss is computed as the symmetric epipolar distance between the 2D landmarks of the input image and the corresponding 2D landmarks in the adjacent-view images.

\subsection{Take-home message}
% 1. DL is awesome but needs large amount of data.
% 2. In 3DFR such amount of data is not available.
% 3. Synthetic training sets, but then the networks does not learn outside the 3DMM.
% 4. New trend with many followers: self-supervision.
% 5. Drawbacks of self-supervision: texture & restrict solution -> 3DMM + drawbacks

% 6. Lack of training data --> not recover details
% 8. Focus on recovering facial details.
% 9. Multi-view, iterative training, divide and conquer
% 10. But results show that details have to be estimated separately from the coarse shape. --> a posteriori with SfS, coarse-to-fine, or recently GANs.
% Most of the promising lines of research depart from the 3DMM...
% 11. Some departed from 3DMM by estimating the space of faces with a net -> non linear face model.
% 12. 3DMM are also very popular because they allow reconstructing a 3D facial surface (mesh) using the classical CNNs --> GCN

Deep learning has proven to be a very powerful tool applicable in many different fields, gaining the attention of great part of the research community, and 3D-from-2D face reconstruction is not an exception. In the last few years, a large amount of works have proposed to reconstruct the 3D facial geometry of a person from one or more images, or even video sequences, using deep learning, as can be seen in Tables \ref{tab:DL1}, \ref{tab:DL2}, and \ref{tab:DL3}.

However, deep learning has an important drawback: the need of huge training sets, in this case, composed of corresponding pairs of 2D images and 3D faces. Since obtaining this amount of ground truth 3D facial scans is not always feasible, the community has explored two strategies to overcome this limitation: building synthetic training sets, and self-supervision. The first one focused on building synthetic data by fitting a 3DMM to real images \citep{XZhuCVPR2016} and/or rendering synthetic images from 3D faces obtained by sampling from the 3DMM \citep{Richardson3DV2016}. Even though these approaches allow for the construction of large heterogeneous training sets, reconstructing 3D faces outside the distribution learnt by the 3DMM may be a difficult task, given that the network is not taught with anything but 3DMM-generated faces.

The second strategy, that is based on self-super\-vision, is growing very fast since it avoids the need of ground truth 3D faces. This self-supervised approach consists in adding a rendering layer at the end of the network that renders a synthetic image from the reconstructed 3D face. Therefore, in the training phase, the difference between the input and the rendered images is minimised so that the network learns to generate 3D faces that produce synthetic images identical to the input images.

Although this approach bypasses the need of paired 2D-3D data, it has two weaknesses. Firstly, the texture of the 3D face must be accurately recovered, which is a difficult task in its own and, in some applications, it is not necessarily of interest. Secondly, the training process has to be carefully regularised using additional features, since minimising only the difference between the images does not guarantee a correct reconstruction of the 3D geometry, due to the ill-posed nature of the problem. In fact, many works have used 3DMMs as prior knowledge to restrict the space of possible solutions obtained with the self-supervised training, thus suffering from similar limitations to those described above for works using synthesised training data.

%. However, even though 3DMMs have greatly boosted the 3D-from-2D face reconstruction field, modelling non-linear shape variations, like facial deformations, with linear models causes fine details to be lost. Consequently, restricting the reconstructions to the 3DMM space may lead to coarse or oversmooothed 3D faces.

In addition, regardless of the training strategy, the shortage of real ground truth 3D facial scans entails another limitation: the reconstruction of fine details. Such details are the ones responsible of making each person unique, and therefore are essential to recover an accurate 3D face. However, synthetic data is not as detailed as real 3D facial scans, since it is generated with a 3DMM, while the self-supervised training strategy learns from real 2D images, which do exhibit facial details, but not in 3D, making their geometric reconstruction very difficult.

%However, if the network cannot learn from real 3D data, it will not be able to recover subtle details, since they are not present in the training dataset.

%

% For this reason, many works have used 3DMM as prior knowledge to restrict the space of possible solutions to those represented by the facial model. However, in spite of the fact that 3DMMs have greatly boosted the 3D-from-2D face reconstruction field, modelling non-linear shape variations, like facial deformations, with linear models causes fine details to be lost. Consequently, restricting the reconstructions to the 3DMM space may lead to coarse 3D faces, not recovering fine details that make a person unique.

As a result, the reconstruction of the fine details has become another big challenge for the community. Some works have explored the benefits of multiple views, and the iterative refinement of the resulting face. Also, many works followed a ``divide and conquer'' strategy by proposing architectures composed of several networks where each of them estimates a single attribute (identity, expression, albedo, lighting, pose, etc.). In this way, each stream can focus on a single task, improving accuracy.

% A different approach that has been widely followed is to recover the fine details ``a posteriori'', once a coarse 3D face is estimated.
Nevertheless, current results suggest that the subtle details have to be estimated separately from the coarse shape. Some works have estimated the details using already presented methods based on shape-from-shading such as \citep{BeelerToG2010,OrElCVPR2015}, whereas others have integrated several networks sequentially where each of them refines the output of the previous one. Also, very recently, there have been researchers that have used generative adversarial networks to force the generator to produce more accurate 3D facial reconstructions.

Even so, the most promising lines of research depart from the 3DMM either by estimating non-linear facial models, or by adopting geometric deep learning methods. On the one hand, the estimation of non-linear  models has been widely explored, mostly, by using encoder-decoder architectures. Under this approach, the network learns to regress parameters that are related to different elements of the 3D face (shape, expression, albedo, pose, etc.) from the input image, and then to recover the 3D face from these parameters. Given the ability of neural networks to capture non-linearities, with this approach facial details can be modelled, and therefore reconstructed on the resulting 3D face.

% On the other hand, in the past few years, the community has begun to explore methods that allow reconstructing directly 3D facial meshes, which cannot be done using classical deep learning methods due to the non-Euclidean nature of meshes. Early works have adopted different Euclidean representations, such as discrete volumes, depth maps, 2D representations, or 3DMM parameters. However, all of them need an additional step to recover the 3D facial surface from the alternative representation estimated by the network, and/or the chosen representation is not sufficiently accurate so as to obtain detailed 3D faces. Consequently, recent works are using geometric deep learning, which is a field that studies the adaptation of common mathematical operations in deep learning, such as pooling or convolutions, to non-Euclidean data like 3D facial meshes, thus avoiding the need of an alternative representation of the 3D face.

On the other hand, geometric deep learning studies the adaptation of common mathematical operations in deep learning, such as pooling or convolutions, to non-Euclidean data like 3D facial meshes, thus allowing designing networks that address the direct reconstruction of 3D facial meshes without the need for alternative, and often limiting, representations of the 3D face.

\section{Other Machine Learning Methods} \label{sec:OtherML}
% that are not DL and do not use 3DMM (i.e. do not fit a 3DMM)
Although deep learning has been a widely used tool in 3D-from-2D face reconstruction, other machine learning methods have also been explored. Most of them approached 3D-from-2D face reconstruction as a regression problem, either by learning a single regressor  \citep{SanchezCVIM2016,RaraICIP2010,RaraIJCB2011,DouBMVC2014} or a cascade of regressors \citep{TianFG2018,JeniFG2015,JeniIVC2017,FLiuECCV2016,FLiuFITEE2017,FLiuPAMI2020}.

\citet{SanchezCVIM2016} trained a regression matrix $\vec{\mathcal{R}}$ to estimate a 3D facial surface $\vec{x}$ from the contour of the input image $\vec{c}$ as  $\vec{x}=\vec{\mathcal{R}c}$. To do so, they concatenated the $M$ vectorised 3D training shapes into the response matrix $\vec{X} \in \mathbb{R}^{M\times3N}$, and concatenated the pixel locations of the $L$ 2D contours of the training images into the predictor matrix $\vec{C} \in \mathbb{R}^{M\times 2L}$. Then, the regression matrix was estimated by solving $\vec{X} = \vec{C\mathcal{R}}$. Even though this approach is very fast, its reconstruction power is limited because only contours are used as facial shape information to reconstruct the 3D geometry. 

On the other hand, \citet{RaraICIP2010} trained two regression models, one to recover the 3D face and another one to recover the facial texture. More precisely, the shape regressor estimated parameters of a PCA model built from the projection of a set of 3D textured faces into the spherical harmonics space (eq. \eqref{eq:LambertianModel_SH}), and equivalently for the texture. With this method, \citet{RaraICIP2010} achieved low computational cost, however, they assumed frontal poses and a Lambertian reflectance model, which limits the reconstruction accuracy in unconstrained contexts.

Differently from \citep{SanchezCVIM2016,RaraICIP2010}, \citet{RaraIJCB2011} built two statistical models; one that encoded the variation of the position of the 2D landmarks in a face, and another one that encoded the variation of the 3D landmarks. Thus, given the coefficients of the 2D-landmarks-model, their proposed regressor estimated coefficients of the 3D-land\-marks-model, which were mapped back to $\mathbb{R}^3$. To obtain dense 3D shapes, they replaced the sparse 3D-landmarks-model with a model built from dense 3D shapes. \citet{DouBMVC2014} extended this work \citep{RaraIJCB2011} by recovering the dense 3D face from the landmarks with an additional coupled dictionary model built from corresponding pairs of sparse 3D landmarks and dense point sets. This additional step allowed \citet{DouBMVC2014} to obtain more accurate reconstructions than \citep{RaraIJCB2011}.

As stated above, a second group of works \citep{TianFG2018,JeniFG2015,JeniIVC2017,FLiuECCV2016,FLiuFITEE2017,FLiuPAMI2020} proposed a cascade of linear regressors to estimate the 3D shape, instead of a single regressor as the works above. 
\citet{TianFG2018} trained a cascade of regressors $\vec{\mathcal{R}}^k$ to predict the shape update $\vec{\Delta}\vec{x}^{k}$ according to the difference between the 2D landmarks in the input image, $\lmk$, and the 3D landmarks of the previous shape projected into the image $\vec{\hat{\ell}}^{\text{ 2D},k-1}$, $\vec{x}^{k} = \vec{x}^{k-1} + \vec{\Delta x}^{k} = \vec{x}^{k-1} + \vec{\mathcal{R}}^{k}\left(\lmk - \vec{\hat{\ell}}^{\text{ 2D},k-1}\right)$. The regressors were trained by minimising the Euclidean distance between the ground truth shape update $\vec{\Delta}\vec{x}^{k}_\text{gt} = \vec{x}_\text{gt} - \vec{x}^{k-1}$ and the estimated shape update by the current regressor over the $M$ training pairs, 
\begin{equation*}
 \vec{\mathcal{R}}^k = \argmin_{\vec{\mathcal{R}}} \sum_{m=1}^M \norm{\vec{\Delta}\vec{x}^{k}_{m,\text{gt}} - \vec{\mathcal{R}}\left(\lmk_m - \vec{\hat{\ell}}^{\text{ 2D},k-1}_m\right)}_2^2.
\end{equation*}
Although this method is able to exploit the information provided from a variable number of input images and is robust to image quality and expressions, it performs less accurately than state of the art methods in case of large pose variations.

Contrary to \citep{TianFG2018}, who takes as input the position of the 2D landmarks, 
\citet{JeniFG2015,JeniIVC2017} and \citet{FLiuECCV2016,FLiuFITEE2017,FLiuPAMI2020} estimated the 2D landmarks in the input image along with the 3D shape. The former \citep{JeniFG2015,JeniIVC2017} used as input a 2D video, reconstructing the 3D face for each frame. They trained a cascaded regressor to estimate a dense set of facial points in the 2D frame, which were in correspondence with the vertices of a 3D face model. Then, the 3D face model was fitted to the frame by minimising the Euclidean distance between the projection of the model vertices and the estimated 2D facial points. Unlike \citep{JeniFG2015,JeniIVC2017}, \citet{FLiuECCV2016,FLiuFITEE2017,FLiuPAMI2020} trained a separate regressor to estimate the 2D landmark locations, which was alternated with the reconstruction of the 3D face. The landmarks regressor updated the 2D landmarks from image features extracted around the estimated 2D landmarks. Then, the shape regressor estimated the update of the shape according to the landmarks update. Whereas in their preliminary work \citep{FLiuECCV2016,FLiuFITEE2017}
Liu \emph{et al.} estimated the face shape update jointly for the identity-related and expression-related deformations, in their latter work \citep{FLiuPAMI2020}, they trained the shape regressor to estimate the identity and the expression updates as separated elements such that $\vec{x}^k = \vec{x}^{k-1} + \vec{\Delta x}_\text{id}^k + \vec{\Delta x}_\text{exp}^k$. With this approach, \citet{FLiuPAMI2020} were able to substantially reduce the reconstruction errors.

Differently from all the previous works, \citet{SWangICIP2018} addressed 3D face reconstruction from an uncalibrated 2D video as a minimisation problem based on dense optical flow. First, a personalised 3D template was obtained from the collection of frames with \citep{RothCVPR2015}, and the method by \citep{GargIJCV2013} was applied to estimate a dense 2D optical flow of the video. With these two elements, the following energy function was used to translate the 2D optical flow to the personalised template,
\begin{equation*}
    \mathbb{E}(\vec{f},\vec{\rho}) = \sum_{i = 1}^N \norm{ \mathcal{P}\Big(\vec{v}_i + \vec{f}(\vec{v}_i),\vec{\rho}\Big) - \mathcal{F}_o\left(\mathcal{P}\Big(\vec{v}_i,\vec{\rho}\right)\Big) }_\epsilon
\end{equation*}
where $\{\vec{v}_i\}_{i=1}^N$ is the set of 3D vertices of the personalised 3D facial template, $\mathcal{P}(\vec{v},\vec{\rho})$ is the projection of $\vec{v}$ according to projection parameters $\vec{\rho}$, $\norm{\cdot}_\epsilon$ is the Huber loss, and $\vec{f}(\vec{v}) = (\vec{\Delta}v^x,\vec{\Delta}v^y,\vec{\Delta}v^z)\in\mathbb{R}^3$ is the per-vertex displacement estimated from the 2D optical flow $\mathcal{F}_o$. Further, two additional terms were added to ensure local smoothness of the deformations $\vec{f}(\cdot)$, and a fourth term to regularise the deformations in the temporal domain. According to the authors, the dense optical flow approach is able to improve the reconstructions for large expressions, avoiding the need to rely on the quality of pre-built facial models.

\section{Applications}\label{sec:Applications}
The possibility of reconstructing the 3D face of a person from uncalibrated image(s) has granted a tool for many applications to estimate 3D information from facial data, avoiding the need for specialised machinery or controlled environments, which are not always feasible. In this section, we show the impact of 3D-from-2D face reconstruction by providing a brief review of some of these applications.

% Aplications: en qué consiste, cómo aplica el 3DFR, citas

\emph{\textbf{Facial recognition.}} The goal of a facial recognition system is to ``recognise'' a person from an image, that is, to match the face from the picture against a database of faces. Traditionally, this technology has been developed using 2D images, given the simplicity of their acquisition. However, the facial appearance of the same person may greatly vary across images, depending on factors such as the pose in which the picture is taken, the scene illumination and the expression of the person, making the recognition task challenging. In contrast, the 3D facial geometry is invariant to the pose and the illumination conditions, and thus it can add robustness to the recognition. As a consequence, and given that obtaining 3D facial data is not always feasible, the community has explored facial recognition approaches based on 3D face reconstruction from uncalibrated images by fitting a 3DMM \citep{BlanzVetterPAMI2003,BlanzFG2002,YHuFG2004,BChuCVPR2014,JZhaoIJCAI2018}, using deep learning methods \citep{FLiuCVPR2018,TranCVPR2017,FLiuICCV2019,JGuoCCBR2018}, or training a cascaded regressor \citep{FLiuPAMI2020}, among others.

% In fact, some of the works presented above \citep{FLiuPAMI2020,FLiuCVPR2018,FLiuICCV2019,TranCVPR2017} proposed a 3D reconstruction method with the aim of improving the face recognition. Also, from the 3DMM fitting perspective, 

% the researchers that presented the 3DMM to the community, explored the applicability of the 3DMM fitting for face recognition in \citep{BlanzVetterPAMI2003,BlanzFG2002}. This approach was also explored by \citet{YHuFG2004} and \citet{JZhaoIJCAI2018}, by minimisng only the landmarks error, and by \citet{BChuCVPR2014}, who adopted the fitting method proposed by \citep{MatthewsIJCV2007}. In contrast, the works \cite{FLiuCVPR2018,FLiuICCV2019,TranCVPR2017,JGuoCCBR2018} adopted an approach based on deep learning. the former three are presented above (see Section \ref{sec:DL}), and \citet{JGuoCCBR2018} adopted the deep learning method proposed by \citet{XZhuCVPR2016}, which is also presented above. 

\begin{figure}[h!]
    \centering
    \includegraphics[width = 0.9 \columnwidth]{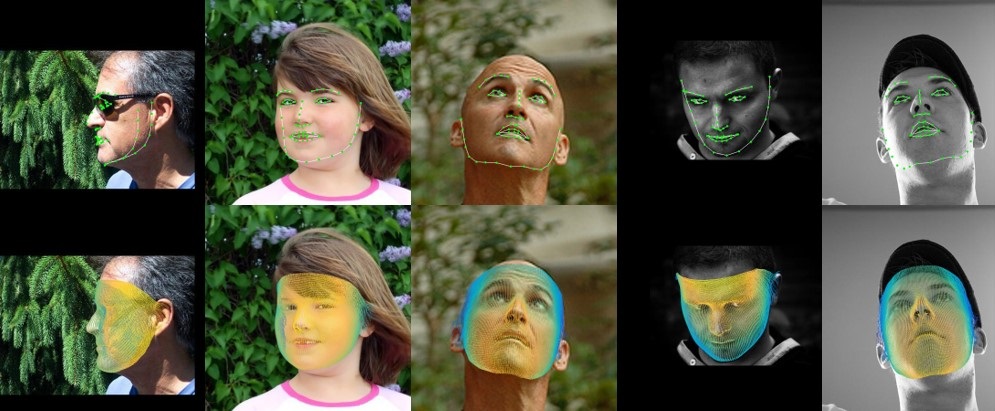}
    \caption{Face alignment examples extracted from \citep{YFengECCV2018}. \emph{First row}: Visible facial landmark points in the input image. \emph{Second row}: 3D facial reconstruction estimated from the input image.}
    \label{fig:FaceAlignment}
\end{figure}

\emph{\textbf{Face alignment.}} Face alignment aims at locating facial landmark points in a 2D image. This field has been widely studied without the need of the 3D facial representation, obtaining satisfactory results in frontal images. However, when the pose deviates from the frontal view, not all landmarks are visible due to self-occlusions. This obstacle has been overcome, to some extent, for moderate pose variations, where non-contour landmarks are still visible, by changing the semantic meaning of contour landmarks to silhouette landmarks \citep{XZhuCVPR2016}. However, for larger out-of-plane rotations, where half of the face is occluded, this 2D approach fails. As a result, many researchers have addressed face alignment jointly with 3D-from-2D face reconstruction \citep{BhagavatulaICCV2017,FLiuECCV2016,GZhangFG2018,JourablooCVPR2016,JourablooICCV2017,JourablooIJCV2017,XTuToM2020,XZhuCVPR2016,XZhuPAMI2019,YFengECCV2018}. By reconstructing the 3D facial geometry from the 2D image, out-of-plane rotations can be easily estimated; thus, the location of the facial landmarks in the 2D image can be determined by projecting the landmarks in the 3D face into the image plane, as shown in Figure \ref{fig:FaceAlignment}.

% FACE ALIGNMENT

\begin{figure}[h!] % FACE ANIMATION
    \centering
    \begin{subfigure}{\columnwidth}
         \centering
             \includegraphics[width = .8 \columnwidth]{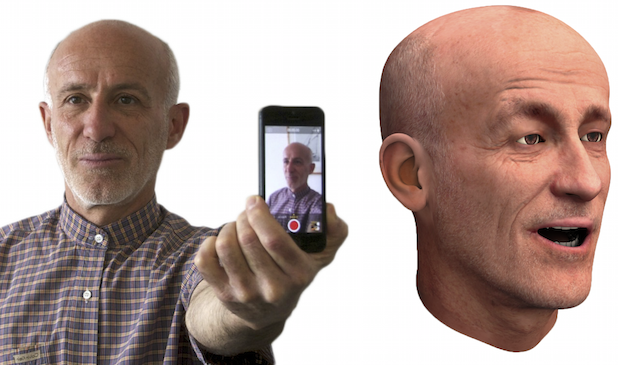}
         \caption{}
         \label{fig:FaceAnimation_AEIchimTOG2015}
     \end{subfigure}
     
    %  \vspace{1em}
     
     \begin{subfigure}{\columnwidth}
         \centering
             \includegraphics[width = .8 \columnwidth]{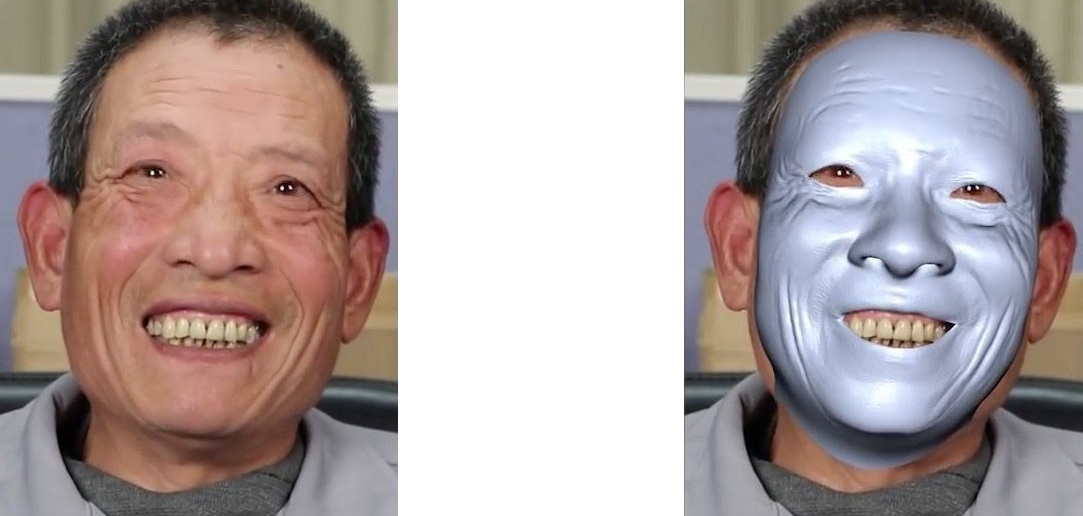}
         \caption{}
         \label{fig:FaceAnimation_CCaoTOG2015}
     \end{subfigure}
     
    %  \vspace{1em}
     
     \begin{subfigure}{\columnwidth}
         \centering
             \includegraphics[width = .8 \columnwidth]{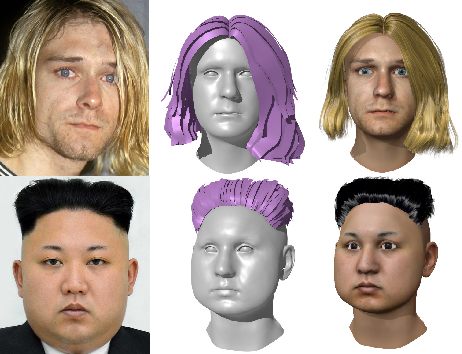}
         \caption{}
         \label{fig:FaceAnimation_LHuTOG2017}
     \end{subfigure}
     
     \caption{Examples of facial animations obtained using 3D-from-2D face reconstruction. (a) Personalised avatar with high-fidelity texture extracted from \citep{AEIchimTOG2015}. (b) Highly detailed avatar extracted from \citep{CCaoTOG2015}. (c) 3D avatar where the hair has also been recovered, extracted from \citep{LHuTOG2017}.}
     \label{fig:FaceAnimation}
\end{figure}

\emph{\textbf{Face animation.}} Reconstructing the 3D facial geometry and the texture from a picture also enables the possibility of constructing personalised avatars, which can be used in areas such as gaming, movies, and virtual reality. Whereas some works focused on obtaining high-fidelity textures \citep{AEIchimTOG2015}, others focused on reconstructing subtle facial details \citep{FShiTOG2014,CCaoTOG2015}, or on reconstructing a complete 3D head, including the hair \citep{CCaoTOG2016,LHuTOG2017}. Figure \ref{fig:FaceAnimation} shows some examples of 3D animations obtained using 3D-from-2D face reconstruction methods.

\begin{figure}[h]
    \centering
    \begin{subfigure}{\columnwidth}
         \centering
             \includegraphics[width = 0.8 \columnwidth]{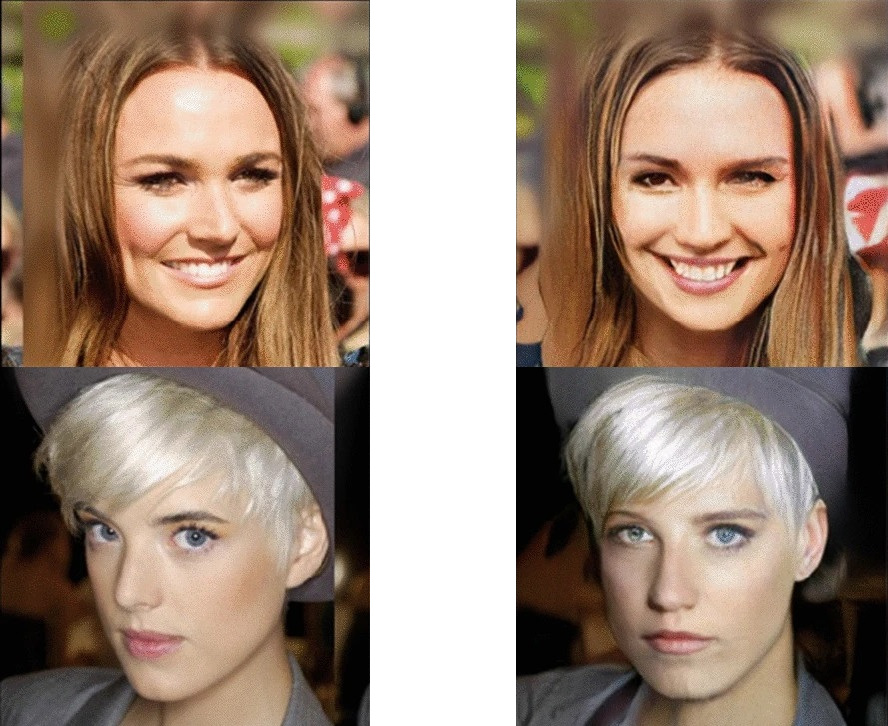}
         \caption{}
     \end{subfigure}
     
     \vspace{1em}
     
     \begin{subfigure}{\columnwidth}
         \centering
             \includegraphics[width = \columnwidth]{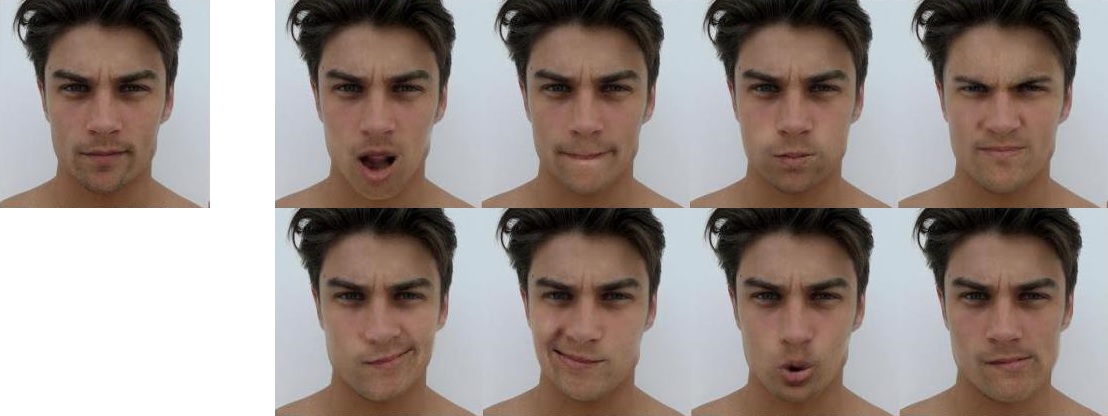}
         \caption{}
     \end{subfigure}

     \caption{Examples of image editing applications using 3D-from-2D face reconstruction. (a) Face frontalisation examples extracted from \citep{JCaoIJCV2020}. \emph{Left}: Original image. \emph{Right}: Frontalised image. (b) Synthesised facial expressions (right) from the input image (left), extracted from \citep{ZGengCVPR2019}.}
     \label{fig:ImageEditing}
\end{figure}

\emph{\textbf{Image editing.}} Estimating all the components participating in the image formation process, such as the 3D facial geometry, albedo, illumination, pose, etc. allows generating new synthetic images by editing one or more of them. For example, face frontalisation is the task of generating a synthetic frontal image from the estimated textured 3D face by modifying the pose to frontal view \citep{XZhuCVPR2015,XYinICCV2017,JCaoNIPS2018,JCaoIJCV2020}. Other examples of image editing are manipulation of the expression \citep{BlanzCGF2003,ZGengCVPR2019,LZhangCGI2005} and manipulation of the illumination conditions \citep{LZhangCGI2005}. These image editing methods are widely used as data augmentation techniques, but also as pre-processing procedures in other applications such as face recognition. Examples of some of these image editing techniques are shown in Figure \ref{fig:ImageEditing}.
% IMAGE EDITING

\begin{figure}[ht]
    \centering
    \begin{subfigure}{\columnwidth}
         \centering
         \includegraphics[width = .9 \columnwidth]{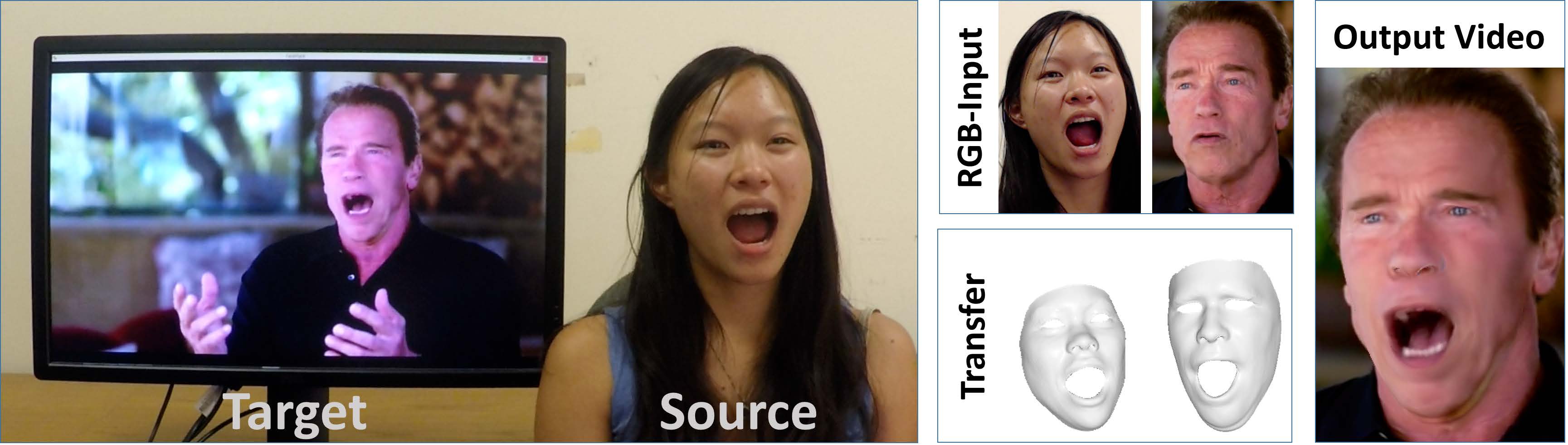}
         \caption{}
     \end{subfigure}
     
     \vspace{1em}
     
     \begin{subfigure}{\columnwidth}
         \centering
        %  \includegraphics[width = \columnwidth]{Figures/ChaudhuriCVPR2019_FaceReenactment2.png} \\
        %  \vspace{0.5em}
         \includegraphics[width = .9 \columnwidth]{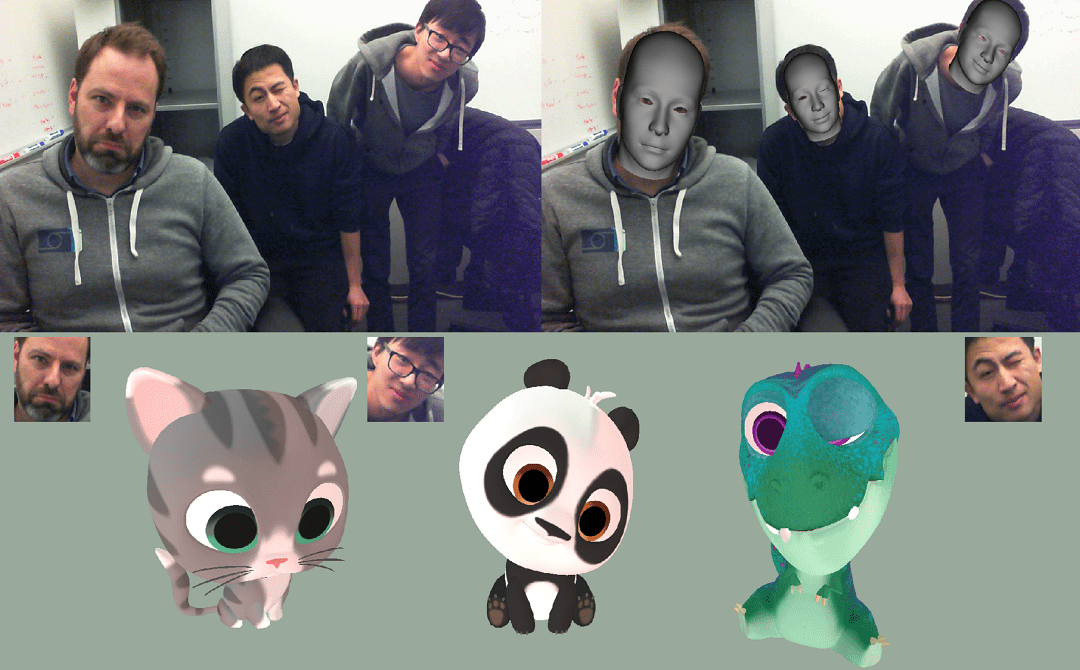}
         \caption{}
     \end{subfigure}
     
     \vspace{1em}
     
     \begin{subfigure}{\columnwidth}
         \centering
         \includegraphics[width = .9 \columnwidth]{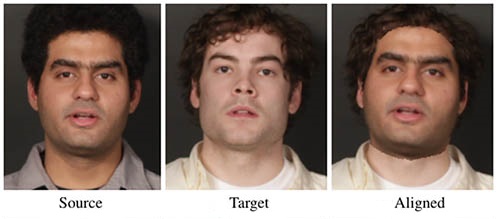}
         \caption{}
     \end{subfigure}

     \caption{Examples of facial reenactment using 3D-from-2D face reconstruction. (a) Video editing by transferring the expression, extracted from \citep{ThiesCVPR2016}. (b) Animation of a 3D avatar transferring expression and pose, extracted from \citep{ChaudhuriCVPR2019}. (c) Face replacement extracted from \citep{DaleTOG2011}.}
     \label{fig:FaceReenactment}
\end{figure}

\emph{\textbf{Facial reenactment.}} Facial reenactment, or facial motion re-targeting, consists in capturing the facial appearance of a person (source) and transferring some of the estimated features (expression, pose, etc.) to another person (target). In this way, the image of the target subject is modified so that he/she exhibits the chosen characteristic of the subject in the source image. Transferring the expression (facial motion) allows modifying a video sequence of a subject so that, for example, it matches the dubbed audio \citep{GarridoTOG2016,ThiesCVPR2016,VlasicTOG2005}. It can also be used to animate an artist-designed 3D character \citep{ChaudhuriCVPR2019,ChaudhuriECCV2020}, or to replace the face of the target subject with the face of the source subject \citep{DaleTOG2011}. Figure \ref{fig:FaceReenactment} shows some examples of these applications.
% IMAGES FROM THE PAPERS

Even though the applications introduced above are the most explored ones, the reconstruction of the 3D face from 2D images has also been used in other applications, such as anti-spoofing \citep{ZYuCVPR2020,ZWangCVPR2020}, age estimation \citep{SavovICCVW2019}, or craniofacial analysis with medical diagnostic purposes \citep{LTuMICCAI2018,LTuMICCAI2019,AlomarVISAPP2021}.

% Others as anti-spoofing, medical, age estimation. Some also took profit of the 3D data to virtual make-up, although not 3D-from-2D face reconstruction.

\section{Summary and Conclusions} \label{sec:Conclusions}

\begin{figure*}[t]
    \centering
    \includegraphics[width=\textwidth]{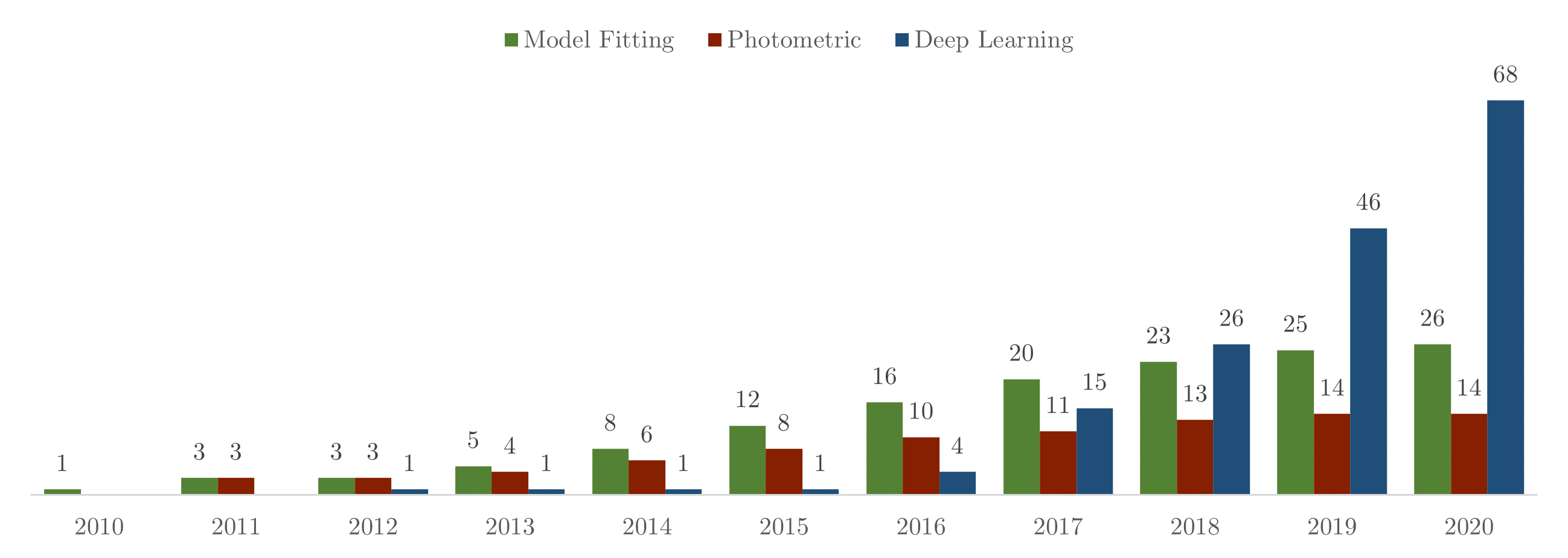}
    \caption{Cumulative number of 3D-from-2D face reconstruction publications in the last decade within the different strategies: statistical model fitting (Section \ref{sec:ModelFittingMethods}), photometric stereo (Section \ref{sec:Photometry}), and deep learning (Section \ref{sec:DL}).}
    \label{fig:CumNumOfPapers}
\end{figure*}

% {\color{red}When using a 3DMM, if it does not model expression, i.e., is build with neutral-expression faces, it is a limitation.}

In this paper, we review the state-of-the-art of 3D-from-2D face reconstruction in the last decade, separating the different proposed approaches into three main strategies: 3D facial model fitting (Section \ref{sec:ModelFittingMethods}), photometric stereo (Section \ref{sec:Photometry}), and deep learning (Section \ref{sec:DL}). Figure \ref{fig:CumNumOfPapers} illustrates the evolution of the field by showing the cumulative number of published papers for each of these strategies.

Recovering the 3D face from uncalibrated 2D pictures is an underdetermined problem, that is, an infinite number of 3D faces can generate the same 2D image if no additional information is known a priori, such as the camera location or the lighting conditions. Therefore, further constraints are needed to reconstruct unequivocally the 3D facial geometry (and, in some cases, the texture), either by making reasonable assumptions, or by including prior knowledge in the reconstruction process.

Traditionally, one of the non-invasive techniques that is used to capture the 3D facial geometry is photometric stereo, where, under a completely controlled environment, the surface normals are estimated by observing how different illumination conditions affect the appearance of the object. Given that these controlled environments are not always feasible, the community has developed a new branch within the photometric stereo field to reconstruct the 3D face from uncalibrated images, i.e., from images obtained under uncontrolled conditions, a.k.a \emph{in the wild}. However, in this scenario, the reconstruction problem is again ill-posed, hence additional constraints have to be added. A widely explored approach to do so is the use of multiple images, which allow researchers to have many samples of lighting conditions. These methods are able to reconstruct highly detailed 3D faces, although the assumptions they consider, such as Lambertian reflectance, are often too restrictive, and the estimated 3D faces can show artefacts since no prior knowledge regarding the geometry is included. Consequently, other works have proposed to restrict the solutions by fitting a pre-designed 3D facial template, but they are highly dependant on the characteristics of the chosen template (age, gender, ethnicity, etc.), which can lead to sub-optimal reconstructions. As a result, the photometric stereo-based strategy has been barely explored in the last decade, as shown in Figure \ref{fig:CumNumOfPapers}. Nevertheless, the most recent works have taken advantage of the prior knowledge provided by statistical models to obtain a coarse reconstruction that is later refined with photometric methods, obtaining impressive results.

Indeed, the use of statistical models has been widely explored in the last decade (see Figure \ref{fig:CumNumOfPapers}), since they help obtaining plausible 3D faces by modelling the space of possible solutions, thus avoiding artefacts or unrealistic 3D facial reconstructions. The 3DMMs are the most wide\-spread statistical facial models, because of their capacity of encoding the geometric (and possibly texture) variations of the human face. Fitting a 3DMM to an image consists in estimating the model parameters by minimising the difference between the image and the projection of the 3DMM onto the image plane, with the help of a set of corresponding 2D-3D points. Most of the publications have approached the fitting as the minimisation of a non-linear cost function, which allowed them to easily incorporate as many terms as needed to further drive the reconstruction process. However, this optimisation is prone to get stuck in local minima and, generally, has a significant computational cost. These drawbacks have been identified by researchers who have proposed linear approaches to avoid them, although this often requires further assumptions, not always realistic. Furthermore, it has been found that estimating the parameters of a 3DMM is not sufficient to recover fine facial details, since these models usually encode global facial deformations. Thus, most 3DMM fitting approaches only reconstruct coarse shapes, and they are not able to recover subtle geometric details that are still important to achieve faithful reconstructions.

In the last few years, there has been a recent trend to explore deep learning algorithms that is rapidly growing, as can be seen in Figure \ref{fig:CumNumOfPapers}. These methods are especially well suited to learn non-linearities and, thus, have the ability to recover subtle details of each person, while keeping a realistic 3D face. Nevertheless, the main drawback of deep learning methods is the huge amount of data needed to train them. In the case of 3D-from-2D face reconstruction, the training dataset shall consist of pairs of 2D facial images and the corresponding 3D facial geometry (optionally with texture), but obtaining a sufficiently large collection of such pairs is challenging. For this reason, early works exploring neural networks for this purpose built synthetic training sets by fitting a 3DMM to real images and/or rendering synthetic images from randomly sampled faces (i.e. a kind of data augmentation). Even though this allowed researchers to train complex neural networks, it has two main drawbacks. Firstly, using the reconstruction from a 3DMM as ground truth limits the reconstruction power of the network to that of the 3DMM-fitting method. Secondly, training the network with synthetically generated images may lower its efficiency when dealing with real images.

As a consequence, the most recent works have reconsidered the need to have 3D ground truth and have addressed the training process in a self-supervised manner. To do so, they added a rendering layer at the end of the network that generates a synthetic image from the estimated textured 3D face and rendering parameters. This technique avoids the need of a large amount of paired 2D-3D data by minimising the difference between the input real image and the one rendered by the network. However, to be able to render a synthetic image from the estimated 3D face, the texture must be accurately recovered, which is a challenging task in its own. In addition, the learning must be regularised properly to restrict the space of solutions, since only the image error is not enough to ensure the uniqueness of the solution.

On the other hand, the lack of real data also limits the learning of fine details, which is currently a major concern of the community, since 3DMM-generated synthetic data is not as detailed as the real one, and learning them from the input image is greatly challenging.  Empirical findings have shown that the best reconstruction of fine details is obtained when they are estimated in a separate task from the global estimation of the 3D face. As a consequence, many researchers followed a coarse-to-fine approach, which allowed them to recover accurately detailed 3D faces, proving this strategy a promising research line.

% The first attempts of using deep learning to reconstruct 3D faces from images adopted single-network architectures, exploring the power of deeper networks and the benefits of iterative refinement of the reconstruction. Also, encoder-decoder architectures have been widely used, since they allow reconstructing the 3D face from more significant features extracted by the encoder. However, these approaches are not capable of recovering globally accurate 3D facial shapes and capturing fine details at the same time, without real training data. Furthermore,

Additionally, the community has begun to explore two other approaches just recently: GANs and GCNs. GANs have been explored from many different perspectives, since they allow training the 3D face reconstruction network to generate much more realistic 3D faces; however, given the lack of real 2D-3D paired data, how to adopt this type of architecture is still an open question. On the other hand, some researchers have identified a drawback common to all the presented works: 3D faces are usually represented by triangular meshes, which is a non-Euclidean data type, but classical deep learning operations can only deal with Euclidean data. Geometric deep learning is a recent research line that explores techniques to extend classical deep learning operations to non-Euclidean data, such as triangular meshes, which has motivated the appearance of a promising new approach to directly reconstruct 3D facial meshes from the input images.

Even though a wide range of different strategies has been presented, a tool that is common to most of them is the 3DMM since it captures the highly complex variations of the human face into a linear subspace of reduced dimension. However, including 3DMMs in the reconstruction process has two drawbacks. Firstly, the quality of the 3DMMs is highly dependant on the type of shape variations included in the training set, since facial shapes can be very different between subjects of different ages, gender, ethnicity, and even expressions. Secondly, 3DMMs model facial deformations globally, which limits their ability to reconstruct fine details. This limitation is also inherited by deep learning methods that employ 3DMMs either to generate synthetic training data or to constrain their internal (latent) representation.

% This last limitation is evident in deep learning methods. As stated above, some works trained their networks with 3DMM-generated synthetic dataset, which causes them to learn to represent 3D faces as modelled by the low-dimensional space provided by the 3DMMs, missing fine details. On the other hand, works that trained a neural network to estimate 3DMM parameters do not overcome the limitations of classical 3DMM-fitting methods, also resulting in coarse reconstructions.

For this reason, 3D facial models and 3D-from-2D face reconstruction are two closely related research fields that have been developed cooperatively. 3D facial models allow ensuring the plausibility of the 3D faces reconstructed from 2D images, which boosted the field forward. On the other hand, 3D face reconstruction methods are limited by the representation provided by facial models, leaving the reconstruction of fine details an open problem. Therefore, improving the modelling power by means of local or non-linear models may help obtaining more accurate detailed 3D reconstructions.

\appendix
\section{Notation} \label{sec:Notation}

\subsection{3D facial model}
% {\doublespacing
\begin{itemize}
    \item[] $M$: number of faces in the model
    \item[] $N$: number of vertices of the meshes
\end{itemize}
Shape model%, $i = 1,\dots,\widetilde{M}$
\begin{itemize}
    \item[] $\vec{x}$: shapes in 3D
    \item[] $\widetilde{M}_\vec{x}$: number of eigenvectors in the 3D facial shape model
    \item[] $\vec{\phi}_i \in \mathbb{R}^{3N}$ : eigenvectors of 3D facial shape model
    \item[] $\vec{\Phi} =
\begin{pmatrix}
\vdots & & \vdots\\
\vec{\phi}_1 & \cdots & \vec{\phi}_{\widetilde{M}_\vec{x}}\\
\vdots & & \vdots
\end{pmatrix} \in \mathbb{R}^{(3N) \times \widetilde{M}_\vec{x}}$
    \item[] $\sigma_{\alpha_i}^2 \in \mathbb{R}$ : eigenvalues of shape model
    \item[] $\alpha_i \in \mathbb{R}$ : shape model parameters
    \item[] $\vec{\alpha} = (\alpha_1, \dots, \alpha_{\widetilde{M}_\vec{x}})$
    \item[] $\vec{x}(\vec{\alpha})$: 3D face generated by $\vec{\alpha}$
    % \item[] $\vec{\delta}$: expression model parameters
    % \item[] $\vec{\Upsilon}$: expression model eigenvectors
\end{itemize}
Texture model%, $i = 1,\dots,\widetilde{N}$
\begin{itemize}
    \item[] $\vec{c}$: albedo of a 3D face (a.k.a texture or colour)
    \item[] $\widetilde{M}_\vec{c}$: number of eigenvectors in the 3D facial texture model
    \item[] $\vec{\psi}_i \in \mathbb{R}^{3N}$ : eigenvectors of texture model
    \item[] $\vec{\Psi} \in \mathbb{R}^{(3N) \times \widetilde{M}_\vec{c}}$
    \item[] $\sigma_{\beta_i}^2 \in \mathbb{R}$ : eigenvalues of texture model
    \item[] $\beta_i \in \mathbb{R}$ : texture model parameters
    \item[] $\vec{\beta} = (\beta_1, \dots, \beta_{\widetilde{M}_\vec{c}})$
    \item[] $\vec{\gamma}_\text{d}$: diffuse lighting parameters
    \item[] $\vec{\gamma}_\text{s}$: specular lighting parameters
\end{itemize}
% }

\subsection{3D facial model fitting}
% {\doublespacing
\begin{itemize}
    \item[] $\Iin$: input image
    \item[] $\Imod$: rendered image from reconstructed 3D face
    \item[] $L$: number of landmarks
    \item[] $\lmk$: landmarks in the input image
    \item[] $\Lmk$: landmarks in the 3D face
    \item[] $\LmkProj$: projection of landmarks in the 3D face $\Lmk$ into the image plane
    \item[] $\lmkProj$: projection of landmarks in the image $\lmk$ into the 3D face
    \item[] $\vec{\kappa}$: intrinsic camera parameters (depends on the camera model, but usually focal length, principal point, ...)
    \item[] $\vec{R}$: rotation matrix
    \item[] $\vec{\tau}$: translation
    \item[] $\vec{\zeta} = \{ \vec{R}, \vec{\tau}, \vec{\kappa}\}$: projection parameters (intrinsic and extrinsic)
    \item[] $\vec{\rho}$: rendering parameters (including $\vec{\kappa}$, $\vec{R}$, $\vec{\tau}$ and illumination parameters)
    \item[] $\mathcal{P}(\vec{v},\vec{\zeta})$: projection of $\vec{v} \in\mathbb{R}^3$ according to parameters $\vec{\zeta}$
    \item[] $\errorI$: distance between $\Iin$ and $\Imod$
    \item[] $\errorlmk$: distance between $\lmk$ and $\LmkProj$
    \item[] $\errorLmk$: distance between $\Lmk$ and $\lmkProj$
    \item[] $\errorReg$: regularisation of the fitting parameters
\end{itemize}
% }
\section*{Acknowledgements}
This work is partly supported by the Spanish Ministry of Economy and Competitiveness under project grant TIN2017-90124-P, the Ramon y Cajal programme, the Maria de Maeztu Units of Excellence Programme (MDM-2015-0502).

% \section*{References}

\bibliography{journal-abbreviations,Bibliography}

\end{document}